\documentclass{article}
\usepackage{iclr2026_conference,times}
\iclrfinalcopy
\renewcommand{\iclrruler}[1]{}
\usepackage[T1]{fontenc}
\input{glyphtounicode}
\usepackage{amsmath,amsfonts,bm}

\def\eqref#1{equation~\ref{#1}}

\def\1{\bm{1}}

\DeclareMathAlphabet{\mathsfit}{\encodingdefault}{\sfdefault}{m}{sl}
\SetMathAlphabet{\mathsfit}{bold}{\encodingdefault}{\sfdefault}{bx}{n}

\usepackage[dvipsnames,svgnames,table]{xcolor}
\usepackage[most]{tcolorbox}
\usepackage{etoc}
\usepackage{placeins}
\usepackage{dblfloatfix}
\usepackage{fvextra}
\usepackage{xspace}
\usepackage{amssymb}
\usepackage{booktabs}
\usepackage{caption}
\usepackage{threeparttable}
\usepackage{tabularx}
\usepackage{array}
\usepackage{enumitem}
\usepackage{multirow}
\usepackage{graphicx}

\setlist[itemize]{leftmargin=0pt}

\usepackage{titlesec}

\AddToHook{cmd/ttfamily/after}{\frenchspacing}

\usepackage{pifont}
\definecolor{checkred}{HTML}{9E0000}
\definecolor{checkgreen}{HTML}{005E22}
\newcommand{\cmark}{{\color{checkgreen} \ding{51}}}
\newcommand{\xmark}{{\color{checkred} \ding{55}}}

\newcommand{\greendelta}[1]{\,{\scriptsize(\textcolor{checkgreen}{\ensuremath{+}#1})}}
\newcommand{\reddelta}[1]{\,{\scriptsize(\textcolor{checkred}{\ensuremath{-}#1})}}

\definecolor{sh-color}{HTML}{ff00a0}

\definecolor{yoonho-color}{HTML}{2104af}
\definecolor{bo-color}{HTML}{0B7A3B}

\definecolor{dy-color}{HTML}{F54C4F}

\definecolor{akari-color}{HTML}{ce42f5}

\definecolor{rs-color}{HTML}{e87ea1}

\definecolor{benchmarkyellow}{HTML}{FFD75B}
\definecolor{benchmarkplum}{HTML}{A52887}

\definecolor{promptlemon}{HTML}{FDF1FF}
\DefineVerbatimEnvironment{VerbatimWrap}{Verbatim}{
breaklines=true,
breakindent=0pt,
breaksymbol={},
fontsize=\footnotesize,
}
\newtcolorbox{plumpromptframe}[2][]{
  enhanced jigsaw,
  breakable,
  colback=promptlemon,
  colframe=black,
  colbacktitle=black,
  coltitle=white,
  fonttitle=\small,
  fontupper=\fontfamily{cmtt}\selectfont\footnotesize,
  boxrule=1pt,
  arc=3.5pt,
  boxsep=0pt,
  left=10pt, right=10pt, top=7pt, bottom=7pt,
  lefttitle=8pt,
  righttitle=8pt,
  toptitle=3pt,
  bottomtitle=3pt,
  titlerule=1pt,
  titlerule style={draw=black, fill=black},
  title=#2,
  notitle after break,
  before skip=15pt,
  after skip=15pt,
  #1
}
\newenvironment{plumpromptbox}[1]
  {\VerbatimEnvironment\begin{plumpromptframe}{#1}\begin{VerbatimWrap}[fontfamily=cmtt]}
  {\end{VerbatimWrap}\end{plumpromptframe}}

\definecolor{yellow}{HTML}{FFD247}
\definecolor{cream}{HTML}{FFF6E9}
\definecolor{mainsettingfill}{HTML}{FFFBEA}

\newtcolorbox{mainsettingbox}{
    enhanced,
    colback=mainsettingfill,
    colframe=black,
    boxrule=0.8pt,
    arc=1.5mm,
    left=3mm, right=3mm,
    top=1.2mm, bottom=1.2mm,
    fontupper=\small,
}

\newcommand{\providericon}[1]{
  \raisebox{-0.43ex}{
    \includegraphics[height=1.05em]{figs/logo/#1.pdf}
  }
  \hspace{0.25em}
}

\newcommand{\hflink}[1]{\href{#1}{\raisebox{-0.3ex}{\includegraphics[height=1.0em]{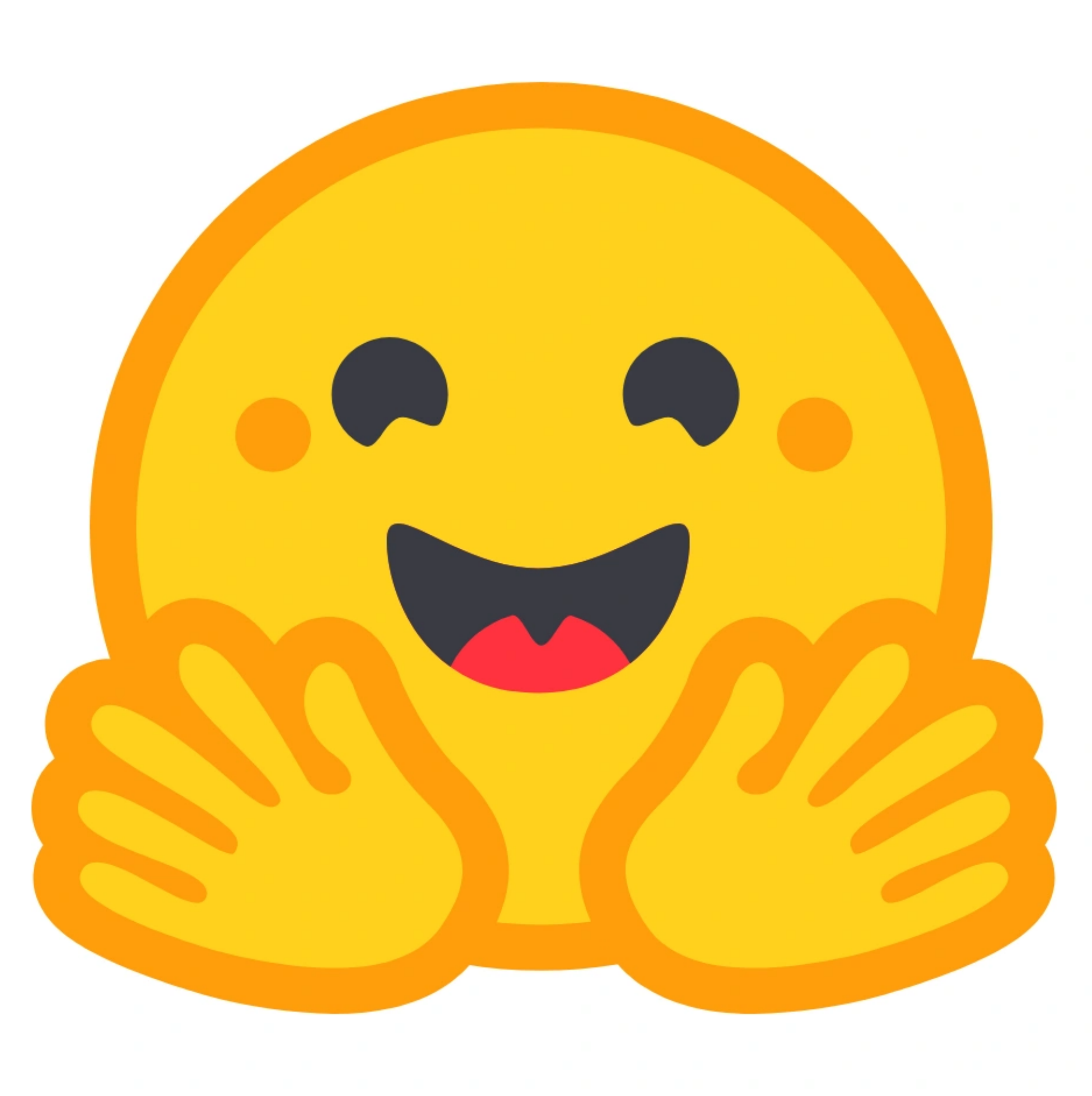}}}}
\newcommand{\githublink}[1]{\href{#1}{\raisebox{-0.3ex}{\includegraphics[height=1.0em]{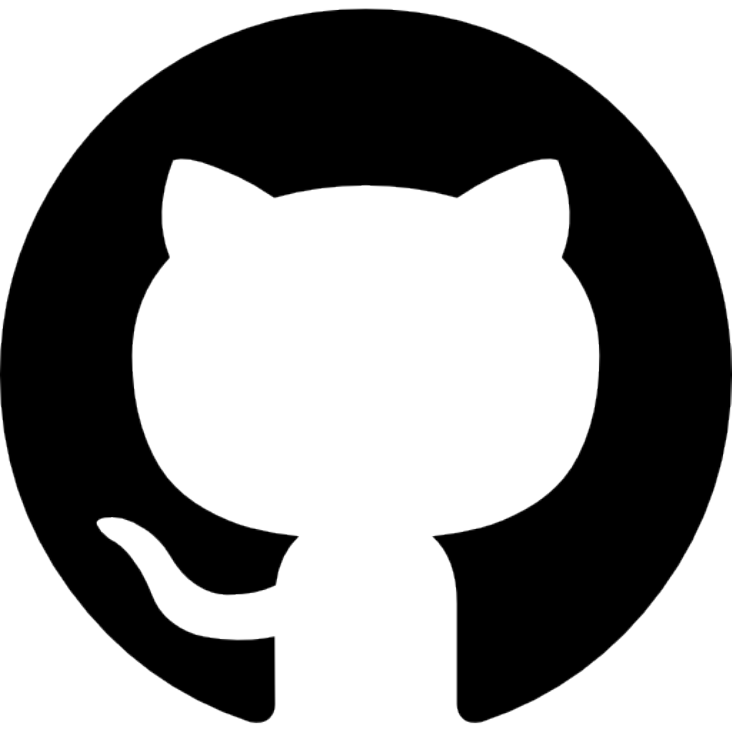}}}}

\newcommand{\resourcefont}{\fontfamily{pcr}\fontsize{9.5pt}{11pt}\selectfont}
\newcommand{\resourcelink}[2]{\href{#1}{{\resourcefont\textcolor{black!85}{#2}}}}

\newcommand{\modelgroup}[2]{
  \rowcolor[HTML]{F2F2F2}
  \multicolumn{#1}{@{}l}{
    \textcolor[HTML]{606060}{#2}
  }\\[0.45ex]
}

\definecolor{citecolor}{HTML}{8D1A71}
\usepackage{hyperref}
\definecolor{refcolor}{HTML}{8D1A71}
\hypersetup{
    bookmarksopen=true,
    bookmarksnumbered=true,
    colorlinks=true,
    citecolor=citecolor,
    linkcolor=refcolor,
    urlcolor=citecolor,
}
\usepackage[capitalize,noabbrev,nameinlink]{cleveref}
\crefname{section}{Section}{Sections}
\crefname{subsection}{Section}{Sections}
\crefname{subsubsection}{Section}{Sections}
\crefname{appendix}{Appendix}{Appendices}
\crefname{subappendix}{Appendix}{Appendices}
\crefname{subsubappendix}{Appendix}{Appendices}
\crefname{figure}{Figure}{Figures}
\crefname{table}{Table}{Tables}
\crefname{equation}{Eq.}{Eqs.}

\newcommand{\benchmark}{\textsc{ScholarCatalyst}}

\newcommand{\rqlevel}{core research query}
\newcommand{\threadlevel}{subfield-specific query}

\newcommand{\nparticipant}{184}
\newcommand{\nauthor}{207}

\definecolor{cqcolor}{HTML}{7A005E}
\definecolor{cqfill}{HTML}{F9EEF6}
\definecolor{sqcolor}{HTML}{008780}
\definecolor{sqfill}{HTML}{ECF7F7}

\pdfmapline{+optimistic-t1 <figs/fonts/T1-WGL4.enc <figs/fonts/Optimistic.ttf}
\newfont{\optimisticfont}{figs/fonts/optimistic-t1 at 8.5pt}
\newcommand{\CQ}{\textcolor{cqcolor}{\optimisticfont CoreQ}}
\newcommand{\SQ}{\textcolor{sqcolor}{\optimisticfont SubQ}}

\newcommand{\logoC}{\raisebox{-0.13em}{\includegraphics[height=1em]{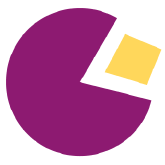}}\hspace{-0.1em}}

\newcommand{\authorblockindent}{}
\newcommand{\eqstar}{\scalebox{1}[0.78]{\ensuremath{\textcolor{benchmarkplum}{\scriptstyle\blacklozenge}}}\ensuremath{\mkern1mu}}
\newcommand{\contributordiamond}{\raisebox{0.05ex}{\smash{\scalebox{1.28}[1.00]{\ensuremath{\textcolor{benchmarkplum}{\blacklozenge}}}}}}

\title{
Scholar\logoC atalyst: A Benchmark for\\ Retrieving Papers that Inspire New Research
\vspace{-1mm}
}
\author{%
{\bf%
\authorblockindent 
    Sohyeon Kim$^{\eqstar1,2}$
    \quad
    Yoonho Lee$^{\eqstar1}$
    \quad
    Bo Liu\,$^{1,3}$
    \quad
    Dayoon Ko\,$^{2}$
    \quad
    Rulin Shao\,$^{3}$
    \vspace{2mm}
} \\%
{\bf%
\authorblockindent 
    Seungone Kim\,$^{4}$
    \quad
    Graham Neubig\,$^{4}$
    \quad
    Pang Wei Koh\,$^{3}$
    \quad
    Aakanksha Chowdhery\,$^{1}$
    \vspace{2mm}
} \\%
{\bf%
\authorblockindent 
    Akari Asai\,$^{4,5}$
    \quad
    Omar Khattab\,$^{6}$
    \quad
    Yejin Choi\,$^{1}$
    \quad
    Gunhee Kim\,$^{2}$
    \quad
    Chelsea Finn\,$^{1}$
} \\[8pt]%
{%
\authorblockindent $^{1}$ Stanford University \quad
    $^{2}$ Seoul National University \quad
    $^{3}$ University of Washington
} \\[1pt]%
{%
\authorblockindent $^{4}$ Carnegie Mellon University \quad
    $^{5}$ Allen Institute for AI \quad
    $^{6}$ MIT
}}

\begin{document}
\etocdepthtag.toc{mainpart}
\maketitle
{\renewcommand{\thefootnote}{\eqstar}\footnotetext{Equal contribution. Contact: \texttt{\{sohyeon,yoonho\}@stanford.edu}}}

\vspace{-1mm}
\begin{abstract}
\vspace{-3mm}
What makes great scientists great?
Even as AI systems start to make progress on open problems, scientists remain far ahead of them at sensing which prior idea, buried in an ever-growing archive of research, a new problem needs.
To study this skill, we draw on researchers who know firsthand which earlier work advanced their completed projects, with papers serving as pointers to the ideas within.
Using our automated pipeline that makes author annotation scalable, we build \textbf{\benchmark{}} by having \nparticipant{} lead authors of \nauthor{} recent computer science papers label which candidates did or could have advanced their project, each with a detailed rationale.
We introduce a \textit{retrieval task} with author-provided judgments: given an initial research question, retrieve these papers from only the literature available when the project began.
Agentic search does no better than embedding retrieval (0.42 vs. 0.48 Recall@20) despite calling that same retriever as a tool.
Even an agent built on Claude Fable 5.1, which may have seen the completed papers during training, reaches only 0.51 R@20.
These results highlight the need for new training recipes that equip models with expert intuition for searching broad corpora.
We envision \benchmark{} as a step toward scientific agents that can take a half-formed idea and point to the prior research it needs.
\end{abstract}
\vspace{-1mm}
\begin{center}
{\small
\renewcommand{\resourcefont}{\fontfamily{pcr}\fontsize{8.5pt}{10pt}\selectfont}%
\renewcommand{\arraystretch}{1.05}%
\begin{tabular}{@{}r@{\hspace{0.8em}}c@{\hspace{0.8em}}l@{}}
\textbf{Project Page} & \githublink{https://github.com/stanford-iris-lab/ScholarCatalyst} & \resourcelink{https://ohmyksh.github.io/project/ScholarCatalyst/}{ohmyksh.github.io/project/ScholarCatalyst} \\
\textbf{Dataset} & \hflink{https://huggingface.co/ScholarCatalyst} & \resourcelink{https://huggingface.co/ScholarCatalyst}{huggingface.co/ScholarCatalyst} \\
\textbf{Code} & \githublink{https://github.com/stanford-iris-lab/ScholarCatalyst} & \resourcelink{https://github.com/stanford-iris-lab/ScholarCatalyst}{github.com/stanford-iris-lab/ScholarCatalyst} \\
\textbf{Contributors} & \contributordiamond & {\resourcefont\hypersetup{linkcolor=black!85}\hyperref[app:contributors]{Full list in Appendix}}
\end{tabular}}
\end{center}
\vspace{0mm}

\begin{figure}[h!]
\vspace{0.5mm}
\centering
\includegraphics[width=0.84\linewidth]{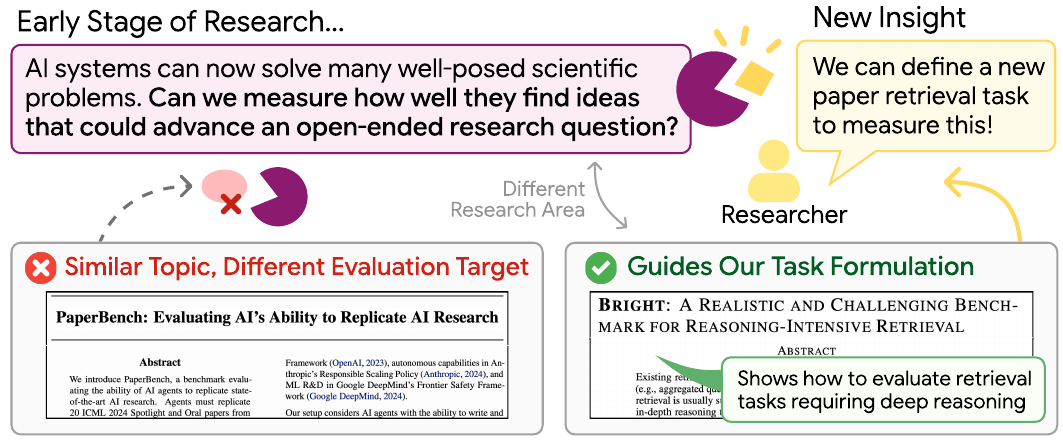}
\vspace{-2mm}
\caption{\textbf{\benchmark{}} asks whether a system can retrieve prior papers that inspire progress on a research project. Such a paper may share little semantic similarity with the question yet offer an insight that advances the project (\textit{right}), while topically related work may not (\textit{left}).
}
\label{fig:benchmark_task}
\end{figure}

\clearpage
\section{Introduction}
{ \vspace{-8.5mm}
\raggedleft
\textit{To invent is to discern, to choose.}---~Henri Poincar\'e\par}
\vspace{2mm}

Scientific research is inherently cumulative, with each project building on previous results and ideas.
Current AI systems for research mainly accelerate progress on an already well-defined subproblem, where the remaining work is closer to engineering.
Deciding which existing idea a new problem should build on is still mostly left to human researchers.
An insightful prior paper can turn a vague question into a well-defined problem, for instance by revealing a limitation that motivates a new direction or suggesting an approach worth adapting, thereby shaping and accelerating---\textbf{\textit{catalyzing}}---subsequent research.
Making this choice well is a defining ability of great scientists, and a critical part of what researchers call \textit{research taste}.
We ask whether models can make that decision through a concrete task: given an early-stage research question, find its \textit{catalyst papers}, by which we mean prior papers whose ideas did or could have meaningfully advanced the project (\cref{fig:benchmark_task}).

Evaluating this retrieval task requires ground truth about which papers were catalysts for a particular project.
Published papers record the outcomes of completed projects, but the ideation that shaped them is largely undocumented, and bibliographies rarely distinguish the sources that provided non-obvious, critical insights from the broader set of cited works.
Existing scientific retrieval benchmarks (Table~\ref{tab:benchmark_comparison}) evaluate facet-level similarity or relevance to a stated information need~\citep{wadden2020fact,wang2023scientific,ajith2024litsearch,asai2024openscholar} or derive positives from citation-intent annotations for queries that withhold the eventual method~\citep{garikaparthi2025mir}.
These signals do not capture whether a retrieved paper could have inspired an early-stage project, especially when it was not cited by the eventual paper.

We introduce \benchmark{}, a literature inspiration retrieval benchmark grounded in researchers' firsthand knowledge of their own projects.
\nparticipant{} researchers who led \nauthor{} recent computer science projects verified 894 research questions as they stood before each project's key findings, then labeled which prior papers did or could have advanced their work and explained why, including papers they had not encountered at the time.
Given a question, a system must rank these papers near the top of a 191K-paper corpus restricted to work published before the project, using models whose knowledge cutoffs predate every source paper.
This knowledge is rarely documented, and we collect it efficiently and at scale through an automated pipeline that only needs a paper's arXiv ID to produce author-reviewable draft data.
Because any newly published paper can become a fresh instance, the pipeline keeps \benchmark{} ahead of model training cutoffs and lets the benchmark grow with the literature.

Across sparse, dense, and multi-vector retrievers and LLM search agents, the strongest system recovers only 48\% of gold (i.e., author-credited) papers in its top 20.
We find that topical similarity is a weak signal: catalyst papers are no more similar to the query than related papers the authors rejected.
Every search an agent issues still goes through a similarity-based retriever, so repeated querying does not help (42\% vs. 48\% for the retriever alone).
We also evaluate Claude Fable 5.1 as a lenient upper reference because its training data may include the source papers or secondary sources describing them, and even with this advantage it misses nearly half of the gold papers in its top 20.
More broadly, our results point to the need for retrieval models with \textit{an expert-level sense of which ideas matter}, and \benchmark{} offers a way to measure progress toward them.

\begin{table*}[!b]
\centering
\scriptsize
\setlength{\tabcolsep}{3pt}
\renewcommand{\arraystretch}{1.12}
\vspace{0mm}
\caption{\textbf{Comparison with literature retrieval benchmarks.}
\benchmark{} combines queries that precede a project's solution (\emph{Pre-discovery query}), direct author annotation (\emph{Author annotation}), and relevance judgment beyond what is published in the completed source paper (\emph{Undocumented relevance}).}
\vspace{-2mm}
\label{tab:benchmark_comparison}
\definecolor{Ours}{HTML}{FFEDA3}
\begin{threeparttable}
\begin{tabularx}{\textwidth}{
@{}
>{\raggedright\arraybackslash}m{0.26\textwidth}
>{\raggedright\arraybackslash}X
>{\centering\arraybackslash}m{0.09\textwidth}
>{\centering\arraybackslash}m{0.08\textwidth}
>{\centering\arraybackslash}m{0.12\textwidth}
>{\centering\arraybackslash}m{0.065\textwidth}
>{\centering\arraybackslash}m{0.075\textwidth}
@{}
}
\toprule
Benchmark
& Relevance
& \shortstack{Pre-discovery\\query}
& \shortstack{Author\\annotation}
& \shortstack{Undocumented\\relevance}
& \shortstack{Queries\\$|\mathcal{Q}|$}
& \shortstack{Corpus\\$|\mathcal{P}|$} \\
\midrule

{SciFact}~\citep{wadden2020fact}
& Claim support
& \xmark
& \xmark
& \xmark
& 1.4K
& 5.2K \\

{DORIS-MAE}~\citep{wang2023scientific}
& Multi-aspect relevance
& \xmark
& \xmark
& \xmark
& 100
& 100 \\

{ScholarQABench}~\citep{asai2024openscholar}
& Claim support
& \xmark
& \xmark
& \xmark
& 2{,}967
& 45M \\

{LitSearch}~\citep{ajith2024litsearch}
& Citation relation
& \xmark
& \cmark
& \xmark
& 597
& 64K \\

{MIR}~\citep{garikaparthi2025mir}
& Methodological citation intent
& \cmark
& \xmark
& \xmark
& 139
& 4.7K \\

\rowcolor{Ours}
{\textbf{ScholarCatalyst (ours)}}
& Author-judged inspiration
& \cmark
& \cmark
& \cmark
& 894
& 191K \\
\bottomrule
\end{tabularx}
\begin{tablenotes}[flushleft]
\scriptsize
\item[\textdagger] LitSearch asks authors to write 246 of its 597 queries to retrieve their own papers.
\end{tablenotes}
\end{threeparttable}
\vspace{2mm}
\end{table*}

\section{Related Work}

\paragraph{Scientific discovery using LLMs.}
Accelerating science is a major opportunity for LLM-based agents, which increasingly can propose relevant hypotheses and research  ideas~\citep{wang2024scimon,wang2024scipip,liu2025harnessing,liu2026researchbench}.
We deliberately ground our task in existing \textit{papers}.
Unlike a newly generated idea, which cannot be verified without running it~\citep{ICLR2025_ea94957d,si2026ideation}, a published paper comes with a record of what worked, why, and where it fell short.
The same grounding is what lets us collect ground truth, since authors can judge existing papers against what actually advanced their project.
Existing scientific-discovery systems likewise draw on prior papers, either starting from a seed or target paper and exploiting its citation graph or metadata~\citep{wang2024scimon,liu2025harnessing}, while others search a task-specific candidate corpus constructed in advance~\citep{wang2024scipip,liu2026researchbench}.
Recent work studies how LLMs can generate scientific insights by combining existing ideas~\citep{he2026giants}, further motivating the need to evaluate which prior work agents choose to draw on.
\benchmark{} directly addresses this evaluation gap by testing whether scientific agents can retrieve insightful prior ideas for an early-stage research question, guided by project authors' firsthand experience of what helped their research.

\paragraph{Scientific literature retrieval.}
Existing scientific retrieval benchmarks evaluate how well systems find evidence for claims~\citep{wadden2020fact,wadden2022scifact}, retrieve documents for specified facets~\citep{mysore2021csfcube,wang2023scientific}, and locate papers given document context or explicit information needs~\citep{ajith2024litsearch,asai2024openscholar,he2025pasa,park2025chain,shao2025scinet}.
Whereas existing benchmarks provide a retrieval objective that is explicit in the query, we pose an open-ended problem that requires inferring the latent scientific relevance of papers to an early-stage research idea.
MIR~\citep{garikaparthi2025mir} comes closest, since its queries also withhold the project's eventual method, but it derives positives from citation-intent annotations indicating methodological influence~\citep{lauscher2022multicite}.
Rather than inferring relevance from the published record, we elicit directly from paper authors otherwise undocumented data about the research process behind it, including early-stage research questions and relevance judgments.
To our knowledge, \benchmark{} is the first scientific literature retrieval benchmark grounded in authors' firsthand accounts of their research process.

\paragraph{Information retrieval systems.}
Lexical retrievers~\citep{robertson2009probabilistic} and embedding retrievers rank corpus papers through term overlap or learned semantic similarity~\citep{karpukhin2020dense,khattab2020colbert,zhang2025qwen3, shanbhogue2026geminiembedding2native}.
For a given query, these methods rank the corpus once and do not adapt the search to the returned papers.
Search agents use LLM reasoning to iteratively formulate queries and invoke these retrievers as tools over multiple turns, enabling more adaptive search for complex information-seeking tasks~\citep{yao2022react,jin2025search,li2025search}.
Deep research systems extend this approach with subquestion decomposition, reading, and reflection over longer horizons~\citep{team2025tongyi,zhang2025deep}.
However, these agents still access the corpus through a retriever's top-$k$ candidates at each step.
Recent work instead gives agents direct access to the raw corpus through terminal tools such as grep and file reads, letting the language model determine how to search and interpret documents~\citep{li2026beyond,sen2026grep}.
Whether such agentic search approaches can retrieve prior work whose scientific inspiration is latent rather than lexically explicit remains unclear.

\section{ScholarCatalyst: Evaluating Scientific Literature Retrieval with Authors' Firsthand Accounts}
\label{sec:benchmark}

\benchmark{} evaluates whether retrieval systems can identify \textit{catalyst papers}, prior work whose ideas could inspire and help advance an early stage of research.
To obtain firsthand judgments about inspiration, we ask project authors which prior papers did or could have helped advance their project and why, knowledge that is largely absent from the public record.
We first define the task of retrieving useful prior work, including its query types, retrieval corpus, and evaluation metrics~(\cref{sec:bench:formulation}).
We then describe our pipeline that supports effective and efficient data collection (\cref{sec:bench:pipeline}) and the subsequent author validation and revision process for finalizing the benchmark instances (\cref{sec:bench:author_data}).

\begingroup
\setlength{\textfloatsep}{10pt plus 2pt minus 2pt}

\begin{figure}[t!]
\centering
\includegraphics[width=\linewidth]{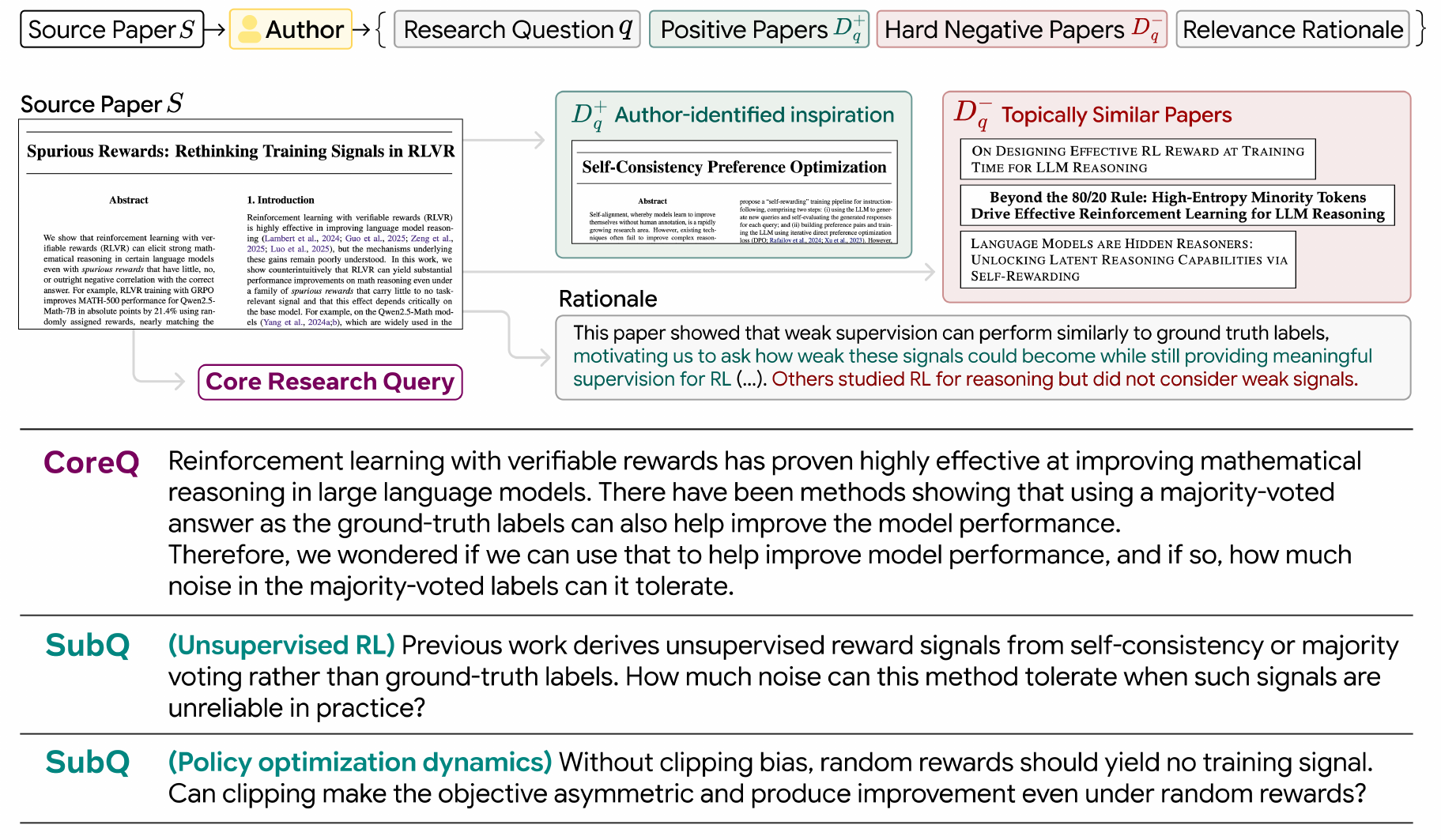}
\vspace{-5mm}
\caption{\textbf{An example \benchmark{} instance.}
An author of the source paper $S$ provides a research question $q$, expressed as a \CQ{} or \SQ{}, along with positive and hard negative papers $\mathcal{D}_q^+, \mathcal{D}_q^-$ and a rationale for these judgments.}
\label{fig:benchmark_example}
\end{figure}
\vspace{-10pt}

\subsection{Task Definition}
\label{sec:bench:formulation}
\newcommand{\tightset}[1]{\smash{#1}}
Given a query $q$ from the authors of the source paper $S$, the task is to retrieve catalyst papers $\tightset{\mathcal{D}^{+}_{q}}$ from $\tightset{\mathcal{P}_{<S}}$, where $\tightset{\mathcal{P}_{<S}}$ is a corpus containing papers published before source paper $S$ was completed.
For each query $q$, we also provide hard-negative documents, $\tightset{\mathcal{D}^{-}_{q} \subset \mathcal{P}_{<S}}$, which are related to $q$ but did not inspire the authors of $S$.

Each query $q$ is a short paragraph that reconstructs the research landscape at the outset of the project underlying the source paper $S$.
It describes what was known, what remained unresolved, and the open question the authors set out to answer, without revealing the eventual solution.
Specifically, we consider two query types that reflect how an early-stage project develops from a broad central research question into particular subfields, with an inspiring paper potentially shaping the next step at either stage.
\begin{itemize}[leftmargin=1.5em, topsep=0.25em, partopsep=0pt, itemsep=0.3em, parsep=0pt]
    \item A {\textbf{\rqlevel{}}} (\CQ{}) asks which prior ideas could help address the project's central research question.
    \item A {\textbf{\threadlevel{}}} (\SQ{}) approaches the same research question from a particular research subfield, asking which prior ideas could help advance the project from that perspective.
\end{itemize}

\Cref{fig:benchmark_example} shows an example \benchmark{} instance $(q, \tightset{\mathcal{D}^{+}_{q}}, \tightset{\mathcal{D}^{-}_{q}})$ derived from source paper $S$.
For each query $q$, the positive paper set $\tightset{\mathcal{D}^{+}_{q} \subseteq \mathcal{P}_{<S}}$ contains papers that authors judge to offer inspiration for pursuing $q$.
The hard-negative set $\tightset{\mathcal{D}^{-}_{q} \subseteq \mathcal{P}_{<S} \setminus \mathcal{D}^{+}_{q}}$ contains papers that the authors reviewed and rejected.
Every hard-negative paper is topically related to $q$, since it was either cited in $S$ or ranked highly for $q$ by a retriever.

By making the retrieval corpus $\tightset{\mathcal{P}_{<S}}$ contain only papers published before $S$~\citep{yetisgen2008evaluation,yetisgen2009new,halawi2024approachinghumanlevelforecastinglanguage}, we approximate the literature available during the research process.
We also disable external web access to prevent retrieval of subsequent writing about the project and the source paper.
We evaluate retrieval performance using Recall@$N$, which measures the fraction of positive papers retrieved in the top $N$ results, and normalized discounted cumulative gain (nDCG@$N$)~\citep{Jrvelin2002CumulatedGE}, which additionally rewards higher rankings of positive papers.

\endgroup

\subsection{Our Automated Pipeline for Efficient Data Collection}
\label{sec:bench:pipeline}
We build an automated data construction pipeline that \textit{only requires a source paper's arXiv ID} and produces the materials required for author annotation.
The pipeline let us collect author-validated data at scale.
It also makes it easy to refresh \benchmark{} with newly published papers as older instances go stale due to leakage into newer models' training data.

\textbf{Source-paper and corpus processing.}
For each source paper \(S\), we use the arXiv API\footnote{\url{https://info.arxiv.org/help/api/index.html}} to obtain metadata and source files, parse the full text into a structured representation, and resolve bibliography entries to canonical scholarly records.
The resulting one-hop citation graph links inline citations to the referenced papers' metadata and abstracts.
Our retrieval corpus \(\mathcal{P}\) contains 190{,}896 papers.
It combines references resolved from our source papers with 181K additional arXiv papers from 2020--2024 across major computer-science categories.
Scoping \(\mathcal{P}\) to computer science and the source papers' publication years keeps the corpus relevant to \benchmark{} without indexing all of arXiv, which now exceeds 3 million articles~\citep{arxiv2026milestone}.
We show detailed corpus statistics in~\cref{app:dataset-stats}.

\textbf{Pre-annotation and candidate construction.}
We use Gemini 3.1 Pro to generate pre-annotations from the structured source paper \(S\) and its bibliography.
These include one \CQ{}, several \SQ{}s, and rationales explaining the proposed relevance of cited papers.
We additionally retrieve uncited candidates with BM25 and Qwen3-Embedding-8B~\citep{robertson2009probabilistic,zhang2025qwen3}.
Gemini 3.6 Flash~\citep{google2026gemini36flash} reranks the pooled candidates for scientific relevance, and the pipeline retains 10 candidates per query to balance coverage of potential false negatives with a manageable author-review workload.

The resulting pre-annotations and candidates are loaded into a web interface for author review (\cref{sec:bench:author_data}).
We describe the implementation details of our pipeline in~\cref{app:reconstruction}.

\begin{figure}[t]
\centering
\includegraphics[width=\columnwidth]{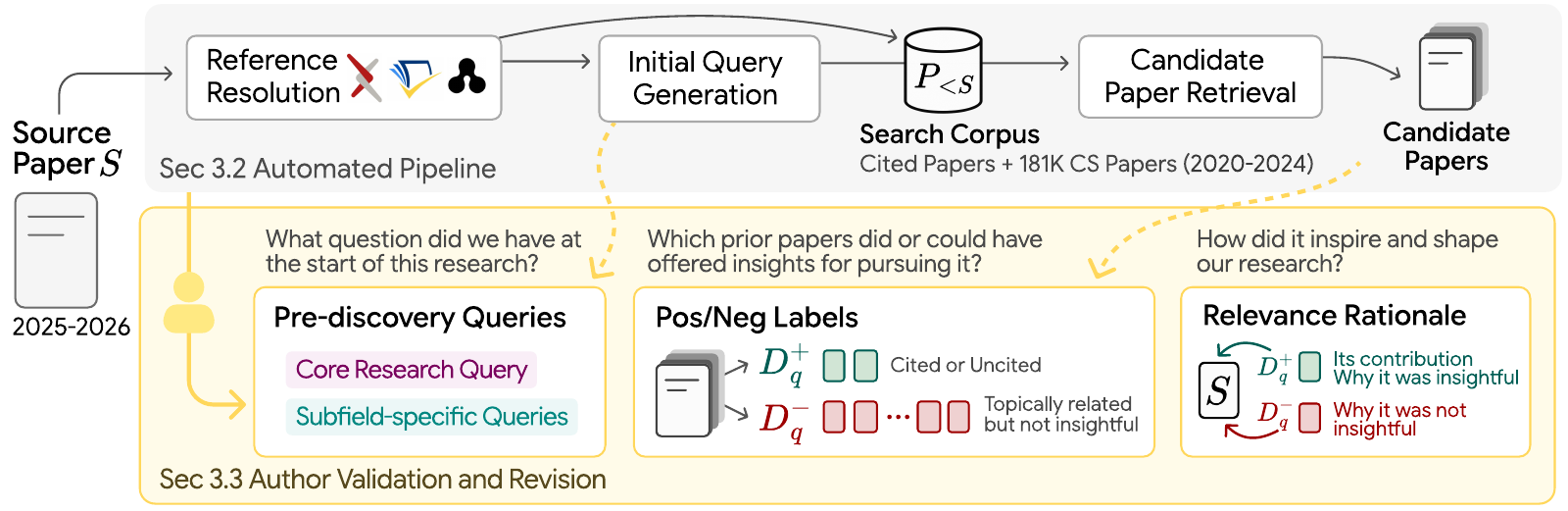}
\caption{
\textbf{\benchmark{} construction process.}
Our automated data-collection pipeline constructs a retrieval corpus and author-review materials, including structured paper data and candidate papers (\cref{sec:bench:pipeline}), which authors review through our interface to validate and revise queries, judge candidate-paper relevance, and provide rationales (\cref{sec:bench:author_data}).
}
\label{fig:benchmark_construction}
\end{figure}

\begin{table}[t]
\centering
\small
\setlength{\tabcolsep}{5pt}
\renewcommand{\arraystretch}{1.1}
\caption{
\textbf{Dataset statistics.}
For each query type, we report the number of queries, the average number of positive papers ($|\mathcal{D}^{+}_{q}|$) and hard negatives ($|\mathcal{D}^{-}_{q}|$), and the average query length in tokens.
}
\label{tab:corpus_stats}
\begin{tabular}{lrrrrr}
\toprule
Query type & Queries & Avg. $|\mathcal{D}^{+}_{q}|$  & Cited / Uncited Pos. (\%) & Avg. $|\mathcal{D}^{-}_{q}|$  & Avg. Length \\
\midrule
{\textcolor{cqcolor}{\optimisticfont Core Research Query}} & 207 & 3.19 & 95.5 / 4.5 & 10.01 & 131.8 \\
{\textcolor{sqcolor}{\optimisticfont Subfield-specific Query}} & 687 & 6.09 & 56.4 / 43.6 & 6.88 & 63.5 \\
\bottomrule
\end{tabular}
\end{table}

\subsection{Collecting Relevance Supervision from Authors}
\label{sec:bench:author_data}

We collect \nauthor{} projects from \nparticipant{} researchers, each describing the intellectual path behind a \textit{source paper} $S$ in computer science (CS) and artificial intelligence (AI), published in 2025--2026.
Each project contributes one \CQ{} task instance and one or more \SQ{} task instances, each of the form \(\bigl(q, \mathcal{D}^{+}_{q}, \mathcal{D}^{-}_{q}\bigr)\), together with a rationale for the relevance judgments.

\begin{enumerate}[
label=\textbf{\arabic*.}, labelindent=0pt,
leftmargin=0.8em, itemindent=*, labelsep=0.3em,
itemsep=4pt, topsep=2pt, parsep=0pt]

\item \textbf{Author recruitment.}
We recruit participants in two ways.
We invite first authors of oral, spotlight, and award papers at major AI conferences in 2025--2026\footnote{ICLR 2026, ICML 2026, NeurIPS 2025, ACL 2026, CVPR 2026, CoLM 2025, and CoRL 2025.}.
We also reach out personally and issue an open call to first, co-first, and corresponding authors of any paper accepted at a major AI conference since 2025.
More than 100 of the \nauthor{} source papers received an oral, spotlight, or award distinction.
Full details of the author interface and recruitment procedures are provided in Appendix~\ref{app:recruitment}.

\item \textbf{Author validation and revision.}
Through a web interface, authors provide an account of the research underlying their source paper $S$ and assess the relevance of candidate papers, as shown in~\cref{fig:benchmark_construction}.
Starting from the pre-annotation of \cref{sec:bench:pipeline}, they can validate, revise, or rewrite any component to reflect their own account of the research (\cref{tab:query-composition-example} shows one validated query set).
For each query $q$, authors label every candidate from \cref{sec:bench:pipeline} as offering an insightful idea for pursuing $q$ or not, which yields the positives $\mathcal{D}^{+}_{q}$ and hard negatives $\mathcal{D}^{-}_{q}$, and they explain each judgment.
Together, these author-validated task instances constitute the benchmark's relevance supervision.

\item \textbf{Post-processing.}
We manually review all finalized task instances for consistency with the task definition, paying attention to whether queries describe the research question from an early-stage perspective.
Of the 894 queries, 764 (85.5\%) pass this review unchanged.
In the rest, we reword a phrase that gives away the eventual solution or treats the work as complete, keeping the authors' meaning and voice and changing nothing else.
\end{enumerate}

\section{Evaluation and Analysis}
Our evaluation corpus of 191K computer-science papers is far smaller than the three million papers on arXiv~\citep{arxiv2026milestone}, among which researchers are able to find work that inspires a project.
Retrieval in this corpus is therefore an easier version of the problem researchers actually face.
In this section, we ask: can current retrieval systems identify the \textit{catalyst papers} that inspired a project from among the many topically related papers in the corpus?
If not, where do they fail?

\subsection{Experimental Setup}
\label{sec:exp:setup}
We evaluate lexical, multi-vector, dense, and agentic retrieval.
We include BM25~\citep{robertson2009probabilistic} as a lexical baseline, since it performs strongly against dense models on BEIR~\citep{thakur2021beirheterogenousbenchmarkzeroshot} and LongEmbed~\citep{zhu2024longembed}, and LateOn~\citep{sourty2026denseon} as a multi-vector baseline.
We evaluate seven dense retrievers. Qwen3-Embedding-4B and 8B~\citep{zhang2025qwen3} and Gemini Embedding 2~\citep{shanbhogue2026geminiembedding2native} are general-purpose models. SPECTER2~\citep{Singh2022SciRepEvalAM} and OpenScholar~\citep{asai2024openscholar} are trained on scientific documents. ReasonEmbed~\citep{chen2025reasonembed} and Inf-Retriever-v1-Pro~\citep{inf-x-retriever-2025} are optimized for reasoning-intensive retrieval. Appendix~\ref{app:extended_embeddings} provides extended results for further embedding retrievers.

We evaluate three search agents that represent common designs in the literature.
The grep agent follows DCI-Agent-Lite~\citep{li2026beyond} and searches titles and abstracts with regular expressions.
The tool-calling agent follows the search-then-select structure for literature search~\citep{he2025pasa,pan2026paperscout}.
The deep research agent adopts the workflow of deep research systems~\citep{openai2025deepresearch,zhang2025deep}: it breaks the query into subquestions and iterates over search, reading, and synthesis.
The latter two agents retrieve candidates for their search queries with Gemini Embedding 2.
We use GPT-4.1~\citep{openai2026gpt41model} and o3~\citep{openai2026o3model} as backbone models since they are among the strongest models with a knowledge cutoff (June 2024) before all source papers.
To compare against a recent frontier model, we also evaluate Claude Fable 5.1~\citep{anthropic2026fable51}, which may have seen most source papers during training (June 2026 cutoff).
Following previous work~\citep{thakur2021beirheterogenousbenchmarkzeroshot, su2025bright, tchuindjo2026obliq}, we report nDCG@20 and Recall@$k$ ($k=5,20,100$).
Full experimental details are provided in~\cref{app:experiments}.

\begin{mainsettingbox}
Our main setting is retrieval from a \textbf{temporally filtered local corpus}.
We disallow web search.
Agent backbones must have a knowledge cutoff before every source paper (all published 2025 or later).
Models past this cutoff serve only as points of comparison ($\ddagger$ in~\cref{tab:main_results}).
\end{mainsettingbox}

\definecolor{DeltaPink}{HTML}{FF9E98}
\definecolor{DeltaGreen}{HTML}{75D981}

\definecolor{FamilyGray}{RGB}{242,242,242}
\newcommand{\familyrow}[1]{
\multicolumn{9}{>{\columncolor{FamilyGray}[\tabcolsep][\tabcolsep]}c}{
  \textit{\textcolor{black}{#1}}
}\\[1pt]
}

\newcommand{\lightrule}{
\arrayrulecolor{black!30}
\specialrule{\lightrulewidth}{\aboverulesep}{\belowrulesep}
\arrayrulecolor{black}
}
\newcommand{\deltared}[2]{
\cellcolor{DeltaPink!#1!white}
\ensuremath{\downarrow}\,#2
}
\newcommand{\deltaref}{
\cellcolor{DeltaGreen!38!white}---
}
\newcommand{\deltagreen}[2]{
\cellcolor{DeltaGreen!#1!white}
\ensuremath{\uparrow}\,#2
}

\newcommand{\qbelow}[2]{
\cellcolor{DeltaPink!#1!white}#2
}
\newcommand{\qabove}[2]{
\cellcolor{DeltaGreen!\the\numexpr#1+18\relax!white}#2
}
\newcommand{\qref}[1]{
\cellcolor{DeltaGreen!18!white}#1
}

\newcommand{\fablecell}[1]{
\cellcolor{FamilyGray}#1
}

\begin{table*}[!b]
\centering
\small
\setlength{\tabcolsep}{4pt}
\renewcommand{\arraystretch}{1.1}
\vspace{-1mm}
\caption{
\textbf{Retrieval performance on \benchmark{} across both query types.}
nDCG@20 and Recall@$k$ (k=5, 20, 100) on \benchmark{} for both query types.
Bold marks the best per column among the rows under the rule.
Cells are shaded relative to Qwen3-Emb-8B, with green for higher values and red for lower values.
}
\vspace{-1mm}
\label{tab:main_results}
\resizebox{\textwidth}{!}{
\begin{threeparttable}
\begin{tabular}{l *{8}{c}}
\toprule
& \multicolumn{4}{c}{\textcolor{cqcolor}{\optimisticfont Core Research Query} (n=207)}
& \multicolumn{4}{c}{\textcolor{sqcolor}{\optimisticfont Subfield-specific Query} (n=687)} \\
\cmidrule(lr){2-5}\cmidrule(lr){6-9}
Model & nDCG@20 & R@5 & R@20 & R@100 & nDCG@20 & R@5 & R@20 & R@100 \\
\midrule
BM25
& \qbelow{40}{0.16} & \qbelow{35}{0.12} & \qbelow{55}{0.23} & \qbelow{55}{0.39}
& \qbelow{60}{0.21} & \qbelow{55}{0.12} & \qbelow{75}{0.26} & \qbelow{75}{0.45} \\
LateOn 0.1B
& \qbelow{40}{0.16} & \qbelow{35}{0.12} & \qbelow{50}{0.24} & \qbelow{60}{0.37}
& \qbelow{55}{0.27} & \qbelow{35}{0.16} & \qbelow{60}{0.33} & \qbelow{60}{0.55} \\
SPECTER2
& \qbelow{55}{0.13} & \qbelow{45}{0.10} & \qbelow{65}{0.20} & \qbelow{60}{0.38}
& \qbelow{75}{0.18} & \qbelow{60}{0.11} & \qbelow{85}{0.23} & \qbelow{85}{0.42} \\
OpenScholar
& \qbelow{40}{0.16} & \qbelow{30}{0.13} & \qbelow{55}{0.23} & \qbelow{55}{0.40}
& \qbelow{70}{0.19} & \qbelow{55}{0.12} & \qbelow{80}{0.24} & \qbelow{75}{0.46} \\
Qwen3-Emb-4B
& \qbelow{5}{0.25} & \qbelow{12}{0.18} & \qabove{8}{\textbf{0.39}} & \qabove{5}{0.59}
& \qbelow{20}{0.35} & \qbelow{15}{0.21} & \qbelow{25}{0.44} & \qbelow{15}{0.69} \\
Qwen3-Emb-8B
& \qref{0.26} & \qref{\textbf{0.21}} & \qref{0.37} & \qref{0.58}
& \qref{\textbf{0.41}} & \qref{\textbf{0.25}} & \qref{\textbf{0.51}} & \qref{\textbf{0.73}} \\
Gemini-Emb-2
& \qabove{5}{\textbf{0.27}} & \qref{\textbf{0.21}} & \qabove{5}{0.38} & \qabove{18}{\textbf{0.62}}
& \qbelow{10}{0.38} & \qbelow{5}{0.24} & \qbelow{18}{0.46} & \qbelow{10}{0.70} \\
ReasonEmbed
& \qbelow{10}{0.23} & \qbelow{10}{0.18} & \qbelow{10}{0.33} & \qbelow{15}{0.53}
& \qbelow{30}{0.33} & \qbelow{18}{0.19} & \qbelow{35}{0.40} & \qbelow{30}{0.65} \\
Inf-Retriever-v1-Pro
& \qref{0.26} & \qbelow{5}{0.20} & \qref{0.37} & \qbelow{18}{0.53}
& \qbelow{35}{0.31} & \qbelow{20}{0.19} & \qbelow{45}{0.38} & \qbelow{35}{0.63} \\
\midrule
GPT-4.1 + Grep Agent
& \qbelow{80}{0.05} & \qbelow{80}{0.04} & \qbelow{90}{0.06} & \qbelow{8}{0.56$^\dagger$}
& \qbelow{90}{0.09} & \qbelow{85}{0.07} & \qbelow{90}{0.09} & \qbelow{25}{0.66$^\dagger$} \\
GPT-4.1 + Tool-calling Agent
& \qbelow{5}{0.25} & \qbelow{8}{0.19} & \qref{0.37} & \qabove{8}{0.60$^\dagger$}
& \qbelow{12}{0.37} & \qbelow{5}{0.24} & \qbelow{25}{0.43} & \qbelow{10}{0.70$^\dagger$} \\
o3 + Deep research Agent
& \qbelow{8}{0.24} & \qref{\textbf{0.21}} & \qbelow{18}{0.33} & \qabove{5}{0.59$^\dagger$}
& \qbelow{25}{0.33} & \qbelow{12}{0.22} & \qbelow{45}{0.38} & \qbelow{12}{0.69$^\dagger$} \\
\midrule
\multicolumn{9}{l}{\textit{Post-cutoff model (comparison only)}$^\ddagger$} \\[1pt]
Fable 5.1 + Grep Agent
& \fablecell{0.41} & \fablecell{0.36} & \fablecell{0.52} & \fablecell{0.67$^\dagger$}
& \fablecell{0.42} & \fablecell{0.28} & \fablecell{0.48} & \fablecell{0.74$^\dagger$} \\
Fable 5.1 + Tool-calling Agent
& \fablecell{0.36} & \fablecell{0.29} & \fablecell{0.51} & \fablecell{0.68$^\dagger$}
& \fablecell{0.41} & \fablecell{0.27} & \fablecell{0.49} & \fablecell{0.71$^\dagger$} \\
Fable 5.1 + Deep research Agent
& \fablecell{0.40} & \fablecell{0.35} & \fablecell{0.53} & \fablecell{0.69$^\dagger$}
& \fablecell{0.42} & \fablecell{0.28} & \fablecell{0.50} & \fablecell{0.73$^\dagger$} \\
\bottomrule
\end{tabular}
\vspace{1mm}
\begin{tablenotes}[flushleft]
\normalfont\footnotesize
\item[$\dagger$] We backfill agent rankings with papers seen during search and Gemini-Emb-2 results (Appendix~\ref{app:agentic-baselines}).
\item[$\ddagger$] We evaluate Claude Fable 5.1 (June 2026 knowledge cutoff) as a stronger recent model for comparison.
\end{tablenotes}
\end{threeparttable}
}
\end{table*}

\subsection{Main Results}
\label{sec:exp:main-results}

\textbf{All systems struggle to retrieve catalyst papers.}
Current retrievers miss most of the papers that inspired a project (\cref{tab:main_results}).
General-purpose dense retrievers perform best, yet place 39\% of author-identified inspiration papers in the top 20 for \CQ{} and 51\% for \SQ{}.
Scientific-document retrievers (SPECTER2, OpenScholar) trail the general-purpose models by at least 16 points on \CQ{} and 27 points on \SQ{} at Recall@20.
BM25 and LateOn trail by 15--16 points on \CQ{} and 18--25 points on \SQ{}.
Together, these results suggest that domain-specific training on citation signals does not by itself solve this task.
Among the pre-cutoff systems in our main experiment, \CQ{} is harder than \SQ{}, consistent with its longer queries and smaller, almost entirely cited positive sets (\cref{tab:corpus_stats}).
We show later that reranking with an oracle candidate pool recovers part of this gap (\cref{{sec:exp:exp_analysis}}).

\textbf{Agentic search does not improve over embedding retrieval.}
We compare LLM search agents built on pre-cutoff models (GPT-4.1 and o3) against the strongest embedding model (\cref{tab:main_results}).
Grep-based search, despite its recent success as a lightweight alternative to embedding retrieval~\citep{li2026beyond,sen2026grep}, performs substantially worse.
It finds only 8\% of gold query--paper pairs during search, compared with 46\% for the tool-calling agent (\cref{fig:agent_coverage}).

We attribute this gap to \textit{candidate coverage}, since regardless of how a model reasons, an agent can interact with the corpus only through its search queries.
As the corpus scales, this constraint becomes more important because an agent can inspect only a small fraction of the corpus.
Consistent with this account, agent recall rises with backbone capability yet remains close to the embedding retriever's even for the strongest model (\cref{fig:backbone_aggregate}).
If coverage is the limiting factor, progress will require embedding retrievers that surface inspiration papers in the first place and agents that build on those results to guide further exploration.
We report Fable 5.1 separately because it falls outside our knowledge cutoff, and we analyze its results, together with agent failure cases and retrieval rationales, in \cref{app:postcutoff,app:failures}.

\subsection{Analysis of Author-Annotated Catalyst Papers}
\label{sec:exp:inspiration_analysis}
\textbf{Inspiration relations are diverse and not captured by similarity.}
Prompting an LLM judge to classify each of the 663 key-inspiration rationales under a nine-type taxonomy (Appendix~\ref{app:inspiration_type}), we find that authors most often credit a technique they adapted (\textit{method}), empirical findings that supported a direction (\textit{evidence}), an idea extended to a new setting (\textit{generalization}), or a limitation that motivated a new approach (\textit{limitation}), with more than one type applying to 55.8\% of rationales (\cref{fig:inspiration_type}).
Similarity to the query does not identify these papers: measuring lexical similarity as the fraction of unique query words in a candidate's title and abstract, and semantic similarity as cosine similarity under Qwen3-Embedding-8B~\citep{zhang2025qwen3}, hard negatives are at least as similar to the query as positives on both measures (\cref{tab:inspiration_not_topical}).
This also limits agentic search, where grep and embedding tools construct the candidate pool before an agent can reason about which papers could inspire a project.

\textbf{The published record does not reveal which papers inspired a project.}
Source-paper citations provide an incomplete signal of inspiration. 43.6\% of \SQ{} positives are not cited by the source paper, and 67.8\% of those share no references with it.
Among cited papers, citation-intent labels do not identify inspirations: only 46.2\% of \CQ{} and 34.0\% of \SQ{} positives carry a \texttt{uses} or \texttt{extends} label~\citep{lauscher2022multicite}.
Even given the entire finished source paper in addition to its bibliography, LLMs fail to identify catalyst papers.
Gemini 3.1 Pro~\citep{google2026gemini31pro} proposes candidate inspiration papers that cover only 32.2\% of \CQ{} positives on average and none of the key inspirations in 28.0\% of instances.
\Cref{app:benchmark-analysis,tab:citation-status-2x2} give further details, including an instance with both uncited positives and cited hard negatives.

\subsection{Analysis of Factors Behind Retrieval Performance}
\label{sec:exp:exp_analysis}

\begin{figure*}[t]
\centering
\captionsetup{font=footnotesize,skip=2mm}
\vspace{-1mm}

\begin{minipage}[b]{.26\textwidth}
\centering\vspace{0mm}
\includegraphics[width=\linewidth]{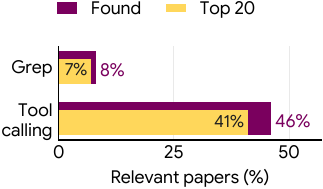}
\caption{\textbf{Grep rarely surfaces gold papers.}
Share of gold papers that each GPT-4.1 agent encounters during search (Found) and returns in its final top 20.}
\label{fig:agent_coverage}
\vspace{0mm}
\end{minipage}\hfill
\begin{minipage}[b]{.32\textwidth}
\centering\vspace{0mm}
\includegraphics[width=\linewidth]{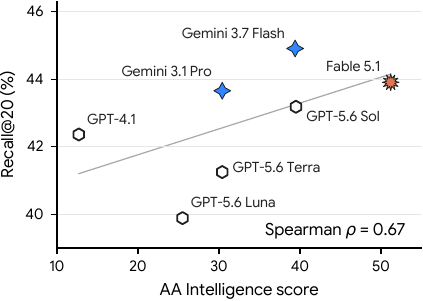}
\caption{\textbf{Agent recall rises with backbone capability.}
Tool-calling-agent Recall@20, plotted against general capability score.}
\label{fig:backbone_aggregate}
\vspace{7mm}
\end{minipage}\hfill
\begin{minipage}[b]{.35\textwidth}
\centering\vspace{-2mm}
\includegraphics[width=\linewidth]{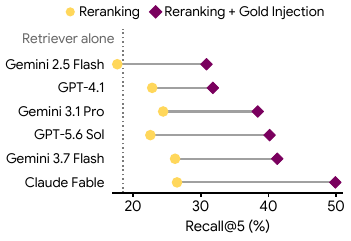}
\caption{\textbf{Stronger rerankers gain more from gold injection.}
Recall@5 after reranking the retriever's top 20, with and without missing gold papers swapped in.}
\label{fig:retriever_verifier_compact}
\vspace{4mm}
\end{minipage}
\vspace{4mm}
\begin{minipage}[t]{.47\textwidth}
\centering\vspace{0pt}
\captionof{table}{\textbf{Retrieval under different source-paper access conditions.}
Recall@20 of the Gemini 3.7 Flash tool-calling agent on the 203-query split.}
\label{tab:access_ablation}
\footnotesize
\setlength{\tabcolsep}{4pt}
\begin{tabular}{@{}lcc@{}}
\toprule
Setting & \CQ{} & \SQ{} \\
\midrule
Agent w/ corpus search & 0.39 & 0.47 \\
\midrule
Agent w/ web search & 0.38 & 0.47 \\
Agent w/ title and abstract & 0.50 & 0.49 \\
Agent w/ source bibliography & 0.74 & 0.57 \\
Closed-book (no corpus) & 0.31 & 0.24 \\
\bottomrule
\end{tabular}
\end{minipage}\hfill
\begin{minipage}[t]{.49\textwidth}
\centering\vspace{0pt}
\captionof{table}{\textbf{Query rewriting does not consistently improve retrieval.} Recall@20 with Gemini-Embedding-2 on 100 queries, with GPT-4.1 rewrites.}
\label{tab:query_gen}
\footnotesize
\setlength{\tabcolsep}{4pt}
\begin{tabular}{@{}lll@{}}
\toprule
Method & \CQ{} & \SQ{} \\
\midrule
Original query & 0.40 & 0.42 \\
\midrule
Single-Query Expansion & 0.44\greendelta{0.04} & 0.37\reddelta{0.05} \\
Multi-Query Generation & 0.37\reddelta{0.03} & 0.41\reddelta{0.01} \\
HyDE (Mean-Pool) & 0.23\reddelta{0.17} & 0.22\reddelta{0.20} \\
HyDE (Multi-Query) & 0.18\reddelta{0.22} & 0.20\reddelta{0.22} \\
\bottomrule
\end{tabular}
\end{minipage}
\vspace{-3mm}
\end{figure*}
\textbf{Stronger rerankers benefit more from better candidate coverage.}
To separate retrieval coverage from recognition performance, we apply LLM-based reranking to the top-$K$ candidates from Gemini-Embedding-2 on a randomly sampled subset of 50 \CQ{} and 50 \SQ{} queries.
We compare reranking with and without gold injection, which replaces negative candidates with positive papers missing from the set while keeping the pool size fixed.
Stronger rerankers gain more in Recall@5 from gold injection (\cref{fig:retriever_verifier_compact}).
The growing gap suggests that candidate coverage limits how much we can benefit from stronger rerankers: their ability to select useful papers matters only when those papers enter the pool.
Better retrieval could therefore yield larger gains as rerankers improve.
Reranking prompts and procedures are provided in~\cref{app:reranking}.

\textbf{Agents benefit from source context but not from fuller reading.}
We vary two kinds of agent context: information about the source paper and how much of each retrieved paper the agent reads.
With a Gemini 3.7 Flash~\citep{google2026gemini37flash} (cutoff March 2026) tool-calling agent on the 203-query cutoff split (\cref{tab:access_ablation}), the source title and abstract raise Recall@20 by 0.11 for \CQ{} and 0.02 for \SQ{}, while web search changes it by at most 0.01.
The source bibliography, which contains 95\% of \CQ{} gold papers, raises it by 0.35.
Without search tools, the model given the title and abstract reaches only 0.31 \CQ{} Recall@20, below corpus search alone, so source context helps by guiding search rather than through memory.
In contrast, letting a GPT-4.1 tool-calling agent read beyond abstracts on 50 queries per type changes Recall@20 by at most 0.04 (\cref{fig:read_policies}), and the best \SQ{} policy still trails the retriever (0.50 versus 0.55).
Together, these results suggest that the bottleneck lies in \textit{knowing what to search for} rather than in understanding the papers already found, consistent with the candidate-coverage explanation above.
Source information is used only in this analysis, and \cref{app:agentic-baselines,app:read-policies} give details.

\textbf{Query rewriting does not consistently improve retrieval.}
We test whether an LLM can improve retrieval by reasoning over the original research question to generate single-query expansions, multiple queries, or hypothetical abstracts~\citep{gao2023precise}.
We evaluate these strategies on 50 queries of each type, using GPT-4.1 for generation and Gemini-Embedding-2 for retrieval.
Neither single-query expansion nor multi-query generation yields a meaningful improvement, while both HyDE variants substantially reduce recall (\cref{tab:query_gen}).
If the retriever's misses were a matter of phrasing, an LLM that reads the research question should be able to recover some of them by rewording.
That none of these strategies does so suggests the embedding space itself does not place inspiration papers near their queries, consistent with the similarity analysis in~\cref{sec:exp:inspiration_analysis}.
We show prompts and implementation details in~\cref{app:query-augmentation}.

\section{Conclusion}
\label{sec:conclusion}

We introduced \benchmark{}, a benchmark for one ingredient of research ability: identifying prior papers whose ideas could help a project move forward.
Drawing on \nparticipant{} researchers' own accounts of what shaped their work, we capture judgments that the published record largely omits.
Current systems recover only a fraction of the papers that authors credit, and agentic search does no better than embedding retrieval, largely because limited candidate coverage keeps many useful papers out of reach.

\textbf{Limitations.}
\benchmark{} covers computer science papers from 2025--2026 with uneven representation across research areas (Table~\ref{tab:author_set_domain}), so our findings may not transfer to other fields.
Our labels are retrospective judgments by paper authors and are thus subject to hindsight.
A more fundamental open question is the skyline, i.e., the \textbf{performance ceiling} of this task.
As a point of reference, the human reference accuracy on ImageNet (94.9\% top-5~\citep{russakovsky2015imagenetlargescalevisual}) was later exceeded, and reannotation studies found that subsequent gains increasingly reflected the idiosyncrasies of the original labels~\citep{beyer2020imagenet,pmlr-v119-shankar20c}.
Catalyst paper retrieval is harder to audit this way, since relabeling by anyone other than the authors would be costly and would measure a different target, namely what an informed outsider can infer.
Still, a preliminary comparison on one paper with three author-annotators in \benchmark{} found that coauthors marking only about seven positives per thread recovered 43--60\% of the first author's labels at 84--88\% precision.
This is well above the best system's overall R@5 of 0.24, which suggests considerable room for improvement.

\textbf{Outlook.}
Recent work on \textbf{reasoning-intensive retrieval} evaluates relevance relations that require reasoning to recognize~\citep{xiao2024rar,su2025bright,tchuindjo2026obliq}.
\benchmark{} pushes further, requiring judgments of how prior ideas could advance a research project.
We see two milestones ahead.
The first is \textit{expert-level} retrieval, matching a domain expert's judgment within their own field.
The second is retrieval \textit{beyond any individual expert}: being able to exercise that level of intuition across a breadth of fields no single researcher can follow.
This breadth may be where such systems help most, since nearly half of the papers our authors credit (49.5\%) come from outside their project's subject area, and 43.6\% of subfield-specific positives were never cited in the source paper.
Researchers may therefore be limited more by what they have seen than by how well they judge it.
The full literature makes this gap far larger than our benchmark shows: Semantic Scholar alone indexes over 225M papers~\citep{kinney2023semantic}, compared with 190k in our corpus.
What stands between current systems and both milestones is \textit{an expert-level sense of which ideas matter}, and we hope \benchmark{} offers a starting point for developing it.
If \textbf{undiscovered public knowledge}~\citep{swanson1986undiscovered} is as common as our results imply, models capable of expert judgment across fields could unlock a large backlog of progress hidden in plain sight.

\clearpage
\section*{Acknowledgements}
The \textsc{ScholarCatalyst} benchmark would not have been possible without the researchers who shared how their discoveries came about (see~\cref{app:contributors}), and we are deeply grateful for their contributions.
We thank Kangwook Lee, Anikait Singh, Joy He-Yueya, Braden Hancock, and Vincent Chen for helpful feedback at different stages of this project.
YL thanks Heewon Son for her support throughout this work.
This work was supported by the Laude Slingshot program, OpenAI, KFAS, Schmidt Sciences AI2050, Stanford HAI, and Hyundai Motor Chung Mong-Koo Foundation.

\bibliography{iclr2026_conference}
\bibliographystyle{iclr2026_conference}

\newpage
\appendix

\clearpage
\etocdepthtag.toc{appendixpart}
\section*{\scshape Appendix}
\vspace{1.25em}
\begingroup
\hypersetup{linkcolor=black}
\etocsettagdepth{mainpart}{none}
\etocsettagdepth{appendixpart}{subsubsection}
\etocsettocstyle{}{}
\setlength{\parskip}{0pt}
\newcommand{\apptocline}[3]{
  \par\noindent\hangindent=\dimexpr#1+#2\relax\hangafter=1
  \rightskip=2.5em plus 1fil\parfillskip=-\rightskip
  \hspace*{#1}\makebox[#2][l]{\etocnumber}#3\nobreak\leaders\hbox{.\hskip 0.5em}\hfill\makebox[2em][r]{\etocpage}\par}
\etocsetstyle{section}{}{}
  {\addvspace{0.8em}\apptocline{0pt}{1.6em}{\textbf{\etocname}}\nopagebreak\vspace{0.5em}}{}
\etocsetstyle{subsection}{}{}
  {\apptocline{2.2em}{2.4em}{\etocname}\vspace{0.5em}}{}
\etocsetstyle{subsubsection}{}{}
  {\apptocline{4.6em}{3.2em}{\etocname}}{}
\tableofcontents
\endgroup
\clearpage

\section{Benchmark Contributors}
\label{app:contributors}

We are deeply grateful to the researchers who contributed to \benchmark{} by reviewing our reconstruction of their own work, revising the research questions and inspiration papers, and sharing the reasoning behind which prior work genuinely shaped their research. 
Their firsthand knowledge is the foundation of this benchmark, and we thank them for generously dedicating their time and expertise.
Contributors who agreed to be acknowledged are listed below in alphabetical order.

\begin{tcolorbox}[
    colback=black!2,
    colframe=white,
    enhanced,
    frame hidden,
    boxrule=0pt,
    arc=0pt,
    left=9pt,
    right=9pt,
    top=23pt,
    bottom=23pt,
    before skip=22pt,
    after skip=0pt,
    overlay unbroken={
        \draw[line width=0.5pt] (frame.north west) -- (frame.north east);
        \draw[line width=0.25pt] ([yshift=-2.5pt]frame.north west) -- ([yshift=-2.5pt]frame.north east);
        \draw[line width=0.5pt] (frame.south west) -- (frame.south east);
        \draw[line width=0.25pt] ([yshift=2.5pt]frame.south west) -- ([yshift=2.5pt]frame.south east);
    },
]
\footnotesize
\renewcommand{\baselinestretch}{1.36}\selectfont
\noindent

Adir Dayan, Ajay Sridhar, Alberta Longhini, Alberto Rota, Alexandru Oarga, Ali Hatamizadeh, Alireza Mousavi-Hosseini, Angana Borah, Anjiang Wei, Atharva Kulkarni, Bhuvan Sachdeva, Bo Liu, Bryan Sangwoo Kim, Burouj Armgaan, Camila Blank, Caroline Choi, Changho Shin, Changyeon Kim, Chaofan Lin, Chengyang He, Chuhan Li, David Anugraha, Dayoon Ko, Dilara Soylu, Dongjin Kang, Dongyang Fan, Ekaterina Kochetkova, Eunkyu Park, Fan Nie, Fanhu Zeng, Fatemeh Pesaran Zadeh, Francesco De Santis, Gaotang Li, Giung Nam, Guancheng Zhou, Guangnian Wan, Guijin Son, Gul Sena Altıntas, Gyuhyeon Seo, Han Luo, Hanane Moussa, Hangoo Kang, Haochen Tian, Haoyu Zhang, Haozhe Chen, Haozhe Wang, Himanshu Gupta, Ho Sy Tuyen, Hoang Pham, Houjun Liu, Hwan Chang, Hyeong Kyu Choi, Hyunji Lee, Hyunjun Lee, Ifdita Hasan Orney, Ilia Sucholutsky, Insu Lee, Itay Itzhak, Jaewoo Ahn, Jarek Liesen, Jason S Lucas, Jen-Yen Chang, Jiaju Ma, Jiaqi Liu, Jiayin Zhu, Jingzhe Shi, Jinheon Baek, Jinkun Hao, Jinwoo Lim, Jiwoo Chung, Johannes Kirmayr, Johnny Tian-Zheng Wei, Jongseo Lee, Joonghyuk Shin, Jubayer Ibn Hamid, Junchi Yu, Jungwoo Kim, Junyoung Lim, Kallol Saha, Kevin Qinghong Lin, Kodai Kawamura, Korneel Van den Berghe, Kunwoong Kim, Letian Fu, Linus Kreitner, Lucy Xiaoyang Shi, Lynnette Hui Xian Ng, Madhav Kanda, Mahyar Ghazanfari, Marcel Torne, Marco Bertolini, Marco Fumero, Maria Parelli, Marie Brockschmidt, Markus Gross, Mert Cemri, Michael Rizvi-Martel, Ming Hu, Mingtong Zhang, Mingyu Kim, Minki Kang, Minkyu Kim, Minseo Kim, Na Min An, Nathaniel L. Diamant, Niels Mündler, Nilesh Jain, Nilesh Prasad Pandey, Omin Kwon, Peng Xia, Pengcheng Wang, Pratik Sachdeva, Qingyang Zhang, Qizheng Zhang, Rahul Sharma, Raj Ghugare, Richard Cornelius Suwandi, Rohit Jena, Rosen Ting-Ying Yu, Roshen Sanjay Nair, Roussel Desmond Nzoyem, Rulin Shao, Runhan Huang, Sanghyun Jo, Seohong Park, Seokwon Song, Seonghyeon Ye, Seungone Kim, Shayan Talaei, Shihao Wang, Shiyin Jiang, Sijia Liu, Sijie Zhao, Siqi Zhu, Taehoon Yoon, Tanvir Ahmed Sijan, Tianhe Wu, Tongtian Zhu, Truong Buu Phan, Valentin Hofmann, Valeria Ruscio, Vasily Ilin, Wei Chen, Weiheng Liu, Wenbin Ouyang, Wenjie Zhu, Woomin Song, Woongyeong Yeo, Xiang Li, Xiaotian Liu, Xuan Luo, Xuanming Zhang, Xuying Ning, Yancheng Zhang, Yang Li, Yangcen Liu, Yanggan Gu, Yangtian Zhang, Yanjiang Guo, Yaozhong Shi, Yasaman Haghighi, Yichao Cai, Yifan Zhang, Yoonho Lee, Yoonkyo Jung, Yoonsang Lee, Yuante Li, Yuejiang Liu, Yujia Zheng, Yumin Choi, Yusuf Dalva, Zeyuan Chen, Zhaoyang Liu, Zheng Huang, Zhenhao Chen, Zhongxing Xu, Zijian Feng, Zixuan Hu
\end{tcolorbox}
\clearpage

\section{The \benchmark{} Benchmark}
\label{app:benchmark}

\subsection{Benchmark Instance}
\label{app:benchmark-instance}

\benchmark{} evaluates whether a retrieval system can find prior work whose ideas would have been insightful for pursuing a research project, rather than papers that are merely topically similar. Each instance is anchored in a completed source paper and pairs its central research question and distinct research directions with author-validated positive papers, topically related hard negatives, and a temporally restricted search corpus. The 207 author-validated instances contribute 894 queries: 207 core research queries and 687 subfield-specific queries. This appendix first defines the information in one benchmark instance (\cref{app:benchmark-instance}), then reports the composition of the instances and corpus (\cref{app:dataset-stats}), and finally examines the kinds of inspiration relations the benchmark captures (\cref{app:benchmark-analysis}).

\begin{table}[h]
\centering
\footnotesize
\setlength{\tabcolsep}{6pt}
\renewcommand{\arraystretch}{1.15}
\caption{Components of a \benchmark{} instance.}
\label{tab:benchmark_instance_structure}
\begin{tabular}{l p{0.62\linewidth}}
\toprule
\textbf{Component} & \textbf{Information} \\
\midrule
Source paper ($S$) & The completed paper defining the instance. Withheld from the retriever, so its solution and bibliography cannot be used to find prior work. \\
Search corpus ($\mathcal{P}_{<S}$) & Papers published before $S$ (temporal cutoff). \\
Core research query \CQ{} & Asks which prior ideas could help address the paper's central research question. \\
Subfield-specific query \SQ{} & Asks which prior ideas from one specific research direction could help advance the paper. \\
Positive papers ($\mathcal{D}^{+}_{q}$) & Papers in $\mathcal{P}_{<S}$ that authors judge to offer an insight that could inspire work on $q$. \\
Hard negatives ($\mathcal{D}^{-}_{q}$) & Papers in $\mathcal{P}_{<S} \setminus \mathcal{D}^{+}_{q}$ that are topically related to $q$ but judged not to offer such an idea. \\
Rationale & Every query carries a rationale for its positive set.\\
\bottomrule
\end{tabular}
\end{table}

A project's overarching question and its subfield-specific questions need not draw on the same prior work.
The \textit{core research query} can identify ideas that organize the project as a whole, whereas \textit{subfield-specific queries} surface papers relevant to particular methodological or empirical directions.
Their overlap and divergence reveal complementary forms of inspiration within a single project.
Table~\ref{tab:query-composition-example} shows what this looks like for one full instance.

\begin{table}[h]
\centering
\small
\renewcommand{\arraystretch}{1.08}
\begin{tabular}{@{}p{0.96\linewidth}@{}}
\toprule

\textbf{Source Paper}\\[-0.2em]
\textbf{Hi Robot: Open-Ended Instruction Following with Hierarchical Vision-Language-Action Models (\texttt{cs.RO})} \\
\midrule

\textbf{Core Research Query}\\[-0.2em]
{\ttfamily
Robots working in human-centric settings receive language that is far messier than the atomic commands studied in language-conditioned imitation learning. A line of work trains policies to map instructions such as 'pick up the coke can' to actions and generalizes well within that regime, but real requests carry constraints, preferences, and mid-execution corrections ("that's not trash", "I'm allergic to pickles") whose meaning depends on what the robot is currently seeing and doing. A separate line of work uses LLMs and VLMs off the shelf to decompose long-horizon tasks over predefined or hand-designed skills. These systems parse richer language but inherit the limited dexterity of their skill libraries and rarely react to a user who interrupts mid-task. Both lines leave a gap between the semantics a system can interpret and the physical behavior it can actually produce, and it is not obvious whether that gap is best closed by making language interpretation more capable or by making dexterous learned policies more promptable. \textcolor{benchmarkplum}{How can a robot with dexterous, learned manipulation skills interpret open-ended prompts and real-time human feedback, grounded in its own observations, and act on them while the task is underway?}} \\
\midrule

\textbf{Subfield-Specific Query 1}\\[-0.2em]
{\ttfamily
Vision-language-action models fine-tune pretrained VLMs to emit low-level continuous actions conditioned on images and a language instruction, and have shown strong dexterity and some language generalization on real hardware. However, their instruction supervision comes from demonstration annotations that are short and literal ("put the cup on the plate"), so it is unclear how much of their language-following ability survives when a prompt carries constraints, negations, preferences, or a correction issued partway through a rollout. \textcolor{benchmarkplum}{What language can an end-to-end VLA actually be steered by at execution time, and where does its instruction-following break down as prompts depart from the literal commands seen in its training data?}} \\
\midrule

\textbf{Subfield-Specific Query 2}\\[-0.2em]
{\ttfamily
A line of work uses pretrained LLMs and VLMs to reason over a scene and break a long-horizon goal into a sequence of calls to predefined or hand-designed skills. These models bring broad semantic and world knowledge, enough to infer that a paper cup is trash while a reusable one is a dish, or that "something sweet" means candy, but they have never seen the robot's observations, its skill repertoire, or what its execution actually looks like mid-episode. \textcolor{benchmarkplum}{What limits an off-the-shelf foundation model's ability to serve as a robot's high-level reasoner: can it stay grounded in the robot's own visual stream and affordances across a long rollout, and does its output remain something the low-level controller can physically carry out?}} \\
\midrule

\textbf{Subfield-Specific Query 3}\\[-0.2em]
{\ttfamily
A line of work already lets users correct a robot while it acts. Some approaches edit trajectories from verbal feedback, others train recovery behaviors from simulator-constructed perturbations, and others route spoken corrections through a language channel between a high-level and a low-level policy. Each demonstrates that mid-execution language can redirect behavior, but each is bounded: by a single task prompt, by the specific correction phrases human annotators happened to produce, or by feedback that never depends on what the robot is currently seeing. Utterances like "that's not trash" or "leave it alone" are unresolvable from text alone. They refer to whatever is in the gripper right now. \textcolor{benchmarkplum}{What determines the range of corrections such a system can absorb, and can a robot ground feedback in its live observations while also handling phrasings and situations that never appeared in its interaction data?}} \\
\bottomrule
\end{tabular}

\caption{\textbf{Example query set for one research project.}
The core research query describes the project's overarching goal, while subfield-specific queries isolate the distinct directions through which it is pursued.}
\label{tab:query-composition-example}
\end{table}

\subsection{Dataset Statistics}
\label{app:dataset-stats}
The author-validated benchmark contains 207 source papers and 894 queries concentrated in machine learning and adjacent AI fields.
Table~\ref{tab:author_set_domain} reports each source paper's primary arXiv category,\footnote{arXiv category taxonomy: \url{https://arxiv.org/category_taxonomy}} where \emph{Other} comprises stat.ML, cs.SE, cs.IR, cs.MA, cs.HC, cond-mat.mtrl-sci, cs.CE, q-bio.NC, cs.IT, cs.GR, and cs.NE.
The author-validated set comprises 894 queries (207 central, 687 directional).
Table~\ref{tab:corpus_stats} summarizes the query and corpus statistics.

\begin{table}[h]
\centering
\caption{Primary arXiv category of the author-validated instances.}
\label{tab:author_set_domain}
\begin{tabular}{lrr}
\toprule
Primary category & Count & \% \\
\midrule
cs.LG (Machine Learning) & 66 & 31.9\% \\
cs.CV (Computer Vision) & 46 & 22.2\% \\
cs.CL (Computation and Language) & 38 & 18.4\% \\
cs.AI (Artificial Intelligence) & 23 & 11.1\% \\
cs.RO (Robotics) & 20 & 9.7\% \\
Other & 14 & 6.8\% \\
\bottomrule
\end{tabular}
\end{table}

\begin{table}[t]
\centering
\begin{minipage}[t]{0.60\linewidth}
\centering
\captionof{table}{Primary arXiv category of the 190{,}896 corpus papers.}
\label{tab:corpus_domain}
\begin{tabular}{lrr}
\toprule
Primary category & Count & \% \\
\midrule
cs.CV (Computer Vision) & 64{,}919 & 34.0\% \\
cs.LG (Machine Learning) & 57{,}151 & 29.9\% \\
cs.CL (Computation and Language) & 33{,}255 & 17.4\% \\
cs.RO (Robotics) & 18{,}922 & 9.9\% \\
cs.AI (Artificial Intelligence) & 10{,}933 & 5.7\% \\
Other arXiv category & 4{,}116 & 2.2\% \\
Non-arXiv (S2, OA) & 1{,}600 & 0.8\% \\
\bottomrule
\end{tabular}
\end{minipage}\hfill
\begin{minipage}[t]{0.34\linewidth}
\centering
\captionof{table}{Publication year of the 190{,}896 corpus papers.}
\label{tab:corpus_year}
\begin{tabular}{lrr}
\toprule
Year & Count & \% \\
\midrule
Before 2020 & 1{,}783 & 0.9\% \\
2020 & 28{,}570 & 15.0\% \\
2021 & 34{,}894 & 18.3\% \\
2022 & 33{,}956 & 17.8\% \\
2023 & 42{,}091 & 22.0\% \\
2024 & 47{,}029 & 24.6\% \\
2025 or later & 2{,}573 & 1.3\% \\
\bottomrule
\end{tabular}
\end{minipage}
\end{table}

The retrieval corpus contains 190{,}896 papers: resolved references cited by our source papers, together with a pool of additional arXiv papers drawn from five main computer science domains and published between 2020 and 2024. Table~\ref{tab:corpus_domain} reports the primary arXiv category of every corpus paper, and Table~\ref{tab:corpus_year} reports its publication year.

Of the 190{,}896 corpus papers, 189{,}296 (99.2\%) are sourced from arXiv\footnote{\url{https://arxiv.org}} directly. The remaining 1{,}600 papers (0.8\%) come from Semantic Scholar (S2)\footnote{\url{https://www.semanticscholar.org/product/api}} and OpenAlex (OA)~\citep{priem2022openalex}, used only when a cited paper could not be found on arXiv itself.
Because the corpus includes every resolved reference our source papers cite, it also ends up containing a small number of papers published in 2025 or later, past the nominal collection window between 2020 and 2024. This does not cause temporal leakage: at evaluation time, we restrict each query's search to only the papers published no later than that query's own source paper.

\subsection{Qualitative Analysis of Inspiration Relations}
\label{app:benchmark-analysis}
This section provides further detail on the analysis of author-identified inspiration in \cref{sec:exp:inspiration_analysis} and presents additional qualitative examples.
To examine the inspirations captured by \benchmark{} beyond topical relevance,
we qualitatively analyze author-provided rationales for catalyst papers.
We first introduce a multi-label taxonomy of inspiration relations, and then
present case studies illustrating cross-subfield transfer, the context-dependent
role of prior work, and the imperfect correspondence between citation and
inspiration.
The authors' accounts offer a rare view of the research process that finished papers seldom document: which prior ideas researchers found useful, and why.
We find that these judgments are diverse in kind and unrelated to similarity, and that neither citations nor an LLM reading the finished paper can recover them.

\begin{table}[h]
\centering
\footnotesize
\setlength{\tabcolsep}{3pt}
\caption{\textbf{Inspiration papers are not more similar to queries than hard negatives.}
Mean BM25 lexical similarity and Qwen-3-8B semantic similarity between each query and its positive papers ($\mathcal{D}_q^+$) or topically similar hard negatives ($\mathcal{D}_q^-$).}
\label{tab:inspiration_not_topical}
\begin{tabular}{@{}lcccc@{}}
\toprule
& \multicolumn{2}{c}{BM25 Sim.} & \multicolumn{2}{c}{Emb. Sim.} \\
\cmidrule(lr){2-3}\cmidrule(lr){4-5}
Query & $\mathcal{D}_q^+$ & $\mathcal{D}_q^-$ & $\mathcal{D}_q^+$ & $\mathcal{D}_q^-$ \\
\midrule
\CQ{}    & 28.7 & 29.2 & 64.5 & 66.2 \\
\SQ{} & 34.5 & 34.4 & 65.1 & 64.2 \\
\bottomrule
\end{tabular}
\end{table}

\begin{figure}[h!]
\centering
\includegraphics[width=0.5\linewidth]{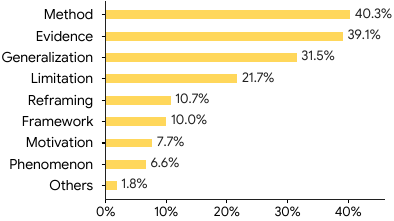}
\caption{\textbf{Distribution of inspiration types.} Share of key-inspiration rationales labeled with each type. Percentages sum to more than 100\% because a rationale can receive multiple types.}
\label{fig:inspiration_type}
\end{figure}

\textbf{Inspiration relations are diverse and not captured by similarity.}
Prompting an LLM judge to classify each of the 663 key-inspiration rationales under a nine-type taxonomy, we find that authors most often credit a technique they adapted (\textit{method}), empirical findings that supported a direction (\textit{evidence}), an idea extended to a new setting (\textit{generalization}), or a limitation that motivated a new approach (\textit{limitation}), with more than one type applying to 55.8\% of rationales (\cref{fig:inspiration_type}).
Similarity to the query does not identify these papers: measuring lexical similarity as the fraction of unique query words in a candidate's title and abstract, and semantic similarity as cosine similarity under Qwen3-Embedding-8B~\citep{zhang2025qwen3}, hard negatives are at least as similar to the query as positives on both measures (\cref{tab:inspiration_not_topical}).
This also limits agentic search, where grep and embedding tools construct the candidate pool before an agent can reason about which papers could inspire a project.

\noindent\textbf{Citation Status Is an Imperfect Proxy for Inspiration.}
Citation status and author-identified inspiration can diverge in both directions.
\Cref{tab:citation-status-2x2} illustrates one source-paper instance in which two of four \CQ{} positives are uncited, while two cited papers are hard negatives.
The uncited positives informed the source paper's exploration and offline-learning mechanisms, whereas the cited hard negatives are topically adjacent but did not shape its approach.
This pattern also appears at scale: 43.6\% of \SQ{} positives are not cited by the source paper, and 67.8\% of those share no references with it.
Even among cited papers, citation intent does not single out inspirations: only 46.2\% of \CQ{} and 34.0\% of \SQ{} positives carry a \texttt{uses} or \texttt{extends} label~\citep{lauscher2022multicite}.

\begin{table}[t]
\centering
\small
\renewcommand{\arraystretch}{1.2}
\begin{tabular}{@{}p{0.1\linewidth} p{0.4\linewidth} p{0.4\linewidth}@{}}
\toprule
\multicolumn{3}{@{}p{0.9\linewidth}@{}}{
\textbf{\CQ{}.} {\ttfamily How can RL agents leverage the temporal dynamics of spiking neural networks when randomly initialized networks fail to collect sequences long enough to expose those dynamics?}
} \\[2pt]
\midrule
& \textbf{\CQ{} Positive} & \textbf{\CQ{} Hard Negative} \\
\midrule
\textbf{Cited} &
Elucidating the theoretical underpinnings of surrogate gradient learning in SNNs;
Neuromorphic Attitude Estimation and Control &
Evolving Connectivity for Recurrent Spiking Neural Networks;
Deep RL with Spiking Q-learning \\
\midrule
\textbf{Uncited} &
Jump-Start Reinforcement Learning;
Revisiting the Minimalist Approach to Offline RL &
8 further topically adjacent papers, including surrogate-gradient variants and neuromorphic control work \\
\midrule
\multicolumn{3}{@{}p{\linewidth}@{}}{
\textbf{Author rationale for uncited positives.}
{\ttfamily Jump-Start Reinforcement Learning suggests using a secondary controller to bridge the warmup period, enabling the spiking network to learn from the start. Revisiting the Minimalist Approach to Offline RL supports combining secondary-controller demonstrations with the spiking actor's own TD3 rollouts.}
} \\
\bottomrule
\end{tabular}
\caption{\textbf{Citation Status Does Not Identify Inspiration.}
For a \CQ{} about leveraging temporal SNN dynamics in RL, citation status is crossed with the author-provided positive/hard-negative judgment. The highlighted rationale explains why the uncited positives were useful.}
\label{tab:citation-status-2x2}
\end{table}

\medskip
\noindent\textbf{Diversity of Inspiration Relations.}
\label{app:inspiration_type}
Prior work inspires new research through diverse mechanisms, including direct method transfer, generalization to a new setting, limitations of existing work, and conceptual reframing.
To characterize \emph{how} a key inspiration paper shaped a target work, we define a nine-type taxonomy, given in~\cref{tab:inspiration_type_examples}, and prompt an LLM judge to label each core-research-query key-inspiration rationale with every type it clearly expresses.
To capture the ways a single paper can shape a project, we allow each rationale to receive multiple type labels. The rationales receive 1.69 labels on average, with 55.8\% receiving two or more and 12.7\% receiving three or more.
This pattern shows that inspiration often operates through several mechanisms at once.

\begin{table}[h!]
\centering
\small
\renewcommand{\arraystretch}{1.15}
\caption{\textbf{Taxonomy of inspiration types.} Definitions describe how a source work shaped the target work. }
\label{tab:inspiration_type_examples}
\begin{tabular}{p{0.17\linewidth} p{0.72\linewidth}}
\toprule
\textbf{Inspiration type} & \textbf{Definition} \\
\midrule
\textbf{\optimisticfont Method} &
A specific method, mechanism, component, formulation, or principle from the source work is adopted or adapted. \\

\textbf{\optimisticfont Generalization} &
An idea, finding, or method from the source work is extended to a new domain, modality, task, setting, or broader scope. \\

\textbf{\optimisticfont Limitation} &
A limitation, failure mode, or unresolved issue in the source work motivates the target work to address it. \\

\textbf{\optimisticfont Framework} &
The target work builds on the source work's overall framework, model, or system rather than a specific component. \\

\textbf{\optimisticfont Phenomenon} &
An empirical phenomenon or observation in the source work motivates the target work to explain, characterize, or formalize it. \\

\textbf{\optimisticfont Reframing} &
An insight from the source work provides a new perspective, abstraction, or interpretation of the target problem. \\

\textbf{\optimisticfont Motivation} &
The source work motivates the research problem or direction pursued in the target work. This label is used only when the motivation is not better captured by limitation, phenomenon, or evidence. \\

\textbf{\optimisticfont Evidence} &
The source work provides evidence or an independent finding that supports the plausibility or direction of the target work without being directly transferred. \\
\bottomrule
\end{tabular}
\end{table}

\begin{table}[h]
\centering
\small
\renewcommand{\arraystretch}{1.3}
\begin{tabular}{p{0.96\linewidth}}
\toprule
\rowcolor[HTML]{F1F1F1}
\textbf{Source Paper}
Towards Understanding the Mechanisms of Classifier-Free Guidance (\texttt{cs.CV}) (NeurIPS 2025 Spotlight) \\[0.9em]
\midrule
\textbf{Core Research Question}\newline
{\ttfamily
Classifier-free guidance (CFG) is a highly effective inference-time guidance method that significantly improves sample quality and condition alignment in diffusion models, but its underlying mechanisms remain poorly understood. Existing theoretical analyses mainly study isotropic Gaussian mixtures or one-dimensional distributions, where much of the complex structure of natural images is absent. This made us wonder (i) whether we could construct a simplified model in which the effects of CFG on image structures become directly visible, hence capturing the underlying mechanism of CFG more faithfully than the simplified settings considered in prior theoretical analyses and (ii) whether the mechanisms identified in this simplified setting extend beyond itself and can help explain, at least partially explain the CFG's mechanism in real-world diffusion models?}\\[1.4em]
\midrule
\rowcolor[HTML]{FFF4CF}
\textbf{Key Inspiration Paper:} Abid, Zhang, Bagaria \& Zou (2018). Exploring patterns enriched in a dataset with contrastive principal component analysis. \emph{Nature Communications} 9, 2134. \\[1.4em]
\midrule
\textbf{Rationale}\newline \textbf{Contribution:} {\ttfamily
This paper shows that the eigendirections associated with differences between the covariance structures of two datasets can reveal features that are distinctive to one dataset relative to the other. These directions, referred to as contrastive principal components, characterize prominent structures that distinguish the target data from the background data.}\\

\textbf{New idea I brought into my own research:} {\ttfamily When analyzing CFG in our simplified linear model, I found that the guidance term involves the difference between two covariance matrices: the covariance of the conditional distribution and that of the unconditional distribution (CPC guidance component in equation (12) of my paper). This suggested that the effect of CFG might be understood through the structure of this covariance difference. I then searched for prior work studying such differences between covariance structures and came across contrastive PCA. This connection suggested that the relevant eigendirections could identify structures that are particularly prominent in the conditional distribution relative to the unconditional distribution, which we later show contribute to the CFG's effect.}\\

\textbf{Why I expect that idea to work:}
{\ttfamily The mathematical structure arising in our analysis of CFG closely resembles that underlying contrastive PCA: both are governed by differences between the covariance structures of two distributions. This similarity suggested that the interpretation of contrastive principal components could provide useful insight into how CFG emphasizes structures that distinguish the conditional distribution from the unconditional one.} \\[0.6em]
\bottomrule
\end{tabular}
\caption{\textbf{Inspiration can cross research domains.}
In the above example, contrastive PCA supplies the covariance-difference perspective used to analyze classifier-free guidance.}
\label{tab:review_2505.19210_abid2018exploring}
\end{table}

\medskip
\noindent\textbf{Context-Dependent Roles of the Same Source Paper.}
A prior paper does not have a fixed role across research projects.
It may offer a transferable method for one target, motivate a different line of inquiry for another, and provide little value for a third despite topical overlap.
Its usefulness depends on the question a project is trying to answer.
Table~\ref{tab:context-dependent-self-consistency} shows one key catalyst paper, selected as a \CQ{} positive by three unrelated source papers, each applying it in a different way.

\begin{table}[h]
\centering
\small
\renewcommand{\arraystretch}{1.3}
\begin{tabular}{p{0.96\linewidth}}
\toprule
\rowcolor[HTML]{FFF4CF} \textbf{Key Inspiration Paper:} Self-Consistency Improves Chain of Thought Reasoning in Language Models (\texttt{cs.CL}, \texttt{arxiv:2203.11171}) \\[0.6em]
\midrule
\rowcolor[HTML]{F1F1F1}
\textbf{Source Paper:} Benefits and Limitations of Communication in Multi-Agent Reasoning (\texttt{cs.MA}, \texttt{arxiv:2510.13903}) \\[0.6em]
\textbf{Role: naive baseline.}\newline
{\ttfamily Proposes multiple CoT with a voting mechanism at the end to improve performance. This work seemed to be the perfect naive baseline as the simplest "multi-agent" protocol where multiple agents think and do not communicate during their thinking process. Here all agents have all the context and think the full CoT, in contrast with multi-agent approaches where context is chunked and CoT is distributed. We used this as a baseline in our empirical experiments section. We expected this to be a good naive baseline as it does not adaptively allocate resources and models are given full context. These were the two main levers by which we conceived multi-agent reasoning to be superior.}\\[0.6em]
\midrule
\rowcolor[HTML]{F1F1F1}
\textbf{Source Paper}\newline
Self-Consistency for LLM-Based Motion Trajectory Generation and Verification (\texttt{cs.CV}, \texttt{arxiv:2603.29301}) \\[0.6em]
\textbf{Role: generalization to a new modality.}\newline
{\ttfamily The key work that our approach extends from, we bring it to the visual domain, whereas this work only works for language-based tasks.}\\[0.6em]
\midrule
\rowcolor[HTML]{F1F1F1}
\textbf{Source Paper}\newline Accelerated Test-Time Scaling with Model-Free Speculative Sampling (\texttt{cs.CL}, \texttt{arxiv:2506.04708}) \\[0.6em]
\textbf{Role: structural premise to exploit.}\newline
{\ttfamily This paper proposes parallel test-time scaling, which is one of the core bases of our work. We aim to accelerate test-time scaling, along the setting that the given paper suggested. We expected the idea to work because speculative decoding is already known to accelerate decoding, and we found redundancy to leverage.}\\
\bottomrule
\end{tabular}
\caption{\textbf{A single paper can inspire different projects in different ways.}
In the above example, Self-Consistency serves as a baseline, a cross-modal generalization, and a systems premise across three source papers.}
\label{tab:context-dependent-self-consistency}
\end{table}

\section{Experimental Setup}
\label{app:experiments}
This appendix specifies the model configurations, agentic baselines, and evaluation conditions used in our experiments. Supplementary analyses are in \cref{app:experimental-details}, and all prompts are in \cref{app:prompts}.

\subsection{Model Configurations}
\label{app:model-config}

Table~\ref{tab:retriever_configs} summarizes the retrieval models used in \benchmark{}, including their architectures, input limits, and whether they receive a task-specific instruction. Resource links point to the corresponding checkpoint, API, or implementation. Table~\ref{tab:llm_backbones} lists the language-model backbones used for agentic search and LLM reranking. GPT-4.1 and o3 run under the knowledge-cutoff rule of Table~\ref{tab:main_results}; the other backbones serve the reranking, backbone-swap, and leakage analyses, and Claude Fable 5.1 also appears in the rows of Table~\ref{tab:main_results} marked $\ddagger$.

\newtcolorbox{inlineprompt}{
  blanker, breakable,
  borderline west={1.1pt}{0pt}{benchmarkplum},
  left=6pt, right=1em,
  before skip=8pt, after skip=4pt,
  fontupper=\fontfamily{cmtt}\selectfont\small
}
For Qwen3-Embedding models, we prepend each query with the following task instruction:
\begin{inlineprompt}
Given a scientific research question, retrieve papers that could inspire or contribute to answering the question, including relevant background knowledge, methods, and ideas.
\end{inlineprompt}

\begin{table*}[t]
\centering
\small
\setlength{\tabcolsep}{2pt}
\renewcommand{\arraystretch}{1.1}
\caption{
\textbf{Retriever configurations used in \benchmark{}.}
Configurations are reported from the official model documentation and checkpoint configuration files.
Maximum lengths denote the reported input limits or recommended retrieval caps of each checkpoint or API.
``Instruction'' indicates whether we prepend the task-specific instruction to queries.
}
\label{tab:retriever_configs}

\begin{threeparttable}
\begin{tabularx}{\textwidth}{
@{}
>{\raggedright\arraybackslash}X
c c c c c c
@{}
}
\toprule
Model
& Size
& Arch.
& {Max.$|Q|$}
& {Max.$|D|$}
& Instr.
& Link \\
\midrule

\modelgroup{7}{Sparse retriever}
\texttt{BM25}~\citep{robertson2009probabilistic}
& --- & Sparse & $\infty$ & $\infty$ & No
& \githublink{https://github.com/xhluca/bm25s} \\

\lightrule
\modelgroup{7}{Multi-vector retriever}
\texttt{LateOn}~\citep{sourty2026denseon}
& 149M & ColBERT & 32 & 300 & No
& \hflink{https://huggingface.co/lightonai/LateOn} \\

\lightrule
\modelgroup{7}{Scientific embedding retrievers}
\texttt{SPECTER2}~\citep{Singh2022SciRepEvalAM}
& 110M & Encoder & 512 & 512 & No
& \githublink{https://github.com/allenai/SPECTER2} \\

\texttt{SciNCL}~\citep{ostendorff2022neighborhood}
& 110M & Encoder & 512 & 512 & No
& \githublink{https://github.com/malteos/scincl} \\

\texttt{OpenScholar}~\citep{asai2024openscholar}
& 110M & Encoder & 512 & 512 & No
& \hflink{https://huggingface.co/collections/OpenSciLM/openscholar-v1}%
  \hspace{0.12em}%
  \githublink{https://github.com/akariasai/openscholar} \\

\lightrule
\modelgroup{7}{General-purpose embedding retrievers}
\providericon{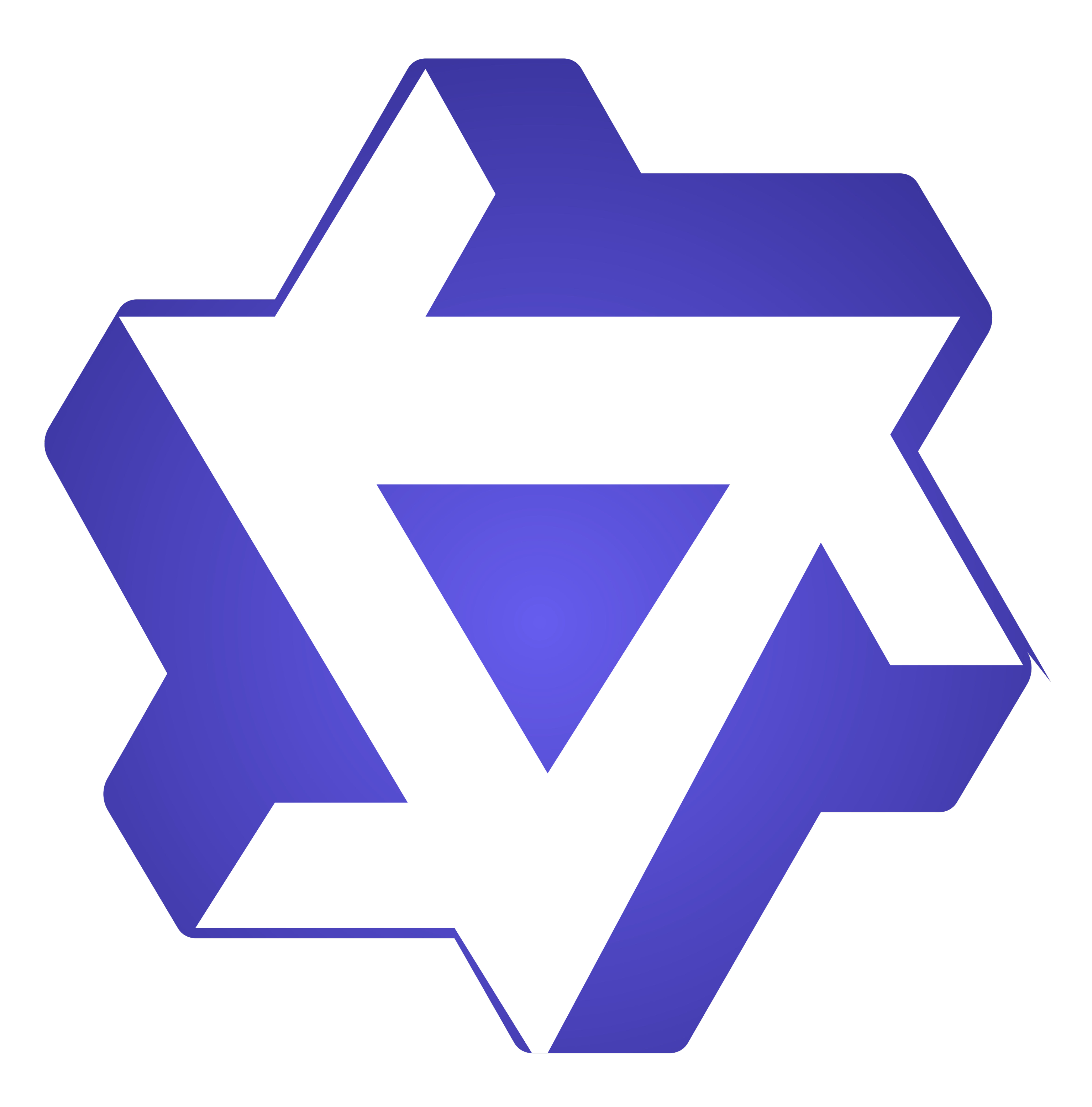}\texttt{Qwen3-Embedding-4B}~\citep{zhang2025qwen3}
& 4B & Decoder & 32K & 32K & Yes
& \hflink{https://huggingface.co/Qwen/Qwen3-Embedding-4B} \\

\providericon{qwen}\texttt{Qwen3-Embedding-8B}~\citep{zhang2025qwen3}
& 8B & Decoder & 32K & 32K & Yes
& \hflink{https://huggingface.co/Qwen/Qwen3-Embedding-8B} \\

\providericon{gemini}\texttt{Gemini Embedding 2}~\citep{shanbhogue2026geminiembedding2native}
& --- & Encoder & 8,192 & 8,192 & No
& \raisebox{-0.3ex}{\includegraphics[height=1.0em]{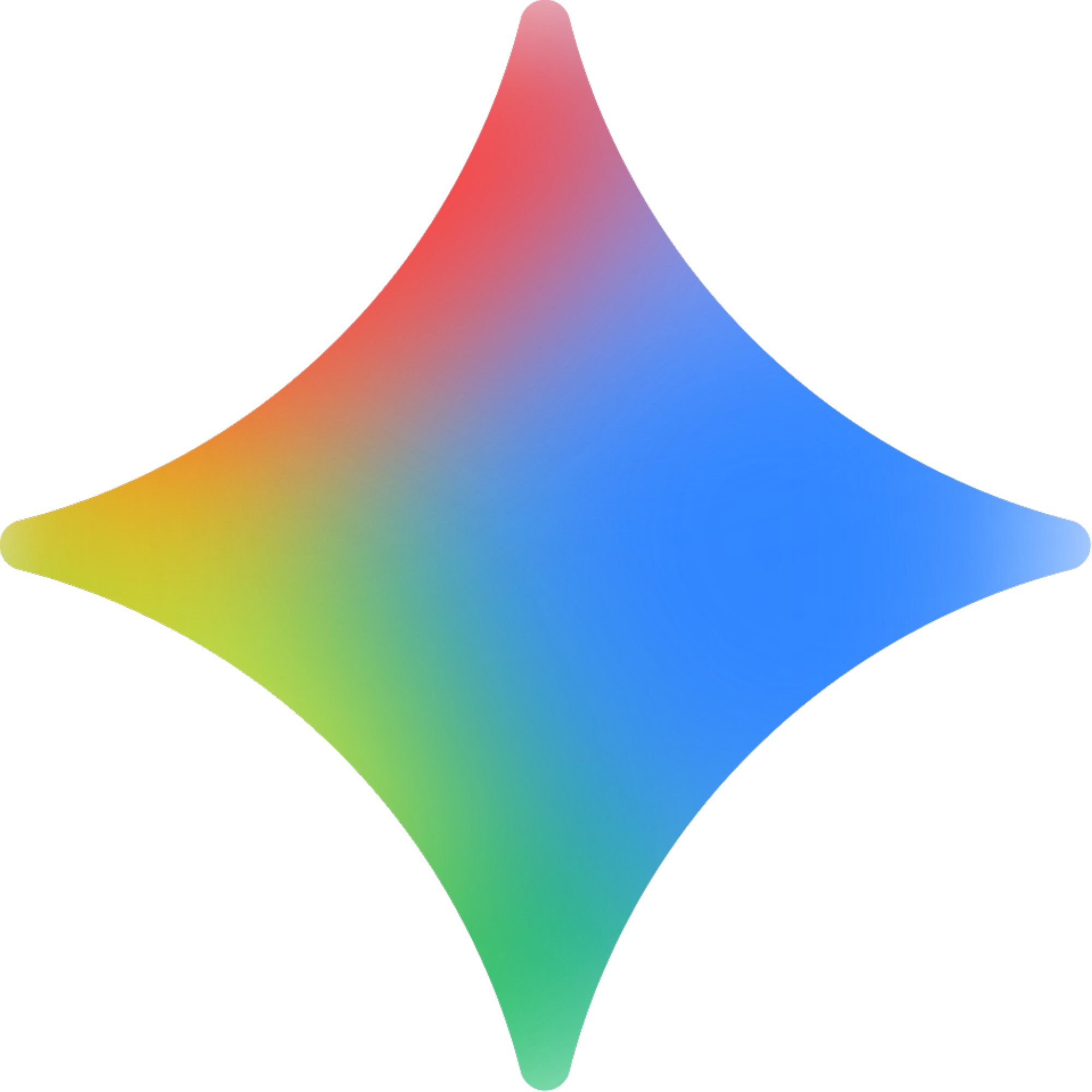}} \\

\texttt{GTE-Qwen2-7B-Instruct}~\citep{li2023towards}
& 7B & Bi-decoder & 32K & 32K & No
& \hflink{https://huggingface.co/Alibaba-NLP/gte-Qwen2-7B-instruct} \\

\lightrule
\modelgroup{7}{Reasoning-optimized embedding retrievers}
\texttt{Reason-Embed-Qwen3-8B}~\citep{chen2025reasonembed}
& 8B & Decoder & 512 & 512 & No
& \hflink{https://huggingface.co/hanhainebula/reason-embed-qwen3-8b-0928} \\

\texttt{Inf-Retriever-v1-Pro}~\citep{inf-x-retriever-2025}
& 7B & Decoder & 8,192 & 8,192 & No
& \hflink{https://huggingface.co/infly/inf-retriever-v1-pro}%
  \hspace{0.12em}%
  \githublink{https://github.com/yaoyichen/INF-X-Retriever} \\

\texttt{NV-Embed-Reasoning-3B}~\citep{nvidia2026llamanvembedreasoning}
& 3B & Bi-decoder & 8,192 & 8,192 & No
& \hflink{https://huggingface.co/nvidia/llama-nv-embed-reasoning-3b} \\

\texttt{Diver-Retriever-4B}~\citep{DIVER}
& 4B & Decoder & 40K & 40K & No
& \githublink{https://github.com/AQ-MedAI/Diver} \\

\bottomrule
\end{tabularx}
\end{threeparttable}
\end{table*}

\begin{table}[t]
\centering
\footnotesize
\setlength{\tabcolsep}{4pt}
\renewcommand{\arraystretch}{1.1}
\caption{
\textbf{Language-model backbones used in \benchmark{}.}
GPT-4.1 and o3 run under the knowledge-cutoff rule of Section~\ref{sec:exp:setup}. The other rows serve the reranking, backbone-swap and leakage analyses, and Claude Fable 5.1 also appears in the rows of Table~\ref{tab:main_results} marked $\ddagger$.
}
\label{tab:llm_backbones}

\begin{threeparttable}
\begin{tabularx}{0.7\columnwidth}{@{}X c@{}}
\toprule
Model &  Knowledge Cutoff \\
\midrule

\modelgroup{2}{Under the knowledge-cutoff rule}
\providericon{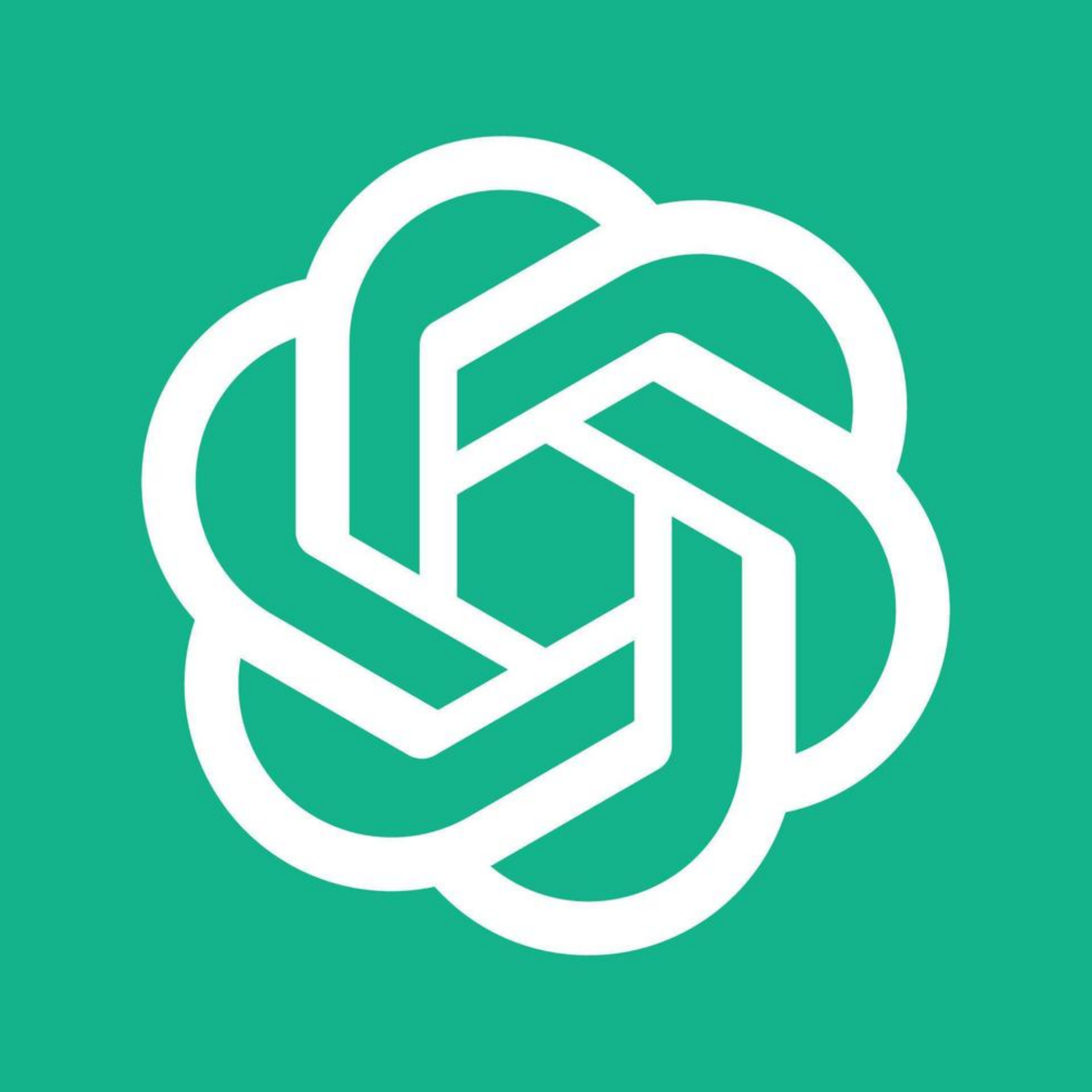}\texttt{GPT-4.1}~\citep{openai2026gpt41model}
& June 2024 \\
\providericon{gpt}\texttt{o3}~\citep{openai2026o3model}
& June 2024 \\
\lightrule
\modelgroup{2}{Analysis backbones}
\providericon{gemini}\texttt{Gemini 2.5 Flash}~\citep{comanici2025gemini}
& January 2025 \\
\providericon{gemini}\texttt{Gemini 3.1} Pro~\citep{google2026gemini31pro}
& January 2025 \\

\lightrule
\providericon{gpt}\texttt{GPT-5.6 Luna}~\citep{openai2026gpt56luna}
& February 2026 \\
\providericon{gpt}\texttt{GPT-5.6 Terra}~\citep{openai2026gpt56terra}
& February 2026 \\
\providericon{gpt}\texttt{GPT-5.6 Sol}~\citep{openai2026gpt56sol}
& February 2026 \\
\providericon{gemini}\texttt{Gemini 3.7 Flash}~\citep{google2026gemini37flash}
& March 2026 \\
\providericon{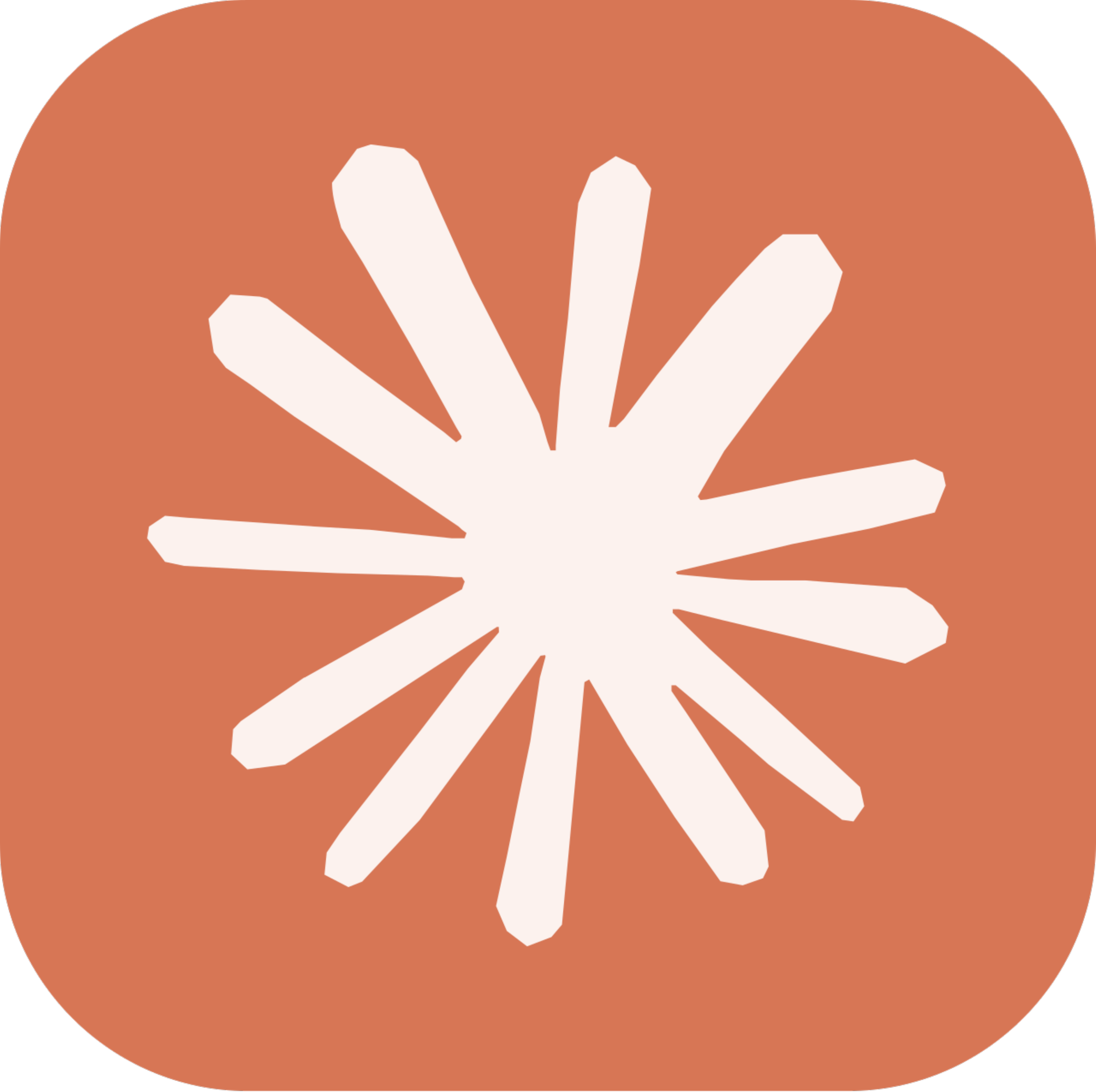}\texttt{Claude Fable 5.1}~\citep{anthropic2026fable51}
& June 2026 \\

\bottomrule
\end{tabularx}
\end{threeparttable}
\end{table}

\subsection{Evaluation Setup for Pre-discovery Retrieval}
\label{app:evaluation-setup}

Our main setting is retrieval from a temporally filtered local corpus. For each query, we exclude corpus papers published after its source paper, so that the retriever can access only literature available when the project began. We disallow web search. Table~\ref{tab:corpus_size_cutoff} reports the resulting searchable corpus size per query. Although the temporal cutoff removes later work, the searchable corpus remains large, with a median of 190{,}574 papers across both query types.

\begin{table}[t]
\centering
\small
\caption{\textbf{Effective searchable corpus size per query after the temporal cutoff.}}
\label{tab:corpus_size_cutoff}
\begin{tabular}{lccccc}
\toprule
& $n$ & Mean & Median & Min & Max \\
\midrule
Full corpus (no cutoff) & \multicolumn{5}{c}{190{,}896} \\
\midrule
\CQ{} (cutoff) & 207 & 190{,}306 & 190{,}574 & 188{,}526 & 190{,}896 \\
\SQ{} (cutoff) & 687 & 190{,}300 & 190{,}574 & 188{,}526 & 190{,}896 \\
All queries (cutoff) & 894 & 190{,}302 & 190{,}574 & 188{,}526 & 190{,}896 \\
\bottomrule
\end{tabular}
\end{table}

\subsection{Agentic Search Baselines}
\label{app:agentic-baselines}

\paragraph{Shared setup.}
Under the main setting, agents search a temporally filtered local corpus of paper titles and abstracts, with no access to the web or the source paper, and their backbones have knowledge cutoffs preceding all source papers.
The corpus-search tool returns the ten papers the retriever ranks highest under the temporal cutoff.
All calls use temperature 0, except GPT-5.6 backbones (\cref{app:postcutoff}), which set a reasoning effort instead, and Claude Fable 5.1, which uses its default thinking setting.

Agents return rankings of different lengths, so we score each on a depth-100 list.
The list starts with the agent's ranking, adds up to 25 papers the agent saw but did not rank, in discovery order, and fills the rest with the retriever ranking of the original query, without duplicates or the source paper.
The cap of 25 keeps an agent that sees many papers but ranks few from displacing the retriever back-fill.
R@5 and R@20 thus come from the agent's own ranking unless it holds fewer than $k$ papers, which happens mainly for the grep agent (440 of 894 queries at R@20, against 10 and 95 for the tool-calling and deep research agents, and at most 93 for any Claude Fable 5.1 row).
Trajectory recall is the share of gold papers among the papers an agent sees during its run.

\paragraph{Tool-calling agent.}
The agent combines the ReAct function-calling pattern~\citep{yao2022react}, the five-round budget of PaperScout~\citep{pan2026paperscout}, and the search-then-select design of PaSa~\citep{he2025pasa}.
It first searches the query as written, then runs up to 5 rounds, each one backbone reply with any number of search calls.
Returned papers enter a pool of at most 60.
The backbone then scores each pooled paper from 0 to 10 from its title and abstract, by how much its ideas could help a researcher pursue the query, and the scores sorted with ties in discovery order give the ranking.
We do not run PaSa or PaperScout as released, because both search the web and PaperScout releases no checkpoint of its trained policy, whose base model has no vendor-stated knowledge cutoff.
We omit their citation-expansion step because the corpus has no citation graph resolved to corpus papers.

\paragraph{Deep research agent.}
The agent follows the plan, search, read, and synthesize procedure of deep research systems~\citep{openai2025deepresearch,zhang2025deep}, with the same search tool and o3~\citep{openai2026o3model} as backbone.
It splits the query into at most 6 sub-questions, searches the query as written, then makes up to 20 searches, 10 reads, and a pool of 150 papers.
A read returns the paper's title and abstract.
After every 5 calls, a reflection step asks which sub-questions remain uncovered.
A final synthesis call ranks up to 50 pooled papers by usefulness, and this list is the ranking, with no per-paper judge.

\paragraph{Grep agent.}
We re-implement DCI-Agent-Lite~\citep{li2026beyond}, which searches the raw corpus using shell commands rather than a retriever.
The released harness has read and shell tools and a 300-turn budget; ours has one shell tool, 20 rounds, and results truncated to 4,000 characters.
The corpus is given as monthly files with one paper per line (identifier, title, abstract, date), and the prompt asks the agent to search abstracts with regular expressions and return an ordered list of identifiers.
The backbone is GPT-4.1.

\paragraph{Claude Fable 5.1 rows.}
We also run all three agents with Claude Fable 5.1~\citep{anthropic2026fable51}, which compares the procedures using a single backbone.
Its June 2026 knowledge cutoff follows all but 6 of the 207 source papers, so these rows fall outside the rule of Section~\ref{sec:exp:setup} and are marked in Table~\ref{tab:main_results}.
The model's safety classifiers sometimes decline a call, which we treat as an empty reply.
Declines occur on 47 of the 894 queries for the tool-calling agent, 95 for the deep research agent, and 33 for the grep agent.
About half of the deep research agent's declines are labeled as requests for the model's reasoning, almost all at or right after the reflection step; all others are labeled biology or cyber-security risk.
A declined judge call ranks the paper after all scored papers, a declined loop turn ends the search, and a declined synthesis leaves the deep research agent with only the seen papers and the back-fill.

\paragraph{Analysis conditions.}
The remaining conditions relax the main setting and are used only for analysis in \cref{app:postcutoff}, not for direct comparison with the query-only agents.
The last three give extra information to the tool-calling agent with Gemini 3.7 Flash~\citep{google2026gemini37flash}, and all are back-filled to depth 100 as above.
\begin{itemize}[leftmargin=*,itemsep=0.5pt,topsep=2pt]
\item \textbf{Closed-book (recall-from-memory).}
A backbone may know the source papers' bibliographies from training, so part of its agent's recall could come from memory.
To measure this, the backbone receives the query and the source paper's title and abstract, has no tools, and names up to 30 prior papers.
Each title is resolved to the top hit of a retriever search under the temporal cutoff, with no similarity threshold, and we drop duplicates, titles whose search returns nothing, and the source paper.
\item \textbf{Source paper.}
This measures how much the query-only restriction costs, since a real user would know more than the query.
The agent's first message includes the source paper's title and abstract.
\item \textbf{Web search.}
This tests whether web access, where the source paper and its citations may appear, can substitute for the local corpus.
The agent also gets a web-search tool (ten results per call) and a page-reading tool (pages truncated to 6,000 characters, at most five per query), and must call a tool in its first two rounds.
A web result enters the pool only if its address contains an arXiv identifier or its normalized title matches a corpus paper.
Other results are shown to the agent but not pooled.
\item \textbf{Bibliography.}
This gives an upper bound on what any leak of the answer could achieve.
We map the source paper's bibliography, taken from its arXiv HTML version, to corpus papers by arXiv identifier or exact title (46 of 54 entries on average).
The leak upper-bound row of Table~\ref{tab:postcutoff_conditions} adds these papers to the agent's pool and prompt, and the bibliography-alone row ranks them in citation order without a model.
\end{itemize}

\section{Additional Analyses}
\label{app:experimental-details}
This appendix gives details and supplementary results for our extended comparison of embedding retrievers and for the analyses of LLM reranking, knowledge cutoffs, query generation, full-text reading, and agent search failures.

\subsection{Extended Results for Embedding Retrievers}
\label{app:extended_embeddings}
This section reports extended results for a broader set of recent embedding retrievers, including models trained on scientific documents and models optimized for reasoning-intensive retrieval.
Recently, several embedding retrievers have been developed for reasoning-intensive retrieval, where identifying relevant documents for complex queries requires reasoning beyond surface-level semantic matching~\citep{xiao2024rar,su2025bright,tchuindjo2026obliq}.
We evaluate SciNCL~\citep{ostendorff2022neighborhood}, OpenScholar~\citep{asai2024openscholar}, GTE-Qwen2-7B~\citep{li2023towards}, ReasonEmbed~\citep{chen2025reasonembed}, INF-X-Retriever~\citep{inf-x-retriever-2025}, Llama NV-Embed-Reasoning-3B~\citep{nvidia2026llamanvembedreasoning}, and DIVER~\citep{DIVER}.
Interestingly, none of the reasoning-optimized models improves over the strongest general-purpose embeddings. Their lack of gains suggests that current objectives for reasoning-intensive retrieval do not adequately capture the relationships that make a prior paper useful for advancing a research project.
As discussed in \Cref{sec:conclusion}, closing this gap may require an expert-level sense of which ideas matter, enabling retrieval of useful connections across fields beyond the scope of any individual expert.

\begin{table*}[t]
\centering
\small
\setlength{\tabcolsep}{4pt}
\renewcommand{\arraystretch}{1.1}
\caption{
\textbf{Extended results for embedding retrievers on \benchmark{}.}
Recall@$k$ ($k$=5, 20, 100) and nDCG@20 across both query types.
This table includes all embedding retrievers evaluated in our experiments.
}
\label{tab:extended_embedding_results}
\resizebox{\textwidth}{!}{
\begin{tabular}{l *{8}{c}}
\toprule
& \multicolumn{4}{c}{\textcolor{cqcolor}{\optimisticfont Core Research Query}}
& \multicolumn{4}{c}{\textcolor{sqcolor}{\optimisticfont Subfield-specific Query}} \\
\cmidrule(lr){2-5}\cmidrule(lr){6-9}
Model & nDCG@20 & R@5 & R@20 & R@100 & nDCG@20 & R@5 & R@20 & R@100 \\
\midrule
\multicolumn{9}{l}{\textit{Sparse retriever}} \\[1pt]
BM25
& \qbelow{40}{0.16} & \qbelow{35}{0.12} & \qbelow{55}{0.23} & \qbelow{55}{0.39}
& \qbelow{60}{0.21} & \qbelow{55}{0.12} & \qbelow{75}{0.26} & \qbelow{75}{0.45} \\
\lightrule
\multicolumn{9}{l}{\textit{Multi-vector retriever}} \\[1pt]
LateOn 0.1B
& \qbelow{40}{0.16} & \qbelow{35}{0.12} & \qbelow{50}{0.24} & \qbelow{60}{0.37}
& \qbelow{55}{0.27} & \qbelow{35}{0.16} & \qbelow{60}{0.33} & \qbelow{60}{0.55} \\
\lightrule
\multicolumn{9}{l}{\textit{Scientific embedding retrievers}} \\[1pt]
SPECTER2
& \qbelow{55}{0.13} & \qbelow{45}{0.10} & \qbelow{65}{0.20} & \qbelow{60}{0.38}
& \qbelow{75}{0.18} & \qbelow{60}{0.11} & \qbelow{85}{0.23} & \qbelow{85}{0.42} \\
SciNCL
& \qbelow{55}{0.13} & \qbelow{45}{0.10} & \qbelow{65}{0.20} & \qbelow{55}{0.40}
& \qbelow{75}{0.18} & \qbelow{65}{0.10} & \qbelow{85}{0.22} & \qbelow{85}{0.42} \\
OpenScholar
& \qbelow{40}{0.16} & \qbelow{30}{0.13} & \qbelow{55}{0.23} & \qbelow{55}{0.40}
& \qbelow{70}{0.19} & \qbelow{55}{0.12} & \qbelow{80}{0.24} & \qbelow{75}{0.46} \\
\lightrule
\multicolumn{9}{l}{\textit{General-purpose embedding retrievers}} \\[1pt]
GTE-Qwen2-7B
& \qbelow{10}{0.23} & \qbelow{15}{0.17} & \qbelow{18}{0.32} & \qbelow{25}{0.51}
& \qbelow{40}{0.30} & \qbelow{25}{0.18} & \qbelow{45}{0.38} & \qbelow{40}{0.62} \\
Qwen3-Emb-4B
& \qbelow{5}{0.25} & \qbelow{12}{0.18} & \qabove{8}{\textbf{0.39}} & \qabove{5}{0.59}
& \qbelow{20}{0.35} & \qbelow{15}{0.21} & \qbelow{25}{0.44} & \qbelow{15}{0.69} \\
Qwen3-Emb-8B
& \qref{0.26} & \qref{\textbf{0.21}} & \qref{0.37} & \qref{0.58}
& \qref{\textbf{0.41}} & \qref{\textbf{0.25}} & \qref{\textbf{0.51}} & \qref{\textbf{0.73}} \\
Gemini-Emb-2
& \qabove{5}{\textbf{0.27}} & \qref{\textbf{0.21}} & \qabove{5}{0.38} & \qabove{18}{\textbf{0.62}}
& \qbelow{10}{0.38} & \qbelow{5}{0.24} & \qbelow{18}{0.46} & \qbelow{10}{0.70} \\
\lightrule
\multicolumn{9}{l}{\textit{Reasoning-optimized embedding retrievers}} \\[1pt]
NV-Embed-Reasoning-3B
& \qbelow{10}{0.23} & \qbelow{10}{0.18} & \qbelow{18}{0.32} & \qbelow{25}{0.51}
& \qbelow{30}{0.32} & \qbelow{18}{0.20} & \qbelow{40}{0.40} & \qbelow{35}{0.63} \\
Diver-Retriever-4B
& \qref{0.26} & \qbelow{8}{0.19} & \qbelow{5}{0.36} & \qbelow{8}{0.56}
& \qbelow{15}{0.37} & \qbelow{8}{0.23} & \qbelow{20}{0.45} & \qbelow{10}{0.70} \\
Reason-Embed-Qwen3-8B
& \qbelow{10}{0.23} & \qbelow{10}{0.18} & \qbelow{10}{0.33} & \qbelow{15}{0.53}
& \qbelow{30}{0.33} & \qbelow{18}{0.19} & \qbelow{35}{0.40} & \qbelow{30}{0.65} \\
Inf-Retriever-v1-Pro
& \qref{0.26} & \qbelow{5}{0.20} & \qref{0.37} & \qbelow{18}{0.53}
& \qbelow{35}{0.31} & \qbelow{20}{0.19} & \qbelow{45}{0.38} & \qbelow{35}{0.63} \\
\bottomrule
\end{tabular}
}
\end{table*}

\subsection{LLM Reranking}
\label{app:reranking}
\begin{figure}[!ht]
\centering
\includegraphics[width=\textwidth]{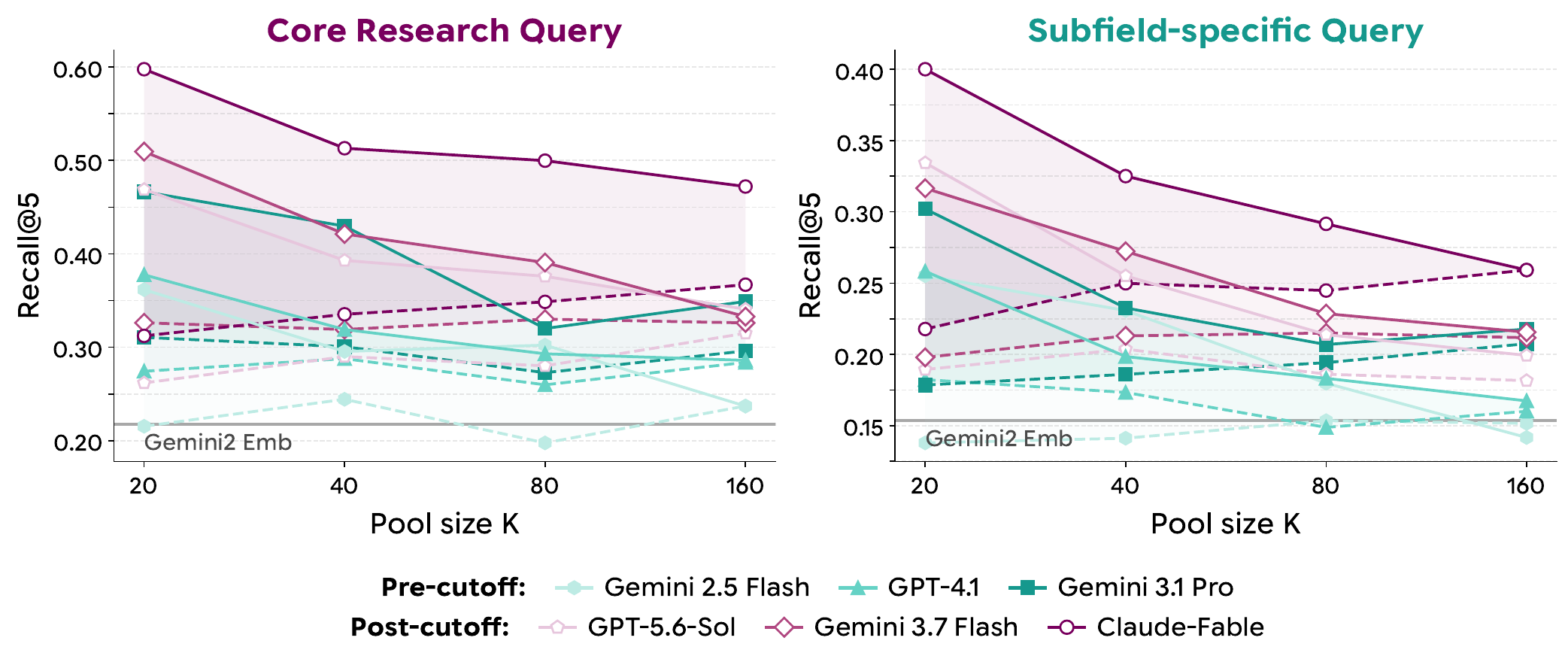}
\caption{\textbf{Reranking performance as the candidate pool grows.}
Recall@5 for six LLM backbones across candidate pools from Gemini-Emb-2.
Solid curves show oracle reranking and dashed curves show standard reranking.
The gray horizontal line marks retriever-only performance.}
\label{fig:reranking_pool_sizes}
\end{figure}

We test whether an LLM can reorder the retriever's candidates to place gold papers higher, using two procedures over candidate pools from Gemini-Embedding-2.
The \textbf{listwise reranker} gives the top 20 candidates to a single LLM call, which returns them in ranked order.
The \textbf{tournament reranker} handles larger pools by ranking batches of $b=20$ papers and promoting the top $k=4$ of each batch to the next round, until one batch remains and its ranking is the final order.
Both use the prompt in \cref{app:analysis-prompts}.
We additionally extend the evaluation to candidate pools of $K \in \{20, 40, 80, 160\}$; \cref{tab:reranking_pool_sizes,fig:reranking_pool_sizes} report standard and oracle reranking results across these pool sizes.

\begin{table}[!ht]
\centering
\small
\setlength{\tabcolsep}{4pt}
\renewcommand{\arraystretch}{1.1}
\caption{\textbf{Reranking performance across candidate-pool sizes.}
Recall@5 on 50 queries of each type after reranking Gemini-Emb-2's top-$K$ candidates.
Bold and underlining indicate the highest and second-highest values, respectively, within each condition and column.}
\label{tab:reranking_pool_sizes}
\resizebox{\textwidth}{!}{
\begin{tabular}{ll *{8}{c}}
\toprule
& & \multicolumn{4}{c}{\textcolor{cqcolor}{\optimisticfont Core Research Query}} & \multicolumn{4}{c}{\textcolor{sqcolor}{\optimisticfont Subfield-specific Query}} \\
\cmidrule(lr){3-6}\cmidrule(lr){7-10}
Verifier & Condition & $K{=}20$ & $K{=}40$ & $K{=}80$ & $K{=}160$ & $K{=}20$ & $K{=}40$ & $K{=}80$ & $K{=}160$ \\
\midrule
GPT-4.1 & Standard & 0.27 & 0.29 & 0.26 & 0.28 & 0.18 & 0.17 & 0.15 & 0.16 \\
        & Oracle   & 0.38 & 0.32 & 0.29 & 0.29 & 0.26 & 0.20 & 0.18 & 0.17 \\
Gemini 2.5 Flash & Standard & 0.22 & 0.24 & 0.20 & 0.24 & 0.14 & 0.14 & 0.15 & 0.15 \\
                 & Oracle   & 0.36 & 0.30 & 0.30 & 0.24 & 0.26 & 0.23 & 0.18 & 0.14 \\
Gemini 3.1 Pro & Standard & 0.31 & 0.30 & 0.27 & 0.30 & 0.18 & 0.19 & 0.19 & 0.21 \\
               & Oracle   & 0.47 & \underline{0.43} & 0.32 & \underline{0.35} & 0.30 & 0.23 & 0.21 & \underline{0.22} \\
Gemini 3.7 Flash & Standard & \textbf{0.33} & \underline{0.32} & \underline{0.33} & \underline{0.33} & \underline{0.20} & \underline{0.21} & \underline{0.22} & \underline{0.21} \\
                 & Oracle   & \underline{0.51} & 0.42 & \underline{0.39} & 0.33 & 0.32 & \underline{0.27} & \underline{0.23} & 0.22 \\
GPT-5.6 Sol & Standard & 0.26 & 0.29 & 0.28 & 0.32 & 0.19 & 0.20 & 0.19 & 0.18 \\
            & Oracle   & 0.47 & 0.39 & 0.38 & 0.34 & \underline{0.33} & 0.26 & 0.21 & 0.20 \\
Claude Fable & Standard & \underline{0.31} & \textbf{0.34} & \textbf{0.35} & \textbf{0.37} & \textbf{0.22} & \textbf{0.25} & \textbf{0.25} & \textbf{0.26} \\
             & Oracle   & \textbf{0.60} & \textbf{0.51} & \textbf{0.50} & \textbf{0.47} & \textbf{0.40} & \textbf{0.33} & \textbf{0.29} & \textbf{0.26} \\
\bottomrule
\end{tabular}
}
\end{table}

These results suggest that reranking requires recognizing which prior papers in a retrieved candidate set contain ideas that may inspire the project. Reranking performance generally improves with the backbone's reasoning capability. Claude Fable 5.1 is the most recent backbone we evaluate and achieves the strongest results. Weaker models provide little benefit and can even underperform the retriever-only baseline on \SQ{} queries.

\subsection{Post-cutoff Backbones and Benchmark Leakage}
\label{app:postcutoff}

\begin{figure}[t]
\centering
\includegraphics[width=0.8\linewidth]{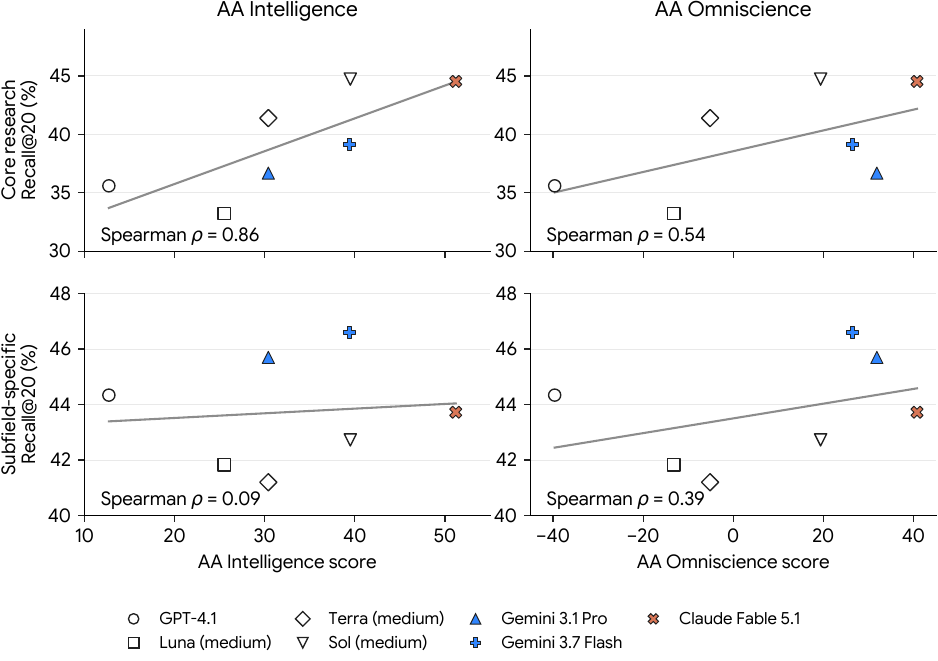}
\caption{\textbf{Backbone capability tracks \rqlevel{} recall but not \threadlevel{} recall.}
Recall@20 of the tool-calling agent with seven backbones on the post-cutoff split, with both date slices pooled, plotted against the Artificial Analysis Intelligence and Omniscience scores~\citep{artificialanalysis2026index}.
The top row shows the 46 \CQ{} and the bottom row the 157 \SQ{}.
Spearman $\rho$ with the Intelligence Index is 0.86 at the \CQ{} and 0.09 at the \SQ{}.
GPT-5.6 Luna, Terra, and Sol use medium reasoning effort.
Lines are least-squares fits, and y-axis ranges differ between rows.
}
\label{fig:backbone_query_correlations}
\end{figure}

\begin{figure}[t]
\centering
\includegraphics[width=0.8\linewidth]{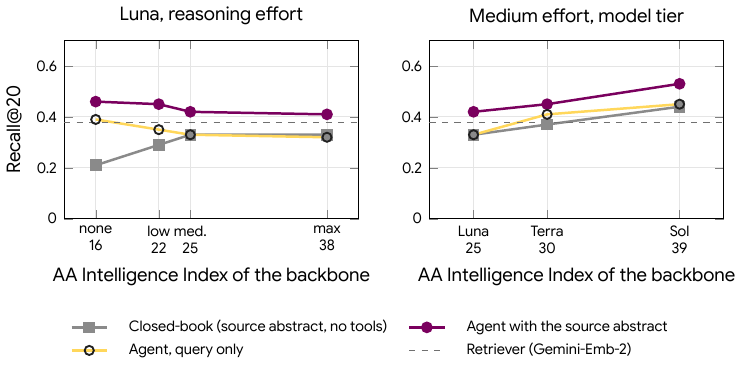}
\caption{\textbf{With the same source context, search helps weaker backbones more than memory alone.}
\CQ{} R@20 over the 46 \CQ{} of the post-cutoff split, with both date slices pooled. All variants share the February 2026 knowledge cutoff~\citep{openai2026gpt56luna,openai2026gpt56sol,openai2026gpt56terra}. Filled marks show the tool-calling agent with the source title and abstract, hollow marks show the query-only agent, and closed-book marks show papers named from memory with the same source context. The left panel varies Luna's reasoning effort, and the right panel compares Luna, Terra, and Sol at medium effort. The dashed line marks Gemini-Emb-2 at 0.38.}
\label{fig:aa_backbones}
\end{figure}
\textbf{Setup.}
Our main experimental setup keeps every backbone's knowledge cutoff before every source paper and gives the agent neither the source paper nor web access.
This section asks what happens when these restrictions are lifted: whether newer backbones gain from having seen the source papers, and how much the source paper, the web, or its bibliography would help.
We use two slices of the author-written set, matched by primary arXiv category.
The pre-cutoff slice contains 23 source papers published between February 12 and December 23, 2025, with two cs.SE slots filled by cs.CE and cs.MA papers.
The post-cutoff slice contains 23 source papers published between May 8 and August 14, 2026, four months later.
A backbone whose cutoff falls between the slices can have seen the first but not the second.

The pre-cutoff slice is harder for every system (\rqlevel{} retriever R@100 of 0.49 against 0.63), so we compare each system with the retriever within the same slice.
\cref{tab:postcutoff_split} evaluates five backbones and their closed-book counterparts, and~\cref{tab:postcutoff_conditions} varies the availability of access to source papers, the web, and the bibliography for Gemini 3.7 Flash.

\textbf{The cutoff split shows no sign that stronger backbones gain from having seen the source papers.}
With GPT-5.6 Sol and Claude Fable 5.1, the tool-calling agent exceeds the retriever at \rqlevel{} R@20 on both slices, while the other three backbones stay close to it (\cref{tab:postcutoff_split}).
Because every system scores higher on the post-cutoff slice, we compare each backbone's margin over the retriever within a slice.
For GPT-5.6 Sol, whose cutoff falls between the slices, the margin is the same on both (0.07 and 0.06), so it shows no gain from having seen the earlier papers.
Claude Fable 5.1 cannot be tested this way because its cutoff falls inside the post-cutoff slice.
Closed-book recall rises with capability, but this fits memorization and general capability equally well, so it does not settle the question.
At the \threadlevel{} level, no agent or closed-book row exceeds the retriever on either slice.

\textbf{With the same source context, search adds more to weak backbones than to strong ones.}
\cref{fig:aa_backbones} holds the February 2026 cutoff fixed across the GPT-5.6 family and varies capability through Luna's reasoning effort and the model tier.
Given the source abstract, the agent beats recall from memory by 0.25 for Luna at no reasoning effort, and the gap shrinks to within noise for the stronger runs.
Raising Luna's effort helps memory and hurts the query-only agent, while a stronger tier helps both.
\cref{fig:backbone_query_correlations} shows a related pattern across seven backbones: capability tracks \rqlevel{} recall but not \threadlevel{} recall.

\textbf{The source abstract helps, web search does not, and the bibliography nearly gives away the answer.}
With Gemini 3.7 Flash, the source title and abstract raise \rqlevel{} R@20 by 0.11 on both slices (\cref{tab:postcutoff_conditions}).
Web search leaves recall almost unchanged, even though the source paper appears in the web results for 85 of the 203 queries.
The bibliography alone contains nearly all gold papers (R@100 above 0.9).
The main protocol therefore withholds the source paper, its bibliography, and the web.

\begin{table}[t]
\centering
\small
\setlength{\tabcolsep}{4pt}
\renewcommand{\arraystretch}{1.1}
\caption{
\textbf{GPT-5.6 Sol and Claude Fable 5.1 exceed Gemini-Emb-2 at \CQ{} R@20 on both cutoff slices.}
\CQ{} Recall@$k$ ($k$=5, 20, 100) and trajectory recall for the tool-calling agent with five backbones and their matching closed-book conditions. Gemini-Emb-2 is the retriever reference within each slice. Closed-book rows receive the source paper's title and abstract and name papers from memory. Agent rows receive the query only. Parentheses indicate vendor-stated knowledge cutoffs.
}
\label{tab:postcutoff_split}
\resizebox{\linewidth}{!}{\begin{tabular}{l *{8}{c}}
\toprule
& \multicolumn{4}{c}{\textbf{Pre-cutoff slice}} & \multicolumn{4}{c}{\textbf{Post-cutoff slice}} \\
\cmidrule(lr){2-5}\cmidrule(lr){6-9}
System & R@5 & R@20 & R@100 & Traj. & R@5 & R@20 & R@100 & Traj. \\
\midrule
Gemini-Emb-2 retriever            & 0.23 & 0.35 & 0.49 &  & 0.25 & 0.41 & 0.63 &  \\
\midrule
\multicolumn{9}{l}{\textit{Tool-calling agent}} \\
GPT-4.1 (June 2024)                          & 0.20 & 0.32 & 0.44 & 0.33 & 0.25 & 0.39 & 0.60 & 0.42 \\
Gemini 3.1 Pro (January 2025)                & 0.21 & 0.36 & 0.47 & 0.40 & 0.26 & 0.37 & 0.53 & 0.45 \\
GPT-5.6 Sol (February 2026)                  & 0.21 & 0.42 & 0.50 & 0.49 & 0.25 & 0.47 & 0.64 & 0.58 \\
Gemini 3.7 Flash (March 2026)                & 0.29 & 0.37 & 0.49 & 0.38 & 0.28 & 0.41 & 0.59 & 0.41 \\
Claude Fable 5.1 (June 2026)                 & 0.36 & 0.42 & 0.60 & 0.64 & 0.26 & 0.47 & 0.66 & 0.63 \\
\midrule
\multicolumn{9}{l}{\textit{Closed-book, given the source title and abstract}} \\
GPT-4.1                                      & 0.07 & 0.14 & 0.57 & 0.14 & 0.05 & 0.12 & 0.64 & 0.15 \\
Gemini 3.1 Pro                               & 0.20 & 0.29 & 0.61 & 0.31 & 0.18 & 0.26 & 0.72 & 0.26 \\
GPT-5.6 Sol                                  & 0.30 & 0.40 & 0.64 & 0.40 & 0.40 & 0.48 & 0.81 & 0.51 \\
Gemini 3.7 Flash                             & 0.25 & 0.34 & 0.66 & 0.34 & 0.20 & 0.29 & 0.72 & 0.28 \\
Claude Fable 5.1                             & 0.36 & 0.55 & 0.66 & 0.56 & 0.44 & 0.54 & 0.80 & 0.55 \\
\bottomrule
\end{tabular}
}
\end{table}

\begin{table}[t]
\centering
\small
\setlength{\tabcolsep}{4pt}
\renewcommand{\arraystretch}{1.1}
\caption{
\textbf{Source-paper context improves \CQ{} retrieval, whereas web search changes little.}
\CQ{} Recall@$k$ ($k$=5, 20, 100) and trajectory recall of the Gemini 3.7 Flash tool-calling agent on the two cutoff slices. The local-corpus row is the main protocol. The remaining rows add the source paper's title and abstract, web search, or mapped source-bibliography entries. The bibliography-alone row ranks the same entries in citation order without a model.
}
\label{tab:postcutoff_conditions}
\resizebox{\linewidth}{!}{\begin{tabular}{l *{8}{c}}
\toprule
& \multicolumn{4}{c}{\textbf{Pre-cutoff slice}} & \multicolumn{4}{c}{\textbf{Post-cutoff slice}} \\
\cmidrule(lr){2-5}\cmidrule(lr){6-9}
Condition & R@5 & R@20 & R@100 & Traj. & R@5 & R@20 & R@100 & Traj. \\
\midrule
Local corpus only                                        & 0.29 & 0.37 & 0.49 & 0.38 & 0.28 & 0.41 & 0.59 & 0.41 \\
+ source paper (title and abstract)                      & 0.34 & 0.48 & 0.57 & 0.51 & 0.34 & 0.52 & 0.69 & 0.56 \\
+ web search                                             & 0.29 & 0.34 & 0.48 & 0.37 & 0.28 & 0.42 & 0.61 & 0.44 \\
+ source-paper bibliography           & 0.36 & 0.68 & 0.93 & 0.93 & 0.42 & 0.81 & 0.98 & 0.98 \\
Source-paper bibliography alone                & 0.23 & 0.60 & 0.93 & 0.93 & 0.18 & 0.60 & 0.98 & 0.98 \\
\bottomrule
\end{tabular}
}
\end{table}

\subsection{Query Generation Methods}
\label{app:query-augmentation}
These methods test whether rewriting the research question with an LLM brings retrieval closer to the gold papers, which often use different vocabulary from the question.
All use GPT 4.1, with the prompts in \cref{app:analysis-prompts} as the system message and the research question as the user message.
\begin{itemize}[leftmargin=*,itemsep=0.5pt,topsep=2pt]
\item \textbf{Single-query expansion.}
The model's output, the question plus a short continuation, replaces the question as the retrieval query.
\item \textbf{Multi-query generation.}
The model returns up to five augmented questions, each used as a separate retrieval query. Their top-1{,}000 rankings are fused with reciprocal rank fusion ($k$=60, \citet{cormack2009reciprocal}).
\item \textbf{HyDE}~\citep{gao2023precise}.
The model writes five hypothetical abstracts of prior papers, and each is used as a retrieval query, with the rankings fused as above.
Its user message is the question prefixed with \texttt{Research question:}.
\end{itemize}

Tables~\ref{tab:qual_single_multi} and \ref{tab:qual_hyde} show generated outputs for the same \CQ{} instance from the fixed \CQ{}/\SQ{} subset.
For this instance, dense retrieval on the original question recovers a gold paper that the generated queries do not recover in their top 100 results.
Table~\ref{tab:qual_single_multi} shows the original question alongside single- and multi-query generation.
Table~\ref{tab:qual_hyde} shows three hypothetical abstracts generated by HyDE for the same question.

\begin{table*}[p]
\centering
\footnotesize
\setlength{\tabcolsep}{5pt}
\renewcommand{\arraystretch}{1.08}
\caption{
\textbf{Generated queries for a \CQ{} instance.}
We show the original question, the single-query expansion, and five multi-query outputs.
}
\label{tab:qual_single_multi}
\begin{tabularx}{\textwidth}{
@{}>{\raggedright\arraybackslash}p{0.18\textwidth}
>{\raggedright\arraybackslash}X@{}
}
\toprule
Original query&
{\ttfamily Enhancing the reasoning capabilities of large language models traditionally involves supervised fine-tuning followed by reinforcement learning applied to reasoning-specific datasets. Recent advancements have shown that applying reinforcement learning directly to base models can autonomously elicit sophisticated reasoning behaviors without relying on supervised fine-tuning data. However, these methods still depend on rule-based verifiers, verified golden answers, or pre-trained reward models to generate reward signals during the reinforcement learning phase. Relying on such explicit feedback limits scalability in domains where ground-truth verification is costly or unavailable. Can we enhance the reasoning capability of LLMs in a fully unsupervised manner?} \\
\midrule
Single-query expansion &
{\ttfamily Enhancing the reasoning capabilities of large language models traditionally involves supervised fine-tuning followed by reinforcement learning applied to reasoning-specific datasets. Recent advancements have shown that applying reinforcement learning directly to base models can autonomously elicit sophisticated reasoning behaviors without relying on supervised fine-tuning data. However, these methods still depend on rule-based verifiers, verified golden answers, or pre-trained reward models to generate reward signals during the reinforcement learning phase. Relying on such explicit feedback limits scalability in domains where ground-truth verification is costly or unavailable. Can we enhance the reasoning capability of LLMs in a fully unsupervised manner? I am particularly interested in prior work exploring self-supervised objectives, intrinsic motivation, or unsupervised reward shaping techniques for language models, as well as approaches that leverage emergent properties or self-consistency signals to improve reasoning without external supervision.} \\
\midrule
Multi-query generation &
{\ttfamily (1) Enhancing the reasoning capability of LLMs in a fully unsupervised manner? I am particularly interested in approaches that leverage self-supervised objectives or intrinsic motivation signals to guide reasoning skill acquisition without any external reward models.} \newline

{\ttfamily (2) Enhancing the reasoning capability of LLMs in a fully unsupervised manner? I would like to explore whether emergent reasoning behaviors can be fostered through unsupervised pre-training on synthetic or automatically generated reasoning tasks.} \newline

{\ttfamily (3) Enhancing the reasoning capability of LLMs in a fully unsupervised manner? I am curious about prior work that investigates the use of mutual information maximization or contrastive learning techniques to improve reasoning without labeled data.} \newline

{\ttfamily (4) Enhancing the reasoning capability of LLMs in a fully unsupervised manner? I am seeking studies that examine the role of curriculum learning or self-play in enabling models to develop reasoning skills autonomously.} \newline

{\ttfamily (5) Enhancing the reasoning capability of LLMs in a fully unsupervised manner? I am interested in research that analyzes the emergence of reasoning abilities in large-scale language models trained solely on raw text corpora, without any explicit reasoning supervision.} \\
\bottomrule
\end{tabularx}
\end{table*}

\begin{table*}[p]
\centering
\footnotesize
\setlength{\tabcolsep}{5pt}
\renewcommand{\arraystretch}{1.08}
\caption{\textbf{HyDE-generated hypothetical abstracts.}
We show three of the five abstracts generated for the same \CQ{} instance as Table~\ref{tab:qual_single_multi}.}
\label{tab:qual_hyde}
\begin{tabularx}{\textwidth}{
@{}>{\raggedright\arraybackslash}p{0.18\textwidth}
>{\raggedright\arraybackslash}X@{}
}
\toprule
Original query &
{\ttfamily Enhancing the reasoning capabilities of large language models traditionally involves supervised fine-tuning followed by reinforcement learning applied to reasoning-specific datasets. Recent advancements have shown that applying reinforcement learning directly to base models can autonomously elicit sophisticated reasoning behaviors without relying on supervised fine-tuning data. However, these methods still depend on rule-based verifiers, verified golden answers, or pre-trained reward models to generate reward signals during the reinforcement learning phase. Relying on such explicit feedback limits scalability in domains where ground-truth verification is costly or unavailable. Can we enhance the reasoning capability of LLMs in a fully unsupervised manner?} \\
\midrule
Hypothetical abstract 1 &
{\ttfamily Emergent Reasoning in Language Models via Self-Consistency Training. Recent work has demonstrated that large language models can exhibit improved reasoning abilities when trained to produce self-consistent outputs. In this study, we investigate a self-consistency training paradigm where models are encouraged to generate multiple solutions to a given problem and are rewarded for producing answers that agree with the majority of their own outputs. We apply this approach to arithmetic and commonsense reasoning tasks, showing that self-consistency training leads to modest improvements in accuracy without requiring external supervision or gold-standard answers. Our analysis suggests that self-consistency acts as a form of internal regularization, promoting the emergence of more robust reasoning patterns. These findings highlight the potential of leveraging a model's own outputs as a weak supervisory signal for enhancing reasoning capabilities.} \\
\midrule
Hypothetical abstract 2 &
{\ttfamily Unsupervised Skill Acquisition in Reinforcement Learning via Intrinsic Motivation. Traditional reinforcement learning relies on extrinsic rewards provided by the environment, which can be sparse or unavailable in many real-world scenarios. We propose an unsupervised RL framework where agents are driven by intrinsic motivation signals, such as curiosity or novelty, to autonomously acquire diverse skills. Our method employs a self-supervised objective that rewards agents for exploring novel states and behaviors, without access to task-specific rewards or external supervision. Experiments in simulated environments demonstrate that agents trained with intrinsic motivation develop a repertoire of complex behaviors, including multi-step reasoning and problem-solving strategies. This work suggests that unsupervised RL can serve as a foundation for skill acquisition in domains where explicit reward signals are difficult to define.} \\
\midrule
Hypothetical abstract 3 &
{\ttfamily Bootstrapping Commonsense Knowledge in Neural Language Models via Iterative Self-Labeling. Acquiring commonsense knowledge remains a challenge for neural language models, particularly in the absence of large-scale annotated datasets. We introduce an iterative self-labeling approach in which a language model generates candidate answers to commonsense questions and then refines its predictions by training on its own high-confidence outputs. Our method leverages confidence estimation to select pseudo-labels, enabling the model to bootstrap its knowledge without external supervision. We evaluate our approach on several commonsense reasoning benchmarks and observe consistent improvements over baseline models trained without self-labeling. Our results indicate that iterative self-labeling can be an effective strategy for enhancing the reasoning abilities of language models in a data-efficient and unsupervised manner.} \\
\bottomrule
\end{tabularx}
\end{table*}

\subsection{Reading Full Papers versus Abstracts}
\label{app:read-policies}

Agents in the main setting see only titles and abstracts.
An abstract states a paper's main result but often omits other critical details, so a gold paper could look unrelated to the question from its abstract alone.
Indexing full text is not a simple fix, since embedding-based retrievers handle long documents poorly~\citep{zhu2024longembed} and a paper's body far exceeds the input length they are trained on.
We therefore keep search over abstracts and instead let the agent read more of the papers it finds, testing whether that extra text helps it judge and pursue candidates.

\paragraph{Full-text subset.}
Full text is available for 180,399 corpus papers, converted from arXiv HTML, \LaTeX{} source, or PDF and split by section heading.
A query qualifies when all of its gold papers have full text.
A source paper enters the subset when its \CQ{} qualifies and at least one of its \SQ{} does, giving 158 \CQ{} and 403 \SQ{}.
We sample 50 queries of each type, separate from the 100 queries used in the other analyses.
The search tool and the back-fill are restricted to the full-text papers.
Reads share the five-round budget with searches.
All other settings follow that of the tool-calling agent.

\paragraph{Read policies.}
Four policies add to the abstract the introduction, the method sections, the bibliography, or both the introduction and the bibliography.
A fifth returns the full paper, and a sixth lets the agent request one section from a list returned by a second tool.
Introductions are matched by their heading, methods by headings containing method, approach, model, framework, algorithm, or architecture, and bibliographies by references or bibliography.
If no heading matches, the tool returns the abstract.
Each read is capped at 12,000 characters, about 41\% of the body at the median for papers read in full, and an included bibliography has its own 6,000-character cap.

\begin{table}[t]
\centering
\small
\setlength{\tabcolsep}{4pt}
\renewcommand{\arraystretch}{1.1}
\caption{
\textbf{Reading beyond abstracts changes agent recall little.}
Recall@$k$ ($k$=5, 20, 100), nDCG@20, and trajectory recall for GPT-4.1 with Gemini-Emb-2 search on 50 \CQ{} and 50 \SQ{} from the 180,399-paper full-text subset. The table gives the full results behind Figure~\ref{fig:read_policies}.
}
\label{tab:read_policies_full}
\resizebox{\linewidth}{!}{\begin{tabular}{l *{8}{c} c}
\toprule
& \multicolumn{4}{c}{\textcolor{cqcolor}{\optimisticfont Core Research Query}}
& \multicolumn{4}{c}{\textcolor{sqcolor}{\optimisticfont Subfield-specific Query}} \\
\cmidrule(lr){2-5}\cmidrule(lr){6-9}
Read policy & nDCG@20 & R@5 & R@20 & R@100 & nDCG@20 & R@5 & R@20 & R@100 & Traj. \\
\midrule
Gemini-Emb-2 retriever         & 0.25 & 0.19 & 0.35 & 0.61 & 0.43 & 0.27 & 0.55 & 0.79 &  \\
Abstract only (no read tool)             & 0.25 & 0.22 & 0.35 & 0.58 & 0.41 & 0.32 & 0.46 & 0.76 & 0.37 / 0.52 \\
Abstract + intro                  & 0.26 & 0.22 & 0.34 & 0.61 & 0.41 & 0.29 & 0.48 & 0.78 & 0.36 / 0.51 \\
Abstract + method                 & 0.24 & 0.22 & 0.32 & 0.61 & 0.41 & 0.28 & 0.47 & 0.78 & 0.33 / 0.53 \\
Abstract + bib                  & 0.26 & 0.23 & 0.34 & 0.61 & 0.41 & 0.30 & 0.48 & 0.79 & 0.36 / 0.52 \\
Abstract + intro + bib   & 0.25 & 0.23 & 0.34 & 0.60 & 0.42 & 0.30 & 0.49 & 0.77 & 0.36 / 0.53 \\
Full paper      & 0.25 & 0.23 & 0.34 & 0.60 & 0.42 & 0.29 & 0.49 & 0.78 & 0.35 / 0.51 \\
Sections chosen by the agent             & 0.26 & 0.23 & 0.35 & 0.60 & 0.42 & 0.30 & 0.50 & 0.77 & 0.36 / 0.53 \\
\bottomrule
\end{tabular}
}
\end{table}

\begin{figure}[t]
\centering
\includegraphics[width=0.5\textwidth]{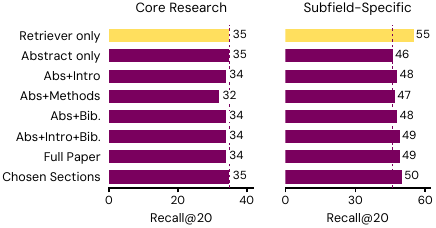}
\caption{\textbf{Reading corpus papers beyond their abstracts adds little.} Recall@20 of the GPT-4.1 tool-calling agent under six read policies and abstract-only reading, on 50 queries of each type from the full-text subset. The top bar is the retriever alone, the second the agent with no read tool.}
\label{fig:read_policies}
\end{figure}

\subsection{Analysis of Agent Search Failures and Rationale Explanations}
\label{app:failures}
We analyze \textit{why} agentic search fails to improve over retrieval, studying three possible causes in turn:
\begin{itemize}[leftmargin=*,itemsep=0.5pt,topsep=0pt]
\item \textbf{Coverage.} We measure how many gold papers agents find outside the retriever's candidates.
\item \textbf{Ranking.} We trace three search trajectories to see where gold papers are found and lost.
\item \textbf{Recognition.} We check whether agents explain the gold papers they rank highly the same way the authors do.
\end{itemize}

\textbf{Coverage: agents rarely find gold papers beyond the retriever's pool.}
\cref{tab:failure_2x2} splits gold papers by two outcomes, whether Gemini-Emb-2 retrieves them in its top 100 and whether each agent ranks them in its own top 20.
The retriever's top 100 covers 70\% of gold papers, and the agents recover almost none of the rest.
About half of gold papers are returned by no search during an agent's run.
The first search, which uses the query as written, accounts for about two-thirds of the gold papers an agent sees, and follow-up searches add few.
The grep agent, which writes its own search commands instead of using the retriever, keeps only 8\% of the gold papers the retriever covers.

\begin{table}[htbp]
\centering
\setlength{\tabcolsep}{4pt}
\renewcommand{\arraystretch}{1.1}
\caption{
\textbf{Agents rarely recover positives beyond the retriever's pool.}
Distribution of the 4,845 author-validated positives across 894 queries by whether Gemini-Emb-2 retrieves them in its top 100 and whether each agent ranks them in its top 20. Each of the first four columns reports the percentage of all positives in the corresponding retriever-agent outcome cell. The final two columns report the agent top-20 rate conditional on whether the retriever places the positive in its top 100.
}
\label{tab:failure_2x2}
\footnotesize
\resizebox{\columnwidth}{!}{
  \begin{tabular}{l cccc cc}
\toprule
& \multicolumn{4}{c}{\textbf{All positives (\%)}}
& \multicolumn{2}{c}{\textbf{Agent top-20 rate (\%)}} \\
\cmidrule(lr){2-5}\cmidrule(lr){6-7}
& \multicolumn{2}{c}{Retriever top 100}
& \multicolumn{2}{c}{Outside top 100}
& Retriever top 100
& Outside top 100 \\
Agent & Top 20 & Missed & Top 20 & Missed & Top 20 & Top 20 \\
\midrule
Tool-calling agent (GPT-4.1) & 40\% & 30\% & 1\% & 29\% & 57\% & 4\% \\
Deep research agent (o3) & 34\% & 36\% & 2\% & 28\% & 48\% & 6\% \\
Grep agent (GPT-4.1) & 6\% & 64\% & 1\% & 29\% & 8\% & 4\% \\
\bottomrule
\end{tabular}

}
\end{table}

\textbf{Ranking: agents drop many of the gold papers they do find.}
Among the gold papers in the retriever's top 100, the tool-calling and deep research agents each keep 57\% and 48\% in their top 20.
\cref{tab:fail_case_A,tab:fail_case_B,tab:fail_case_C} trace one agent's searches and final ranking on each of three \rqlevel{} queries, marking gold papers wherever a search returns them.
The grep agent retrieves no gold papers in any of the three.
\begin{enumerate}[label=\textbf{\optimisticfont Case \Alph*.},leftmargin=*,itemsep=0.5pt,topsep=2pt]
\item {\optimisticfont{Ranking failure.}}
The retriever ranks two gold papers first and fifth, and both agents retrieve them in their first search.
The tool-calling agent keeps only one in its ranking, and the deep research agent keeps neither.
\item {\optimisticfont{Follow-up recovery.}}
A follow-up search by both agents finds a gold paper that the retriever ranks beyond 100.
\item {\optimisticfont{Query drift.}}
The retriever's top 30 contains three of the four gold papers.
The agents' follow-up searches narrow to a subtopic and miss them, and the deep research agent leaves the one gold paper it does retrieve out of its ranking.
\end{enumerate}

\textbf{Recognition: agents usually explain gold papers the way the authors do.}
For the gold papers an agent does rank highly, we check whether its backbone understands why they matter.
For each of the 238 \rqlevel{} gold papers in the tool-calling agent's top 20, GPT-4.1 explains from the query, title, and abstract how the paper could help pursue the query, and a GPT-4.1 judge compares this explanation with the author's rationale.
Of the 219 pairs whose author rationale is long enough to compare, 76\% describe the same relationship, and almost all others describe a different but valid one (Table~\ref{tab:rationale_examples}).
Because the same model writes and judges, this agreement is an upper estimate.
The analysis also shows only that the backbone can explain a paper's relevance once given it, not that this understanding is why the paper entered the ranking.

\begin{figure}[p]
\centering
\scriptsize
\setlength{\tabcolsep}{3pt}

\renewcommand{\arraystretch}{0.80}
\captionof{table}{\optimisticfont{Case A. Ranking failure.}
The deep research agent (o3) retrieves two of the four gold papers in its first search, one of them again in its second, and lists neither in its final ranking. Each gold paper's position in the search result is in parentheses.}
\label{tab:fail_case_A}
\begingroup
\small
\setlength{\tabcolsep}{4pt}
\renewcommand{\arraystretch}{1.15}

\definecolor{trajectoryplum}{HTML}{79005E}
\definecolor{finalrankfill}{HTML}{FDF1FF}

\newcommand{\casequery}[1]{
  {\normalsize\textcolor{trajectoryplum}{\texttt{#1}}}
}
\newcommand{\casenone}{\textcolor{black!55}{none}}

\newcommand{\finalrank}[1]{
  \multicolumn{3}{@{}p{\linewidth}@{}}{
    \colorbox{finalrankfill}{
      \parbox{\dimexpr\linewidth-2\fboxsep\relax}{\raggedright #1}
    }
  } \\
}

\begin{tabularx}{\linewidth}{
  r
  >{\raggedright\arraybackslash}p{0.53\linewidth}
  >{\raggedright\arraybackslash}p{0.29\linewidth}
}
\toprule
\multicolumn{3}{p{\dimexpr\linewidth-2\tabcolsep\relax}}{
\raggedright
\textbf{\texttt{[[query]]}}
{\ttfamily CoT is the main tool used to tackle reasoning problems. There has been much work done investigating the theoretical capabilities of CoT in models from an expressive perspective\ldots{}}
} \\
\midrule

Call & Search query & Positive papers returned \\
\midrule

1
& \casequery{[[query]]}
& Lower Bounds for Chain-of-Thought\ldots{} (\#1);
  Let Me Think!\ldots{} (\#5) \\
2
& \casequery{multi-agent chain-of-thought language models communication complexity limits}
& Lower Bounds for Chain-of-Thought\ldots{} (\#2) \\
3
& \casequery{multi-agent debate large language models theory communication complexity}
& \casenone \\
\finalrank{
Calls 4--22 return no additional gold papers. The final ranking includes neither of the two gold papers found.
}
\bottomrule
\end{tabularx}
\endgroup

\vspace{3mm}

\renewcommand{\arraystretch}{0.95}
\captionof{table}{\optimisticfont{Case B. Follow-up recovery.}
The deep research agent (o3) finds a gold paper the retriever ranks beyond 100 with its second search and places it second in its final ranking. Each gold paper's position in the search result is in parentheses.}
\label{tab:fail_case_B}
\begingroup
\small
\setlength{\tabcolsep}{4pt}
\renewcommand{\arraystretch}{1.15}

\definecolor{trajectoryplum}{HTML}{79005E}
\definecolor{finalrankfill}{HTML}{FDF1FF}

\newcommand{\casequery}[1]{
  {\normalsize\textcolor{trajectoryplum}{\texttt{#1}}}
}
\newcommand{\casenone}{\textcolor{black!55}{none}}

\newcommand{\finalrank}[1]{
  \multicolumn{3}{@{}p{\linewidth}@{}}{
    \colorbox{finalrankfill}{
      \parbox{\dimexpr\linewidth-2\fboxsep\relax}{\raggedright #1}
    }
  } \\
}

\begin{tabularx}{\linewidth}{
  r
  >{\raggedright\arraybackslash}p{0.53\linewidth}
  >{\raggedright\arraybackslash}p{0.29\linewidth}
}
\toprule
\multicolumn{3}{p{\dimexpr\linewidth-2\tabcolsep\relax}}{
\raggedright
\textbf{\texttt{[[query]]}}
{\ttfamily Diffusion models are slow because they repeatedly perform denoising over many timesteps. Dynamic caching methods speed up this process by measuring the distance between intermediate features\ldots{}}
} \\
\midrule

Call & Search query & Positive papers returned \\
\midrule

1
& \casequery{[[query]]}
& Timestep Embedding Tells\ldots{} (\#9) \\
2
& \casequery{spectral evolution diffusion models low frequency first generative modeling}
& A Fourier Space Perspective on Diffusion Models (\#1) \\
\finalrank{
Calls 3--13 return no additional gold papers. The final ranking places \emph{A Fourier Space Perspective on Diffusion Models} second and \emph{Timestep Embedding Tells...} thirteenth.
}
\bottomrule
\end{tabularx}
\endgroup

\vspace{3mm}

\captionof{table}{\optimisticfont{Case C. Query drift.}
The retriever's top 30 holds three of the four gold papers, but the tool-calling agent (GPT-4.1) drifts to generic transfer learning and never sees any of them. Each gold paper's position in the search result is in parentheses.}
\label{tab:fail_case_C}
\begingroup
\small
\setlength{\tabcolsep}{4pt}
\renewcommand{\arraystretch}{1.15}

\definecolor{trajectoryplum}{HTML}{79005E}
\definecolor{finalrankfill}{HTML}{FDF1FF}

\newcommand{\casequery}[1]{
  {\normalsize\textcolor{trajectoryplum}{\texttt{#1}}}
}
\newcommand{\casenone}{\textcolor{black!55}{none}}

\newcommand{\finalrank}[1]{
  \multicolumn{3}{@{}p{\linewidth}@{}}{
    \colorbox{finalrankfill}{
      \parbox{\dimexpr\linewidth-2\fboxsep\relax}{\raggedright #1}
    }
  } \\
}

\begin{tabularx}{\linewidth}{
  r
  >{\raggedright\arraybackslash}p{0.53\linewidth}
  >{\raggedright\arraybackslash}p{0.29\linewidth}
}
\toprule
\multicolumn{3}{p{\dimexpr\linewidth-2\tabcolsep\relax}}{
\raggedright
\textbf{\texttt{[[query]]}}
{\ttfamily LLM agents struggle to put past experience to use in complex scenarios. There has been a lot of recent work along similar lines that jointly optimizes memory or reasoning banks together with the model weight\ldots{}}
} \\
\midrule

Call & Search query & Positive papers returned \\
\midrule

1
& \casequery{[[query]]}
& \casenone \\
2
& \casequery{meta-learning approaches for generalization in LLM agents}
& \casenone \\
3
& \casequery{transfer learning in reinforcement learning agents with memory}
& \casenone \\
4
& \casequery{representation learning for transferable skills in AI agents}
& \casenone \\
5
& \casequery{case-based reasoning and analogical transfer in AI agents}
& \casenone \\
\finalrank{
Call 6 returns no gold papers. The final ranking contains no gold papers.
}
\bottomrule
\end{tabularx}
\endgroup

\end{figure}

\begin{table}[htbp]
\centering
\normalsize
\setlength{\tabcolsep}{3pt}
\renewcommand{\arraystretch}{1.05}
\caption{
\textbf{Agents usually explain a positive paper the way its author does.}
For each \rqlevel{} positive in the tool-calling agent's top 20, GPT-4.1 explains how the paper could help pursue the query, and a GPT-4.1 judge compares that explanation with the author's rationale.
Of the 219 comparable pairs, 76\% describe the same relationship, and almost all others give a different relationship that is still valid.
Two examples below are judged the same and one different but valid.
The judge's explanation is in gray.
Bolding within the rationales is ours.
}
\label{tab:rationale_examples}
\definecolor{plum}{HTML}{79005E}
\newcommand{\judge}[1]{
  {\normalsize\textcolor{plum}{\texttt{#1}}}
}

\begin{tabularx}{\linewidth}{p{0.18\linewidth} X X}
\toprule
Verdict & Author's rationale & Backbone's rationale (GPT-4.1) \\
\midrule
\multicolumn{3}{p{\dimexpr\linewidth-2\tabcolsep\relax}}{{TTRL: Test-Time Reinforcement Learning}} \\
\midrule
Same relationship &
This is the most significant paper that impacted my initial experiments. I was following their setup and used their code to run test-time training with unsupervised labels, but I found the \textbf{correctness of the labels seems to not matter} too much on the Qwen-Math models they used. [\ldots{}] &
The TTRL paper’s main contribution is the demonstration that reinforcement learning can be effectively performed on reasoning tasks in large language models \textbf{using only unlabeled data}, by leveraging majority-voted answers as a proxy for ground-truth rewards. [\ldots{}] \\
\multicolumn{3}{p{\dimexpr\linewidth-2\tabcolsep\relax}}{\judge{Judge: Both rationales identify that TTRL demonstrates the effectiveness of using majority-voted, potentially noisy, unsupervised labels as rewards for reinforcement learning in LLMs, directly informing the author's [\ldots{}]}} \\
\midrule
\multicolumn{3}{p{\dimexpr\linewidth-2\tabcolsep\relax}}{{Consent in Crisis: The Rapid Decline of the AI Data Commons}} \\
\midrule
Same relationship &
The paper did a large-scale, \textbf{longitudinal audit} of the consent protocols for the web domains underlying AI training corpora, and pointed out there is a decline in data owners' willingness to contribute data to AI training. This paper made an important observation [\ldots{}] &
The key contribution of ``Consent in Crisis: The Rapid Decline of the AI Data Commons'' is its rigorous, \textbf{longitudinal audit} quantifying how quickly and extensively web data sources are restricting AI training use, particularly [\ldots{}] \\
\multicolumn{3}{p{\dimexpr\linewidth-2\tabcolsep\relax}}{\judge{Judge: Both rationales identify the paper's empirical audit of declining data consent as foundational for understanding the real-world consequences of respecting opt-outs on LLM training, emphasizing the importance of the [\ldots{}]}} \\
\midrule
\multicolumn{3}{p{\dimexpr\linewidth-2\tabcolsep\relax}}{{DOME: Taming Diffusion Model into High-Fidelity Controllable Occupancy World Model}} \\
\midrule
Different, valid &
One of the existing sota methods, with high reconstruction accuracy but \textbf{severe degradation of future prediction} performance, inspired us to think about the limitations of autoregressive correlation design. &
The DOME paper’s primary contribution is the introduction of a \textbf{spatial-temporal diffusion transformer} that directly models 3D occupancy evolution over time, bypassing the limitations of discrete tokenization and rigid geometric priors [\ldots{}] \\
\multicolumn{3}{p{\dimexpr\linewidth-2\tabcolsep\relax}}{\judge{Judge: The author focuses on DOME's limitations in autoregressive design for future prediction, while the model highlights DOME's diffusion-based architecture and its advantages over tokenization and priors, both valid but [\ldots{}]}} \\
\bottomrule
\end{tabularx}

\end{table}

\subsection{Cost and Search Effort of the Agents}
\label{app:agent-cost}
Table~\ref{tab:agent_cost} reports the search effort, API cost and time per query of each agent in Table~\ref{tab:main_results}.
With GPT-4.1 and o3, the grep and deep research agents carry a growing context through 14 to 20 tool calls per query and cost about eight times as much as the tool-calling agent, which makes few searches and scores each pooled paper in a short, separate call.
With Claude Fable 5.1 the tool-calling agent issues more than twice as many searches within the same five rounds, and every agent costs more than a dollar per query.
Times are medians over queries run in parallel and include API latency and retries, so they vary with provider load.

\begin{table}[h]
\centering
\small
\setlength{\tabcolsep}{4pt}
\caption{\textbf{Search effort, cost, and latency of agent baselines.}
All numerical columns except total cost report per-query values.
Tool and backbone calls are means, while time is the median.
Total cost is for all 894 queries.
Costs use list prices and include prompt caching for Claude Fable 5.1.
A tool call is a search, read, or shell command.
Backbone calls include the tool-calling agent's per-paper scoring calls.}
\label{tab:agent_cost}
\resizebox{\linewidth}{!}{
\begin{tabular}{llrrrrr}
\toprule
Agent & Backbone & Tool calls & Backbone calls
& Cost (\$) & Total cost (\$) & Median time (s) \\
\midrule
Grep & GPT-4.1 & 20.3 & 18.5 & 0.37 & 332 & 22 \\
Tool-calling & GPT-4.1 & 5.8 & 44.3 & 0.05 & 42 & 25 \\
Deep research & o3 & 14.3 & 18.3 & 0.39 & 345 & 60 \\
\midrule
Grep & Claude Fable 5.1 & 19.3 & 20.2 & 1.14 & 1{,}016 & 139 \\
Tool-calling & Claude Fable 5.1 & 13.9 & 63.2 & 1.32 & 1{,}178 & 88 \\
Deep research & Claude Fable 5.1 & 22.7 & 12.4 & 1.39 & 1{,}240 & 146 \\
\bottomrule
\end{tabular}}
\end{table}

\section{Automated Data Construction Pipeline}
\label{app:reconstruction}
We construct \benchmark{} through an end-to-end pipeline that transforms raw arXiv papers into benchmark instances for inspiration literature retrieval. Starting from recent computer science papers, the pipeline constructs a collection of source papers, reconstructs benchmark annotations, assembles a retrieval corpus with a time cutoff, and augments benchmark instances through candidate retrieval and validation. Figure~\ref{fig:benchmark_construction} summarizes the overall construction pipeline.

Existing publicly available scientific corpus~\citep{lo2020s2orc}
and kaggle arXiv datasets\footnote{\url{https://www.kaggle.com/datasets/sumitm004/arxiv-scientific-research-papers-dataset}}
do not provide the combination of recency, full-text representation, bibliography information, and citation relations required by our setting.
We therefore build directly from raw arXiv data, processing each paper into a structured representation with explicit citation links that supports all subsequent stages of benchmark construction.

\subsection{Source-paper Processing}
\label{app:source-processing}
We enumerate candidate source papers from recent oral and spotlight lists for recent AI conferences, match each accepted paper to its arXiv preprint by title, and exclude surveys and review articles.
For each paper, we obtain a structured representation from the arXiv HTML rendering when available and otherwise by converting the corresponding \LaTeX{} source using LaTeXML\footnote{LaTeXML project: \url{https://math.nist.gov/~BMiller/LaTeXML/}}, which better preserves document structure and citation information than PDF-based processing~\citep{frankston2024html}.
Because some papers cannot be reliably converted due to source formatting or parsing failures, we apply quality filtering to remove papers with incomplete text or citation structure.
The resulting full text is converted into a structured representation in which inline citations are replaced with explicit citation markers linked to bibliography entries, providing a citation-grounded representation for all downstream stages.

We resolve every cited reference to a canonical scholarly record using the arXiv\footnote{\url{https://arxiv.org}}, Semantic Scholar\footnote{\url{https://www.semanticscholar.org/product/api}}, and OpenAlex~\citep{priem2022openalex} APIs based on the bibliography text.
Citation parsing, inconsistent bibliography formatting, and search engine limitations occasionally prevent reliable resolution of the intended paper.
We retain only source papers for which at least 70\% of cited references are successfully resolved, and discard the rest before they proceed to the remaining pipeline stages, ensuring sufficient citation coverage for annotation reconstruction.

\subsection{Pre-annotation Generation}
\label{app:initial-draft}
From the structured source paper, we automatically reconstruct the benchmark instances. The model (Gemini 3.1 pro) reads the full paper together with its bibliography to recover the motivating research question (query $q$, one \CQ{} and several \SQ{}s), inspiration papers (positive papers $D_q^+$) for each query. It further generates detailed rationales explaining how each selected paper contributed to the resulting research. These annotations form the benchmark instances used throughout the evaluation.
This reconstruction uses the prompts provided in Appendix~\ref{app:reconstruction-prompts}.

\subsection{Candidate Retrieval}
\label{app:candidate-retrieval}
We augment the benchmark beyond the paper's cited bibliography by recovering insightful papers that were not cited.
For each retrieval query, we retrieve the top-15 candidate papers from the corpus using both sparse (BM25~\citep{robertson2009probabilistic}) and dense (Qwen3 8B embeddings~\citep{zhang2025qwen3}) retrieval systems, deduplicate them, and rerank with a reasoning model (Gemini-3.6-flash) that scores each candidate for scientific relevance to the query rather than mere topical similarity.
We retain the top 10 for author review, where authors mark inspiring papers as positives and closely related but non-inspiring papers as hard negatives.

\textbf{Retrieval corpus.}
We construct a retrieval corpus containing 190,896 papers. The corpus includes resolved references cited by our source papers together with a pool of 181K additional arXiv papers published between 2020 and 2024, drawn from five main computer science domains (cs.CL, cs.CV, cs.LG, cs.RO, and cs.AI), each restricted to papers with at least 1 citation. To construct a realistic literature pool beyond the cited literature, we estimate the citation distribution of recent ICML, ICLR, and NeurIPS oral and highlight papers and use it to determine the publication-year and category distributions of additional corpus papers. At retrieval time, we additionally apply a per-query temporal cutoff based on the query's source-paper publication date, restricting candidate documents to those published no later than the query paper itself. This precludes temporal leakage, ensuring that no candidate postdates the query paper it is retrieved for.

The prompts used in the pipeline are in \cref{app:reconstruction-prompts}.

\subsection{Author Validation of the Automated Pipeline}
\label{appendix:pipeline-validation}
\textbf{Evaluation against author judgments.}
The final author-written set results from authors validating and revising the pipeline's reconstructions, enabling us to measure how closely the pipeline's initial outputs align with author judgments.
In total, 184 researchers completed the validation form for 207 source-paper instances, together reviewing 687 subfields. Table~\ref{tab:author_validation} summarizes the validation results across all annotation stages.

Authors rated the reconstructed \CQ{} as accurate or mostly accurate for 98.1\% of the 207 instances (60.9\% fully accurate). The same holds for individual subfields: across all 687 subfields, the reconstructed \SQ{} was rated accurate or mostly accurate 95.9\% of the time (64.9\% fully accurate). The pipeline's reconstructed questions thus capture what authors recall motivating their work, both for a paper as a whole and for each of its research directions.

For \SQ{}s, the candidate step shows the value of pooling.
Authors retained 85.1\% of originally cited candidates as positives, making citations a high-yield starting point.
More importantly, authors judged 35.1\% of pooled, non-cited candidates---papers absent from the source paper's bibliography---as genuine inspirations.

Finally, we evaluate the pipeline's key-inspiration nomination for \CQ{}s.
For each paper's \CQ{}, the pipeline selects cited papers as likely key inspirations.
We compare these nominations with the cited papers that the author ultimately labels positive.

\textbf{Model-generated dataset.}
Given this fidelity, we apply the pipeline to 250 additional papers to form the model-reconstructed set, yielding 250 central queries and 863 directional queries.

\begin{table}[h]
\centering
\small
\caption{Author validation results, computed on the final 207-paper, 687-\threadlevel{} benchmark. \emph{Reconstructed query quality} is rated per \CQ{} or \SQ{}. \emph{Candidate pool judgement} is rated per individual candidate document the author judged for an \SQ{}. \emph{Key-inspiration selection vs.\ author} is evaluated only for \CQ{}s, treating the pipeline's per-paper nomination as a binary prediction over every cited candidate the author judged for that paper.}
\label{tab:author_validation}
\begin{tabular}{llr}
\toprule
Category & Metric & Value \\
\midrule
\multirow{4}{*}{\parbox{2.6cm}{Reconstructed query quality}}
  & \CQ{} rated fully accurate & 60.9\% \\
  & \CQ{} rated accurate or mostly accurate & 98.1\% \\
  & \SQ{} rated fully accurate & 64.9\% \\
  & \SQ{} rated accurate or mostly accurate & 95.9\% \\
\midrule
\multirow{2}{*}{\parbox{2.6cm}{Candidate pool\\ judgement}}
  & Original citation retained as positive & 85.1\% \\
  & Pooled candidate promoted to positive & 35.1\% \\
\midrule
\multirow{3}{*}{\parbox{2.6cm}{Key-inspiration\\selection vs.\ author}}
  & Precision & 58.2\% \\
  & Recall & 26.0\% \\
  & Pipeline's full nominated set exactly equals the author's & 9.2\% \\
\bottomrule
\end{tabular}
\end{table}

\section{Author Data Collection}
\label{app:author-data}

\subsection{Author Recruitment}
\label{app:recruitment}
\begin{figure}[t]
\centering
\includegraphics[width=\linewidth]{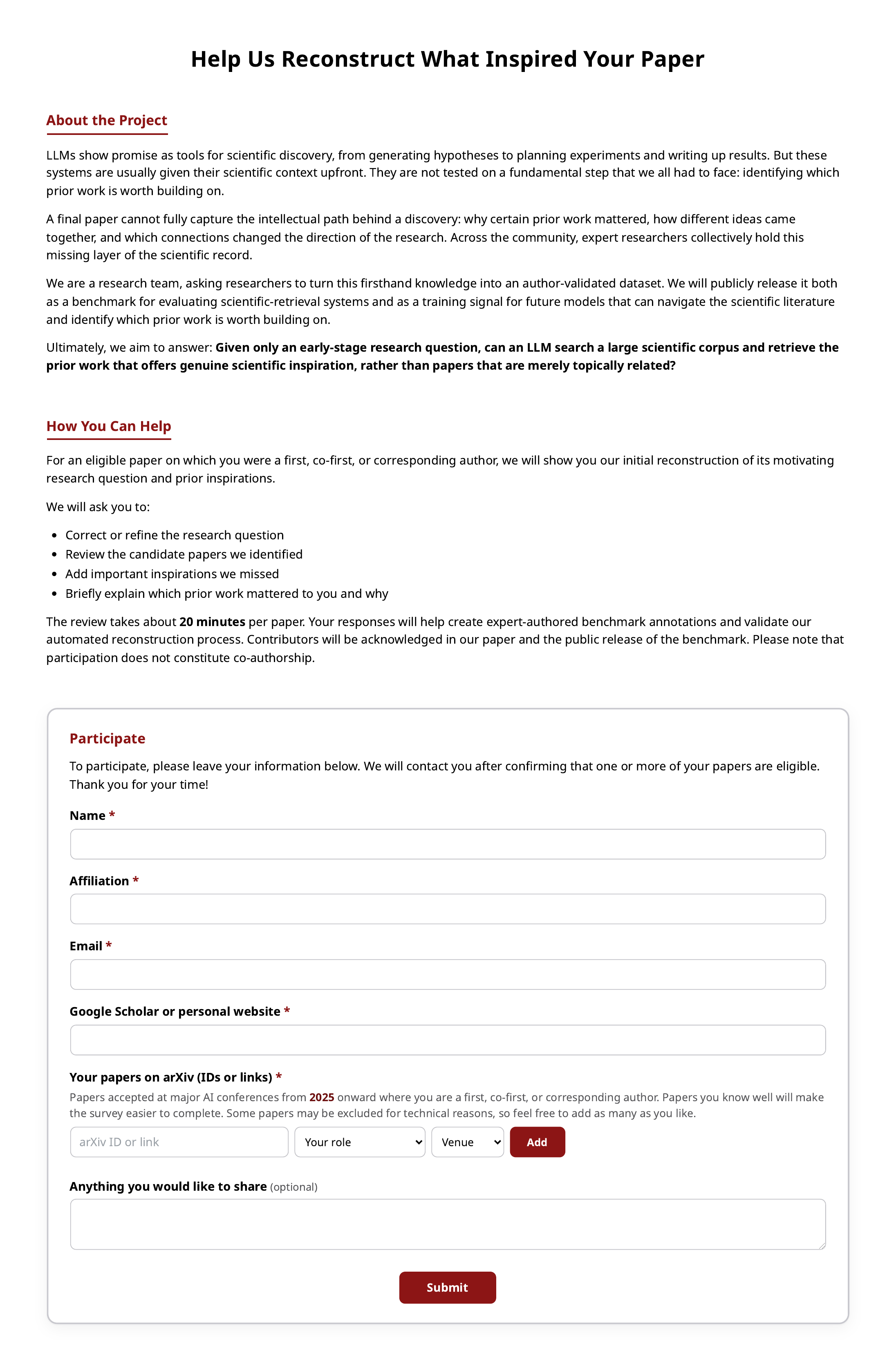}
\caption{Recruitment page shown to participants, where eligible authors submit their arXiv papers accepted at major AI conferences from 2025 onward.}
\label{fig:recruitment_page}
\end{figure}

\begin{figure}[p]
\centering
\includegraphics[height=0.96\textheight]{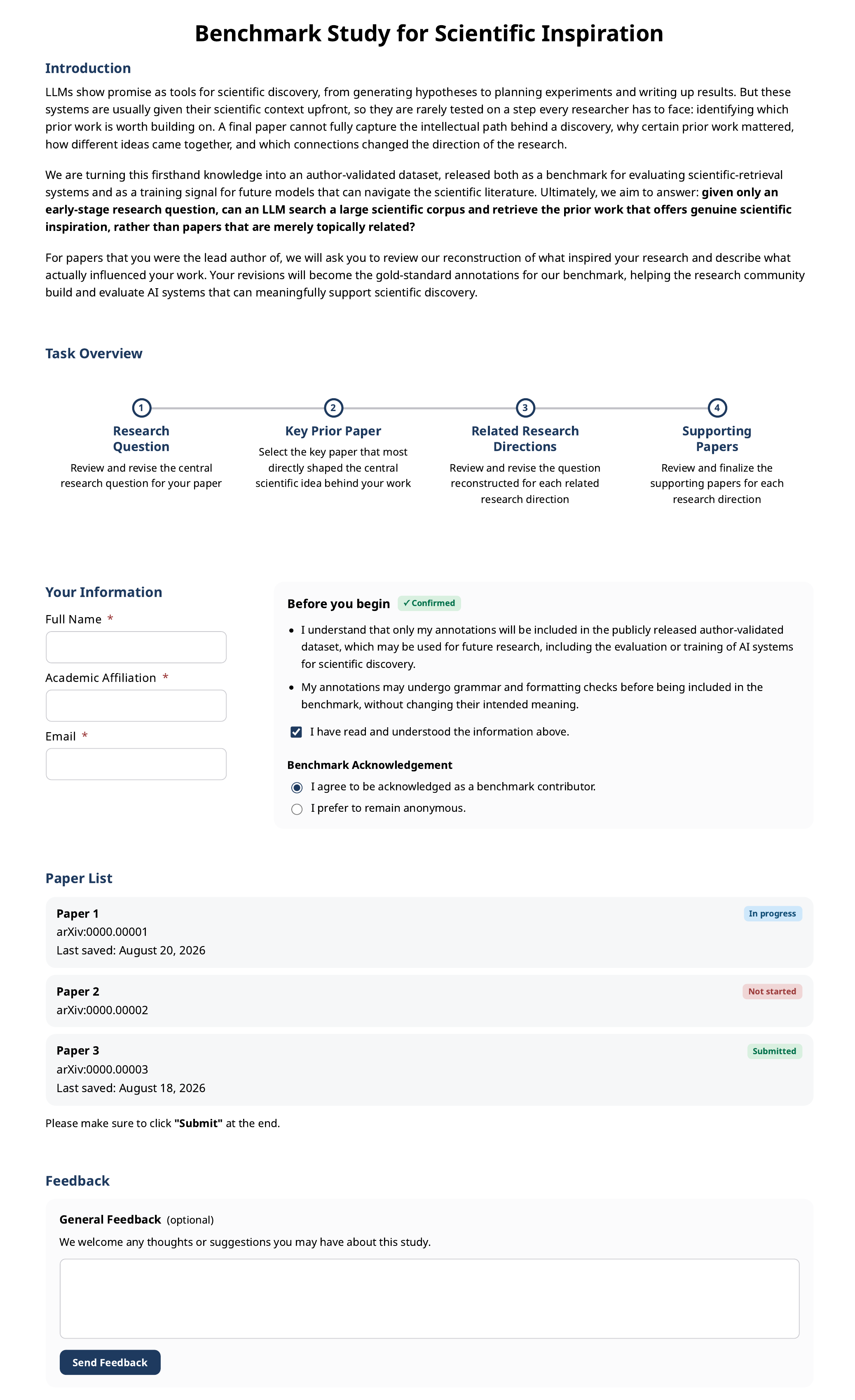}
\caption{The web-based annotation interface presented to authors.}
\label{fig:annotation_interface}
\end{figure}

We recruit first-author, co-first-author, and corresponding authors of highlighted papers (oral, spotlight, and award) at recent iterations of ICLR, ICML, NeurIPS, CVPR, CoLM, CoRL, and ACL, together with institution-wide public recruitment open to any first, co-first, or corresponding author of a paper accepted at a major AI conference from 2025 onward.
Participants were invited through direct email and public calls. The recruitment page in Figure~\ref{fig:recruitment_page}, and the annotation interface in Figure~\ref{fig:annotation_interface}.
In total, 184 authors contributed 207 author-written instances.

\subsection{Annotation Interface}
\label{app:annotation-interface}
We developed a web-based interface through which authors annotate their own research projects.
For each source paper, authors write or revise a research question intended to reflect their pre-solution information state.
They organize the question into distinct research subfields and identify prior papers that provided meaningful scientific inspiration for each direction.
Authors then review additional candidate papers, labeling papers that provided meaningful inspiration as positives and plausible but non-inspirational papers as hard negatives.
For each judgment, they provide a rationale explaining how the paper did or did not contribute to the development of the target work.
The resulting author-written instances contain research questions, positive and hard-negative labels, and open-ended rationales.

\begin{figure}[p]
\centering
\includegraphics[width=\linewidth]{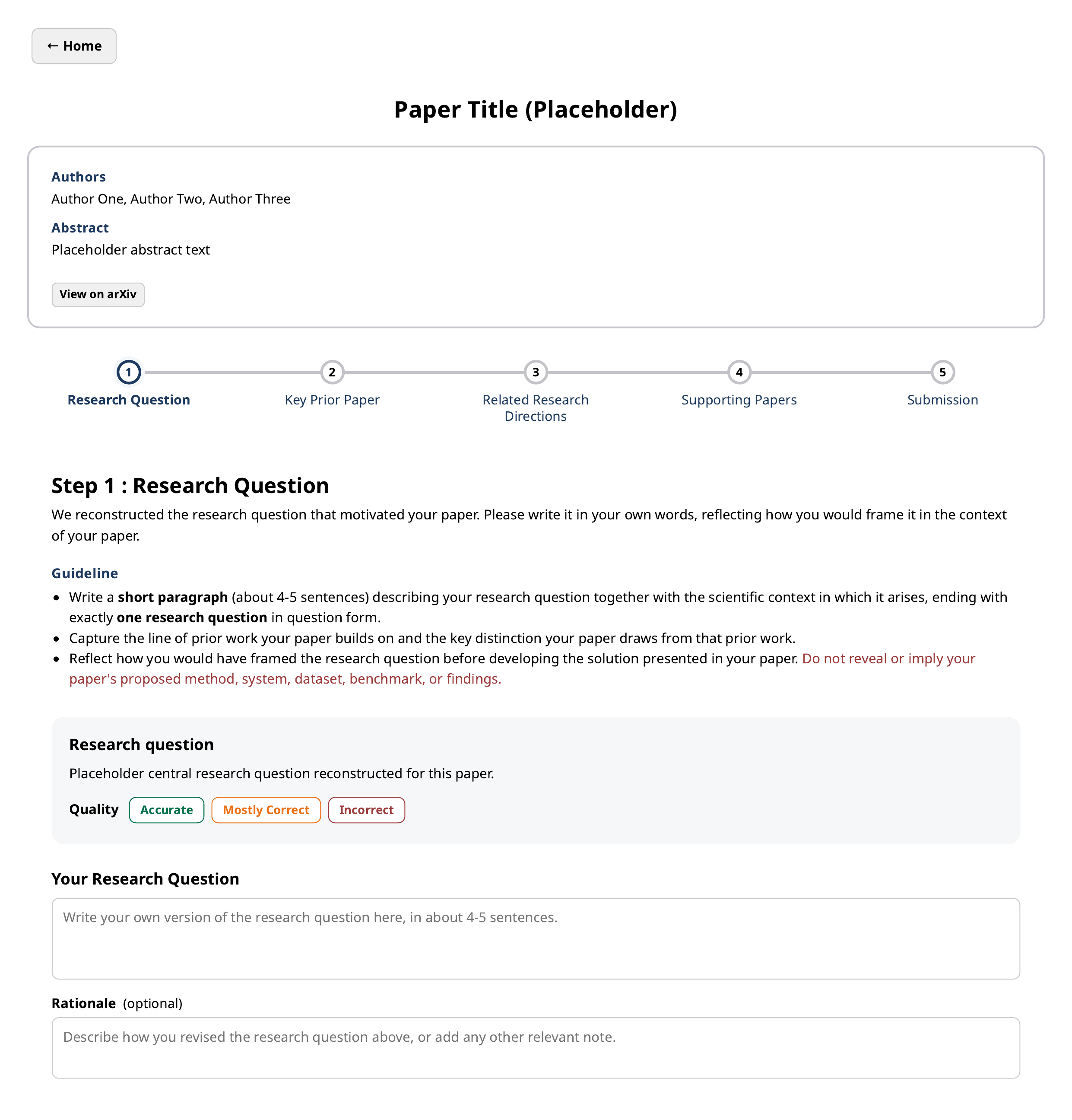}
\caption{Author annotation, Step 1 (Research Question). Authors rate our reconstructed research question, rewrite it in their own words to reflect their pre-solution framing, and optionally add a rationale.}
\label{fig:stage_1}
\end{figure}

\begin{figure}[p]
\centering
\includegraphics[width=\linewidth]{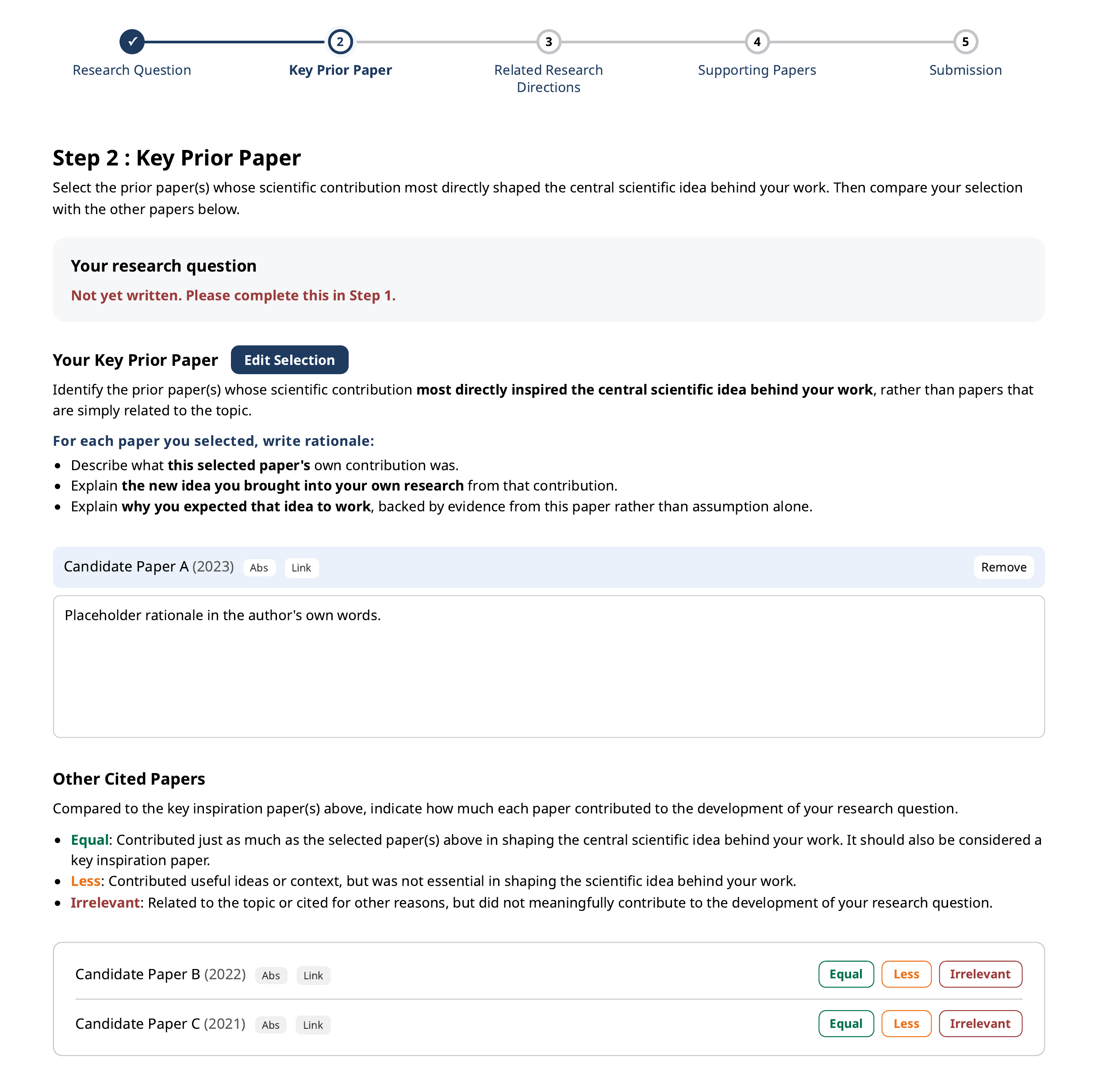}
\caption{Author annotation, Step 2 (Key Prior Paper). Authors select the prior paper(s) whose contribution most directly shaped the central idea, write a rationale for each, and mark every other cited paper as an equal, lesser, or irrelevant influence.}
\label{fig:stage_2}
\end{figure}

\begin{figure}[p]
\centering
\includegraphics[width=\linewidth]{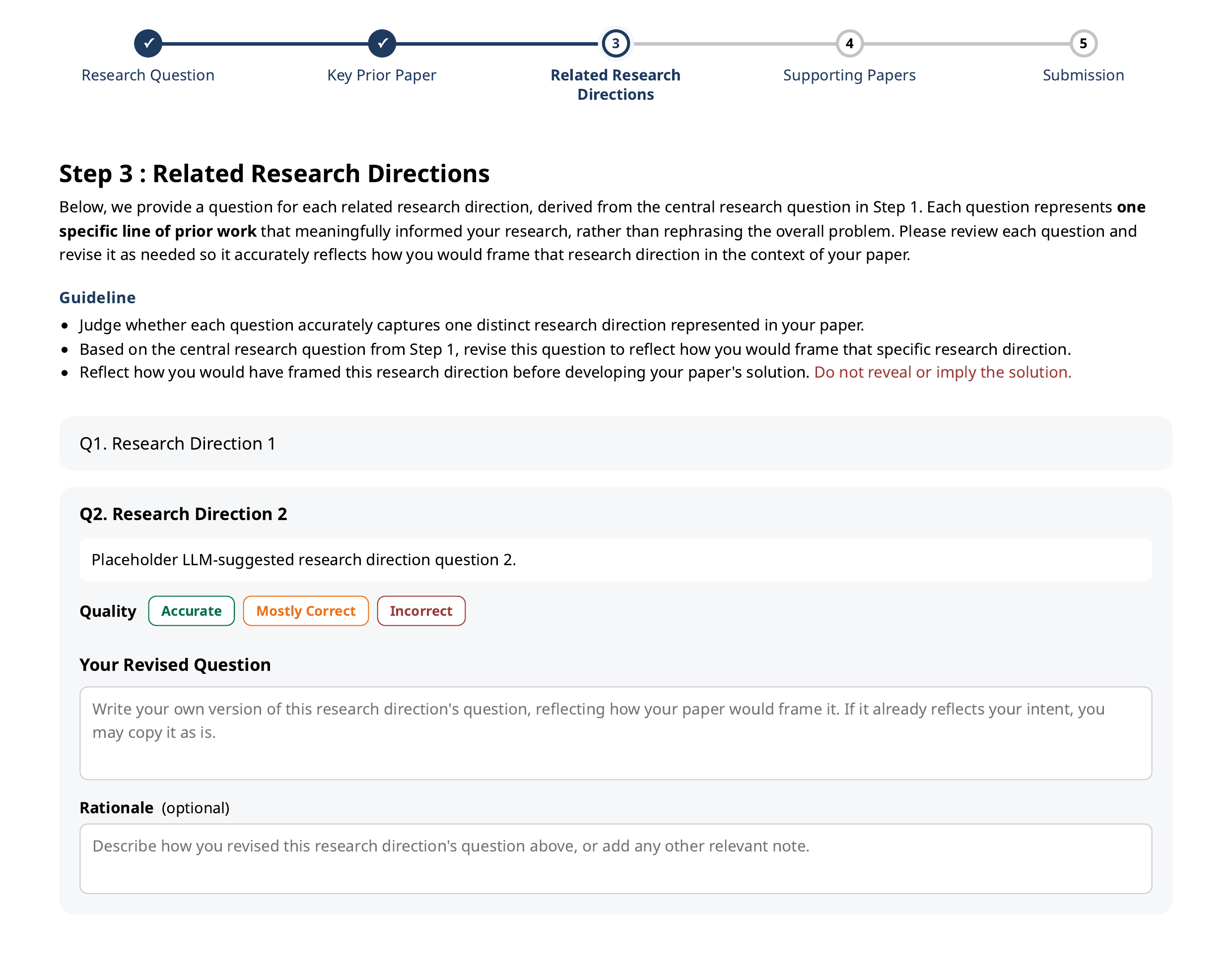}
\caption{Author annotation, Step 3 (Related Research Directions). For each reconstructed research direction, authors rate its question, revise it to reflect how their paper frames that direction, and optionally add a rationale.}
\label{fig:stage_3}
\end{figure}

\begin{figure}[p]
\centering
\includegraphics[width=\linewidth]{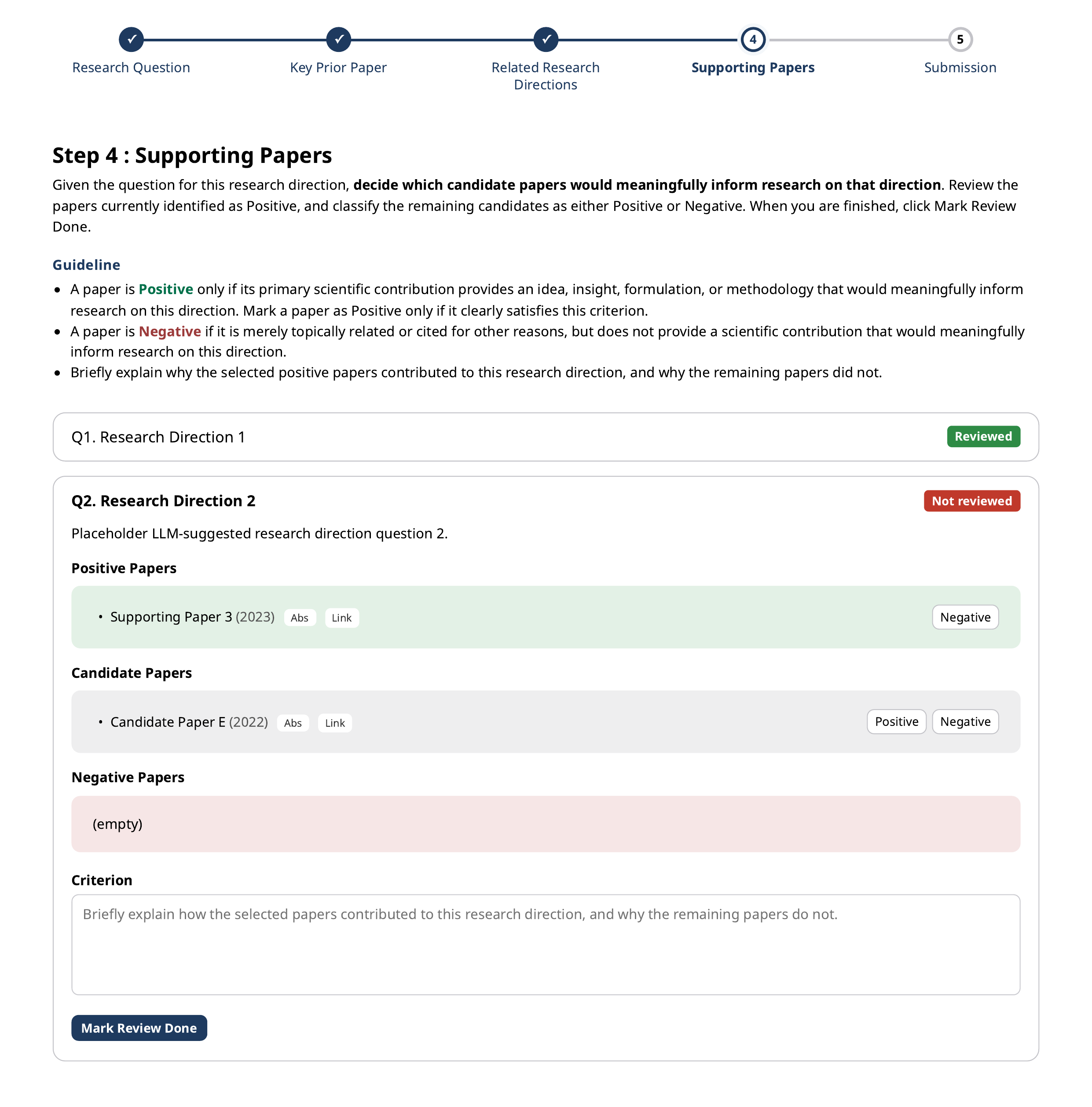}
\caption{Author annotation, Step 4 (Supporting Papers). For each research direction, authors classify candidate papers as positive or negative according to whether the paper's contribution would meaningfully inform that direction, with a brief justification.}
\label{fig:stage_4}
\end{figure}

\section{Prompts}
\label{app:prompts}

\subsection{Agent Prompts}
\label{app:agent-prompts}

\begin{plumpromptbox}{Prompt for the Tool-Calling Agent Search Loop}
You are a research agent. Given an early-stage research question, find prior papers whose ideas could genuinely help a researcher pursue it, not papers that are merely on the same topic. You have at most 5 rounds of tool calls. Call search with different angles of the question. Do not assume or reveal a solution to the question. When you have enough candidates, reply with the single word DONE.
\end{plumpromptbox}

\begin{plumpromptbox}{Prompt for the Tool-Calling Agent Candidate Judge}
Research question:
<research question>

Candidate paper:
Title: <title>
Abstract: <abstract>

Would this paper's ideas meaningfully help a researcher pursue the research question, beyond being topically related? Reply with one integer from 0 (no) to 10 (certainly), nothing else.
\end{plumpromptbox}

\begin{plumpromptbox}{Prompt for the Deep Research Agent Search Loop}
You are a research agent. Given an early-stage research question, find prior papers whose ideas could genuinely help a researcher pursue it, not papers that are merely on the same topic. Work through the sub-questions of your plan: call search with different angles (20 searches at most) and read on promising papers to see their full abstract (10 reads at most). Do not assume or reveal a solution to the question. When the plan is covered, reply without calling a tool.
\end{plumpromptbox}

\begin{plumpromptbox}{Prompt for Deep Research Planning}
Research question:
<research question>

Split this question into at most 6 sub-questions a literature search should answer to find prior work whose ideas could help pursue it. Reply with a JSON array of strings and nothing else.
\end{plumpromptbox}

\begin{plumpromptbox}{Prompt for Deep Research Reflection}
Reflection. The plan was:
<the plan, one sub-question per line>

Which sub-questions are still uncovered by the papers found so far, and what will you search or read next? Reply in a few sentences, without calling a tool.

Continue with the searches and reads your reflection names.
\end{plumpromptbox}

\begin{plumpromptbox}{Prompt for Deep Research Synthesis}
Papers found (<pool size>):
[<id>] <title>
...

Research question:
<research question>

From these papers only, pick the ones whose ideas would meaningfully help a researcher pursue the question, most useful first, at most 50. Reply with a JSON array of their ids and nothing else.
\end{plumpromptbox}

\begin{plumpromptbox}{Prompt for the Grep Agent}
You are a research agent searching a local paper corpus for prior work whose ideas could genuinely help a researcher pursue an early-stage research question, not papers that are merely on the same topic.
The current directory holds the corpus as files named YYYY-MM.jsonl, one paper per line as JSON with the fields id, title, text (the abstract) and published; only papers the researcher could have read are present. Search the abstracts, not only the titles. Use bash with ripgrep, for example
  rg -i 'phrase one|phrase two' *.jsonl | head -c 4000        # matching papers with id, title and abstract
  rg -i -c 'phrase' *.jsonl | sort -t: -k2 -nr | head          # which months have the most matches
  rg -i -o '"id": "[^"]+", "title": "[^"]{0,120}' 2023-*.jsonl | rg -i 'phrase' | head -30   # compact id + title list
Try 5 to 10 different angles, including specific technical terms, method names and problem formulations that a relevant paper would use even if the question does not. Read the abstracts of at least 20 candidates before answering; a paper counts only if you have seen its abstract. You have at most 20 commands. Do not assume or reveal a solution to the question. When done, reply with a JSON array of paper ids, most useful first, and nothing else.
\end{plumpromptbox}

\subsection{Reranking and Query Generation Prompts}
\label{app:analysis-prompts}

\begin{plumpromptbox}{Prompt for Listwise and Tournament Reranking}
[System]
You are an expert at scientific literature retrieval.
Your goal is to identify prior papers that could help a researcher pursue the given research question.
Given a research question and a list of candidate papers, rank ALL papers from most to least likely to be useful for advancing this research question.
Rank a paper higher when its scientific contribution would meaningfully inform how a researcher thinks about or approaches this question.
Rank it lower when it is merely topically related and does not provide insight for this research.
Output only a JSON array of paper IDs in ranked order, most relevant first.
Example: ["id_a", "id_b", "id_c"]

[User]
Query: <research question>

Papers:
id=<paper_id_1> title=<title_1>
abstract=<abstract_1>

id=<paper_id_2> title=<title_2>
abstract=<abstract_2>

...
\end{plumpromptbox}

\begin{plumpromptbox}{Prompt for Single-Query Expansion}
You are helping a researcher find prior scientific papers that could help them pursue an early-stage research question. Augment the researcher's question to make retrieval more effective. Keep the original question text unchanged, and add a short continuation that makes the search more specific, using concrete terms that prior work in this area would use. Do not change or remove any part of the original question. Do not introduce a different research direction or a solution. Write natural prose only, no keyword lists. Output the original question followed by the continuation.
\end{plumpromptbox}

\begin{plumpromptbox}{Prompt for Multi-Query Generation}
You are helping a researcher find prior scientific papers that could help them pursue an early-stage research question. Generate up to 5 augmented versions of the researcher's question. Each version must keep the original question text unchanged and add a short continuation that makes the search more specific. Make the continuations diverse: each should point toward a distinct plausible line of prior work that could inform the same research situation. Do not introduce a different research direction or a solution. Write natural prose only, no keyword lists. Output one augmented question per line and nothing else.
\end{plumpromptbox}

\begin{plumpromptbox}{Prompt for HyDE Hypothetical Abstracts}
You are helping a researcher find the prior work that inspired a new research idea.

Below is a research question written by the authors of a paper before they solved it. Your job is NOT to answer the question. Your job is to imagine the earlier papers that the authors might have read and drawn on when forming this question.

Write 5 hypothetical abstracts of such prior papers. Each abstract should:
- Read like a real paper abstract (title + 120-180 words): problem, method, key finding.
- Describe a paper that already existed before the question was written, so it must not solve the question itself. It contributes one idea, tool, observation, or framing that the authors could have borrowed.
- Come from a different angle than the others. Cover at least one adjacent field or task where the same underlying idea appears under different terminology.
- Use the vocabulary that paper's own community would use, not the vocabulary of the question.

Do not name real papers, authors, or years. Do not mention the research question or the target paper. Output only the abstracts.

Output format (JSON):
[
  {"title": "...", "abstract": "..."},
  ...
]
\end{plumpromptbox}

\subsection{Pre-annotation Construction Prompts}
\label{app:reconstruction-prompts}
This section details the exact prompts used in our pre-annotation construction pipeline. We show the prompt for reconstructing the research question and key inspiration papers, for identifying meaningful subfield direction, and for extracting key citations from each subfield.

\bigskip
\begin{plumpromptbox}{Prompt for Research Question and Key Inspiration Reconstruction}
You are an expert research analyst.

Given a research paper and its bibliography, your task is to reconstruct the
research question the authors were pursuing before they developed the solution
presented in the paper, and identify the cited prior paper(s) whose contribution
most directly inspired the target paper's central contribution.

<paper>
{paper_text}
</paper>

<bibliography>
{bibliography}
</bibliography>

Each citation in the paper is marked inline as `[CITE:<key>]` and listed in the
bibliography above. Use the exact `<key>` string when referring to citations in
your output.

**Task:**
1) Identify the line of research this paper belongs to and reconstruct the
research context that led to the authors' research question. Describe this from
the authors' own perspective rather than as a neutral outside summary. Use only
information that would have been available before this work was conducted.
2) Write the research question the authors would have asked from this context,
before they knew how to solve it. Briefly describe only the research context that
naturally led to the research question, then end with exactly one central
research question.
3) Given this research question and the target paper, identify the cited prior
paper(s) whose contribution most directly inspired the target paper's central
contribution. A key inspiration paper is not simply the most relevant,
representative, or highly cited prior work. Instead, it is the prior work whose
contribution most directly inspired the target paper's central contribution.

**Research Question Requirements:**
- Reconstruct only the research context available before the work was conducted.
- Reflect the authors' research perspective rather than a broad field-level
question. Briefly describe only the scientific problem and its limitations that
naturally led to the research question.
- End with exactly one research question in question form.
**Avoid:**
- Revealing or implying the paper's eventual answer, solution, proposed paradigm,
or high-level approach.
- Using the paper's own methods, systems, datasets, benchmarks, findings, or
other solution-specific concepts.
- Using terminology, names, or phrases introduced by the target paper itself.

**Key Inspiration Requirements:**
**Selection Criteria:**
- Select the cited prior paper(s) using the exact bibliography key(s).
- Select only the cited paper(s) whose contribution most directly inspired the
target paper's central contribution.
**Rationale Requirements:**
- Briefly explain the selected paper's main contribution.
- Explain what idea from this work informed the target paper's central
contribution and how it influenced that contribution.
- Explain why this work should be regarded as a primary source of inspiration.
- Focus on explaining the scientific connection between the selected work and the
target paper, rather than summarizing either paper independently.
**Avoid:**
- Selecting papers simply because they are representative, foundational, or highly
cited unless they directly inspired the target paper's central contribution.
- Referring to the selected work by its citation key, title, or author names.
Instead, describe it through its contribution and scientific role.

**CRITICAL:**
- Every `cite_key` must exactly match a key listed in the bibliography.
- NEVER invent, modify, shorten, normalize, paraphrase, or partially copy a
`cite_key`.
- Write as continuous natural language rather than keywords, concept lists, paper
titles, or multiple questions.
- Prefer clear, direct sentences. Do not use em dashes or semicolons.

**Return a JSON object:**
{
  "research_question": "a short paragraph describing the research context that
led to the research question, ending with exactly one research question",
  "key_inspiration": [
    {
      "cite_key": "the exact bibliography key",
      "rationale": "a dense paragraph explaining the selected paper's main
contribution, what idea from this work informed the target paper's central
contribution, how this idea influenced that contribution, and why this work
should be regarded as a primary source of inspiration."
    }
  ]
}

Return ONLY the JSON object, nothing else.
\end{plumpromptbox}

\clearpage
\begin{plumpromptbox}{Prompt for Identifying Key  Subfields}
You are an expert research analyst. Given a research paper, identify the distinct
inspiration threads underlying the paper and formulate a retrieval query for each
identified thread.

<paper>
{paper_text}
</paper>

**Task**
Identify between 3 and 5 distinct inspiration threads reflected in how the paper
discusses, organizes, and builds upon prior work.

An inspiration thread is one coherent line of prior work representing a distinct
scientific direction that informs the development of the paper. Different
inspiration threads should correspond to distinct lines of prior work at a
comparable level of abstraction, rather than broad research directions and their
finer grained subdirections.

For each inspiration thread:

1. Inspiration Thread Identification
- Identify the distinct inspiration threads reflected in the paper's discussion of
prior work and motivation (e.g., the Introduction, Related Work, and other
relevant sections).
- Each inspiration thread should correspond to one coherent line of prior work
representing a distinct scientific direction.
- Define all inspiration threads at a comparable level of abstraction. Avoid
creating separate threads for closely related subdirections within the same line
of prior work.
- Do not merge distinct lines of prior work simply because they belong to the same
broader research area. Likewise, do not split a single line of prior work into
multiple threads because of method variants or incremental improvements.
- Assign a short, canonical name that reflects the shared scientific direction
represented by the thread. Prefer naming the underlying scientific direction over
a representative paper, specific method, or implementation.

2. Thread Description
- Describe the scientific direction represented by the thread, the shared
scientific insight unifying the prior work belonging to it, and how this direction
contributes to the development of the target paper.
- Focus on the scientific direction and shared scientific insight rather than
paper-specific methods, techniques, terminology, or implementations introduced by
individual papers.
- Clearly distinguish this thread from the other inspiration threads by explaining
what makes its scientific direction unique.

3. Thread-Specific Retrieval Query
- Generate a retrieval query by specializing the source paper's research question
toward the scientific direction represented by this inspiration thread. The query
should reflect how the authors would investigate this perspective in the context
of their own research problem.
- The query should reflect how the authors framed this scientific perspective
within their overall research problem before developing the solution.
- Briefly describe the line of prior work represented by this perspective, then end
with exactly one research question.
- The query should be specific enough to retrieve the prior work that most
substantially contributed to the target paper from this perspective, rather than
broadly retrieving related work.
- Write the retrieval query as continuous natural language rather than keywords,
concept lists, paper titles, or Boolean style search queries.
- Do not reveal or imply the paper's eventual solution, proposed paradigm, or
high-level approach.
- Do not mention methods, systems, datasets, benchmarks, coined terminology, or
other solution-specific concepts introduced by the paper itself.

**CRITICAL:**
- Write as continuous natural language rather than keywords, concept lists, paper
titles, or multiple questions.
- Prefer clear, direct sentences. Do not use em dashes or semicolons.

**Return a JSON object with one field "threads" containing a list of thread
objects.**

**Format**
{
  "threads": [
    {
      "thread_name": "short, canonical name of the scientific direction
represented by this inspiration thread",
      "description": "the scientific direction represented by this inspiration
thread, the shared scientific insight represented by the thread, and the
scientific role this direction serves in relation to the paper's contribution",
      "query": "natural-language retrieval query for retrieving papers from this
inspiration thread"
    }
  ]
}
\end{plumpromptbox}

\begin{plumpromptbox}{Prompt for Extracting Subfield-specific Inspiration}
You are an expert research analyst.
Given a research paper, its bibliography, and a set of pre-identified inspiration
threads, identify the key cited papers in the bibliography for each inspiration
thread.

<paper>
{paper_text}
</paper>

<bibliography>
{bibliography}
</bibliography>

<threads>
{threads_block}
</threads>

**Task**
For each inspiration thread, identify the cited papers that contributed most
substantially to the development of the target paper from the scientific
perspective represented by that thread.

**Guidelines:**
For each inspiration thread:
1. Key Inspiration Paper Selection
- Select only papers whose primary scientific contributions were essential to the
development of the target paper within this perspective.
- Select at most 5 papers for this thread, choosing the ones whose primary
scientific contributions were most essential to the development of the target
paper within this perspective.
- Use the paper's discussion of prior work as the primary evidence for selecting
key papers for inspiration threads.
- Avoid selecting a paper merely because it shares the thread's general topic.
Select it only if its specific contribution was essential to the target paper's
approach within this thread.
- Do not select papers that are cited primarily for background, comparison,
datasets, benchmarks, or implementation details unless they also made a core
scientific contribution within this perspective.
- Use the inline citation markers (e.g., [CITE:key]) together with the surrounding
discussion to determine the scientific role of each cited paper.

2. Rationale
- Explain the shared scientific contribution represented by the selected papers.
- Explain how this contribution was interpreted, adapted, extended, or transformed
within the target paper.
- Explain why this shared contribution makes these papers the most important prior
work from this scientific perspective, compared with other related cited papers.
- Do not refer to the selected papers by their citation keys, titles, or author
names. Instead, describe their shared scientific contribution in natural language.

**CRITICAL:**
- `cite_key` must exactly match a key listed in the bibliography.
- NEVER invent, modify, shorten, normalize, paraphrase, or partially copy a
`cite_key`.
- Write as continuous natural language rather than keywords, concept lists, paper
titles, or multiple questions.
- Prefer clear, direct sentences. Do not use em dashes or semicolons.

**Return a JSON object:**
{
  "threads": [
    {
      "thread_name": "exact thread name as given",
      "cite_keys": ["key 1", "key 2", ..., "key n"],
      "rationale": "A dense paragraph explaining how the selected papers
collectively contributed to the target paper from the perspective represented by
this thread, what key scientific contributions they provided, how those
contributions influenced the target paper, and why these papers were selected
over other related cited papers within this perspective."
    }
  ]
}

Return ONLY the JSON object, nothing else.
\end{plumpromptbox}

\begin{plumpromptbox}{Prompt for Reranking Retrieved Candidates}
You are an expert in scientific literature analysis. Your task is to rank
candidate papers according to how well they satisfy the scientific relevance
criterion below.

{query_block}

Reference Scientific Contribution:
{reference_rationale}

Scientific Relevance Criterion:
- Use the reference scientific contribution above to infer the scientific
contribution criterion for this question.
- Rank candidates higher if their primary scientific contributions provide
ideas, insights, formulations, or methodologies that satisfy this criterion and
would meaningfully inform research on this question.
- Rank candidates lower if they are merely topically related, address a similar
problem, use similar methods, datasets, or benchmarks, or belong to the same
broad research area without satisfying the scientific contribution criterion
represented above.
- Evaluate candidates based on their scientific contributions rather than their
topical similarity.

Candidates:
{listing}

Rank all candidate papers from most to least likely to satisfy the scientific
relevance criterion above.

Return only valid JSON in the following format:
{"ranked": [every candidate number, ordered from most to least suitable]}
\end{plumpromptbox}

\end{document}